\documentclass[sigconf]{acmart}
\acmConference[To appear in ACM CCS]{ACM Conference on Computer and Communications Security}
\setcopyright{none} 
\acmYear{2021}

\usepackage{thmtools}
\usepackage{thm-restate}
\usepackage{hyperref}
\usepackage{cleveref}

\newtheorem{assumption}{Assumption}
\usepackage{colortbl}
\usepackage{tikz}
\usepackage{amsmath}

\usepackage{amssymb}
\usepackage{bm}
\usepackage{algorithm}  
\usepackage{algpseudocode}
\usepackage{multicol}
\usepackage{multirow}
\usepackage{enumitem}
\usepackage{graphicx}
\usepackage{amsfonts}
\usepackage{url}
\usepackage[caption=false,font=footnotesize]{subfig}
\usepackage{color}

\usepackage{filecontents}

\AtBeginDocument{%
  \providecommand\BibTeX{{%
    \normalfont B\kern-0.5em{\scshape i\kern-0.25em b}\kern-0.8em\TeX}}}

\copyrightyear{2021}
\acmYear{2021}
\setcopyright{acmcopyright}\acmConference[CCS '21]{Proceedings of the 2021 ACM SIGSAC Conference on Computer and Communications Security}{November 15--19, 2021}{Virtual Event, Republic of Korea}
\acmBooktitle{Proceedings of the 2021 ACM SIGSAC Conference on Computer and Communications Security (CCS '21), November 15--19, 2021, Virtual Event, Republic of Korea}
\acmPrice{15.00}
\acmDOI{10.1145/3460120.3484796}
\acmISBN{978-1-4503-8454-4/21/11}
\begin{document}

\fancyhead{}
\fancyfoot{}
\title{A Hard Label Black-box Adversarial Attack Against Graph Neural Networks} 

\author{Jiaming Mu$^{1}$, Binghui Wang$^{2}$, Qi Li$^{1}$, Kun Sun$^{3}$, Mingwei Xu$^{1}$, Zhuotao Liu$^{1}$} 
\affiliation{
$^{1}$Institute for Network Sciences and Cyberspace, Department of Computer Science, and BNRist, Tsinghua University
\\
$^{2}$ Illinois Institute of Technology  $^{3} $George Mason University\\
\{mujm19@mails, qli01@, xumw@, zhuotaoliu@\}tsinghua.edu.cn, bwang70@iit.edu, ksun3@gmu.edu
}

\newcommand{\qi}[1]{\textcolor{orange}{\{{Qi:} #1\}}}

\newcommand{\KS}[1]{\textcolor{blue}{\{{KS:} #1\}}}

\newcommand{\M}[1]{\textcolor{black}{ #1}}

\begin{abstract}

  Graph Neural Networks (GNNs) have achieved state-of-the-art performance in 
  various graph structure related tasks such as node classification and graph classification. However, GNNs are vulnerable to adversarial attacks. Existing works mainly focus on attacking GNNs for node classification; nevertheless, the 
  attacks against GNNs for graph classification have not been well explored.  
  
In this work, we conduct 
a systematic 
  study on adversarial attacks against GNNs for graph classification via perturbing the graph structure. In particular, we focus on the most challenging attack, i.e., \emph{hard label black-box} attack, where an attacker has no knowledge about the target GNN model and can only obtain predicted labels 
  through querying the target model. 
  To achieve this goal, we formulate our attack as an optimization problem,
  whose objective is to 
  minimize the number of edges to be perturbed in a graph while maintaining the high attack success rate. 
  The original optimization problem is intractable to solve, and we relax the optimization problem to be a tractable one, which is solved with theoretical convergence guarantee. 
  We also design a coarse-grained searching algorithm and a  query-efficient gradient computation algorithm to decrease the number of queries to the target GNN model. Our experimental results on three real-world datasets demonstrate that our attack can effectively attack representative GNNs for graph classification with less queries and perturbations. 
  We also evaluate 
  the effectiveness of our attack under two defenses: one is  well-designed adversarial graph detector 
  {and the other is that the target GNN model itself is equipped with a defense to prevent adversarial graph generation.} 
  Our experimental results show that such 
  {defenses} are not effective enough, which highlights more advanced defenses.

\end{abstract}






\maketitle

\newcommand{\reducedrefsize}{\fontsize{8}{8}\selectfont}
{\reducedrefsize \textbf{ACM Reference Format:}\\
{Jiaming Mu, Binghui Wang, Qi Li, Kun Sun, Mingwei Xu, Zhuotao Liu. 2021. A Hard Label Black-box Adversarial Attack Against Graph Neural Networks. In \textit{Proceedings of the 2021 ACM SIGSAC Conference on Computer and Communications Security (CCS '21), November 15--19, 2021, Virtual Event, Republic of Korea}. ACM, NewYork, NY, USA, 18 pages. \url{https://doi.org/10.1145/3460120.3484796}}}

\section{Introduction}
\label{sec:intro}

Graph neural networks (GNNs) have been widely applied to various graph structure related tasks, 
e.g., node classification~\cite{kipf2016semi}, link prediction~\cite{zhang2018link}, and graph classification~\cite{zhou2019meta,xu2018powerful,ying2018hierarchical}, and achieved state-of-the-art performance.
{For instance, in graph classification, given a set of graphs and each graph is associated with a label, a GNN learns the patterns of the graphs by minimizing the cross entropy between the predicted labels and the true labels of these graphs~\cite{xu2018powerful} and predicts a label for each graph. 
GNN has been used to perform graph classification in various applications such as 
malware detection~\cite{wang2019heterogeneous}, brain data analysis~\cite{ma2019deep}, superpixel graph classification~\cite{avelar2020superpixel}, and protein pattern classification~\cite{tang2020adversarial}.} 

While GNNs significantly boost the performance of graph data processing, existing studies show that GNNs are vulnerable to  adversarial attacks~\cite{dai2018adversarial,wu2019adversarial,lin2020adversarial,tang2020adversarial,zugner2018adversarial,sun2019node, tang2020adversarial}. 
However, almost all the existing attacks focus on attacking GNNs for node classification, leaving
attacks against GNNs for graph classification largely unexplored,
though graph classification has been widely applied
~\cite{wang2019heterogeneous,ma2019deep,avelar2020superpixel,tang2020adversarial}. 
{Specifically, given a well-trained GNN model for graph classification and a target graph, 
an attacker aims to  
perturb
the structure (e.g., delete existing edges, add new edges, or rewire edges~\cite{ma2019attacking}) of the target graph
such that the GNN model will make a wrong prediction 
for the target graph. 
Such adversarial attacks could cause serious security issues. 
For instance, in malware detection~\cite{wang2019heterogeneous},  by intentionally perturbing a malware graph constructed by a certain malicious program, the malware detector could misclassify the malware to be benign. 
Therefore, we highlight that it is vital to explore the security of GNNs for graph classification under attack.
}

In this work, 
we investigate the most challenging and practical attack, termed \textit{hard label and black-box adversarial attack}, 
against GNNs for graph classification. 
In this attack, an attacker cannot obtain any information about the target GNN model
and can only obtain \textit{hard labels} (i.e., no knowledge of the probabilities associated with the predicted labels) through querying the GNN model. 
In addition, we consider that the attacker performs the attack by  perturbing the graph structure. 
The attacker's goal is then to fool the target GNN model by utilizing the hard label after querying the target model and with the minimal graph structural perturbations.

We formulate the attack as \M{a discrete} optimization problem, which aims to minimize the graph structure perturbations while maintaining high attack success rates.  \M{Note that our attack is harder than the existing black-box attacks (e.g., \cite{cheng2018query, cheng2019sign}) that are continuous optimization problems. 
} 
It is intractable to solve the formulated optimization problem due to the following reasons: (i) The objective function involves the $L_0$ norm, i.e., the number of perturbed edges in the target graph, and it is hard to be computed.
(ii) The searching space for finding the edge perturbations increases exponentially as the number of nodes in a graph increases. That is, it is time-consuming and query-expensive to find appropriate initial perturbations. 
To address these challenges, we propose a three-phase method to {construct our attack}. 
First, we convert the intractable optimization problem to a tractable one via relaxing the $L_0$ norm to be the $L_1$ norm, where gradient descent can be applied. 
Second, we propose a coarse-grained searching algorithm to significantly \M{ reduce the search space and 
efficiently identify initial perturbations, i.e., a much smaller number of edges in the target graph to be perturbed. Note that this algorithm can effectively exploit the graph structural information.}
Third, we propose a query-efficient gradient computation (QEGC) algorithm to deal with hard labels and adopt the sign stochastic gradient descent (signSGD) algorithm to solve the reformulated attack problem. Note that our QEGC algorithm only needs one query each time to compute the sign of gradients. We also derive theoretical convergence guarantees of our attack. 

We systematically evaluate our attack and compare it with {two baseline attacks} on three real-world datasets, i.e., COIL~\cite{riesen2008iam,coil},
IMDB~\cite{yanardag2015deep},  
and NCI1~\cite{wale2008comparison,shervashidze2011weisfeiler} from three different fields~\cite{Morris+2020} and three representative GNN methods.
Our experimental results demonstrate that our attack can effectively generate adversarial graphs with smaller perturbations and significantly outperforms the baseline attacks. For example, when assuming the same number \M{of} edges (e.g., $10\%$ of the total edges in a graph) can be perturbed, 
our attack can successfully attack around $92\%$ of the testing graphs in the NCI1 dataset, while the state-of-the-art RL-S2V attack~\cite{dai2018adversarial} can only attack around $75\%$ of the testing graphs. Moreover, only 4.33 edges on average are perturbed by our attack, while the random attack perturbs 10 times more of the edges. 
Furthermore, to show the effectiveness of our coarse-grained searching algorithm, 
{we compare the performance of three different searching strategies.
} 
{The results show that coarse-grained searching can significantly speed up the initial searching procedure, e.g., it can reduce $84.85\%$ of the searching time on the NCI1 dataset. 
It can also help find initial perturbations that can achieve {higher} attack success rates, e.g., the success rate is improved by around 50\%. }
We also evaluate the effectiveness of the proposed query-efficient gradient computation algorithm. Experimental results show that it decreases the number of queries dramatically. For instance, on the IMDB dataset, our attack with query-efficient gradient computation only needs $13.90\%$ of the queries, compared with our attack without it.  

We also explore the countermeasures against the adversarial graphs generated by our attack. {Specifically, we propose two different defenses against our adversarial attack: one to detect adversarial graphs and the other \M{to} prevent adversarial graph generation.}
For the former defense, we train a binary GNN classifier, whose training dataset consists of both normal graphs and the corresponding adversarial graphs generated by our attack. 
Such a classifier 
aims to distinguish the structural difference between adversarial graphs and normal graphs. 
Then the trained classifier is used to
detect adversarial graphs generated by our attack on the testing graphs. 
Our experimental results indicate that such a detector is not effective enough to detect the adversarial graphs. For example, when applying the detector on the COIL dataset with 20\% of the total edges are allowed to be perturbed, $47.50\%$ of adversarial graphs can successfully evade the detector. 
For the latter one, we equip GNN methods with a defense strategy, in order to prevent the generation of adversarial graphs. Specifically, we generalize the low-rank based defense~\cite{entezari2020all} for node classification to graph classification. 
The main idea is that only 
low-valued singular components of the adjacency matrix of a graph are affected by the adversarial attacks. Therefore, we propose to discard low-valued singular components to reduce the effects caused by attacks.
Our experimental results show that such a defense achieves \M{a} clean accuracy-robustness tradeoff. 
Our contributions 
are summarized as follows:
\begin{itemize}[leftmargin=*]
\setlength{\itemsep}{0.1pt}
\item To our best knowledge, we develop the first {optimization-based} attack against GNNs for graph classification in the hard label and black-box setting. 
 \item We formulate our attack as an optimization problem 
 and solve the problem with convergence guarantee to implement the attack.
 \item We design a 
 coarse-grained searching algorithm and query-efficient algorithm 
 to significantly reduce the costs of our attack. 
\item We propose two different types of defenses against our attack. 
\item We systematically evaluate our attack and defenses on real-world datasets to demonstrate the effectiveness of our attack.
\end{itemize}





\section{Threat Model}
\label{sec:threat model}


\noindent \textbf{Attack Goal.} 
We consider adversarial attacks against GNNs for graph classification.
Specifically, given a well-trained {GNN model} $f$ for graph classification and a \emph{target graph} $G$ with a label $y_0$, an attacker aims to perturb the target graph (e.g., delete existing edges, add new edges, or rewire edges in the graph) 
such that the perturbed target graph (denoted as $G'$) is misclassified by the GNN model
$f$. 
The attacks can be classified into \textit{targeted attacks} and \textit{non-targeted attacks}. In targeted attacks, an attacker will set a \emph{target label}, e.g., $y_c$, for the target graph $G$. Then the attack succeeds, if the predicted label
of the perturbed graph is $y_c$. In non-targeted attacks, the attack succeeds as long as the predicted label of the perturbed graph is different from $y_0$. 
In this paper, we focus on non-targeted attacks and we will also show that our attack can be applied to targeted attacks in Section~\ref{sec:re-formulate}.

\noindent \textbf{Attackers' Prior Knowledge.}
We consider the strictest hard label black-box setting. Specifically, we assume that the attacker can only 
query the target GNN model $f$ with an input graph and obtain only the predicted hard label (instead of a confidence vector that indicates the probabilities that the graph belongs to each class) for the graph.   
All the other knowledge, e.g., training graphs, structures and  parameters of target GNN model, is unavailable to the attacker. 

\noindent \textbf{Attacker's Capabilities.}
An attacker can perform an adversarial attack by perturbing one of three components in a graph: (i) perturbing nodes, 
i.e., adding new nodes or deleting existing nodes; (ii) perturbing node feature matrix, 
i.e., modifying nodes' feature vectors; 
and (iii) perturbing edges,
i.e., adding new edges, deleting existing edges or rewiring edges, which ensures the total number of edges is unchanged. In this paper, we focus on 
perturbing edges,
which is practical in real-world scenarios.
For example, in a social network, an attacker can influence the interactions between user accounts (i.e., modifying the edge status). However, it is hard for the attacker to close legitimate accounts (i.e., deleting nodes) or to modify the personal information of legitimate accounts (i.e., modifying the features). \M{The attacker can also conduct adversarial attacks via adding new nodes to the target graph, which is called \textit{fake node injection attack}. However, its attack performance is significantly impacted by  locations of injected nodes, e.g., an attacker needs to add more edges if the injected nodes is on the boundary of a graph, which is however easily detected. What's worse, fake node injection only involves adding edges but it cannot delete edges. Thus, we focus on more generic cases, i.e., perturbing edges, in this paper.} More specifically, we assume that the attacker can add new edges and delete existing edges to generate perturbations.
To guarantee unnoticeable perturbations, we set a \textit{ budget} $b \in \left[0, 1\right]$ for perturbing each target graph. That is, perturbed graphs with a \emph{perturbation rate} $r$, i.e., fraction of edges in the target graph is perturbed, exceeds the budget $b$ are invalid. 

As an attacker is often charged according to the number of queries, e.g., querying the model deployed by machine-learning-as-a-service platforms, 
we also assume that an attacker attempts to reduce the number of queries to save economic costs. 
In summary, the attacker aims to guarantee the attack success rate with as few queries as possible. 
Note that it is often a trade-off between the budget 
and the number of queries. For instance, with a smaller budget,  
the attacker needs to query the target GNN model more times.
Our designed three-phase attack (see Section \ref{sec:algorithm}) will obtain a better trade-off. 

\section{Problem Formulation}
\label{sec:4formulat}
Given a target GNN model $f$ 
and a target graph $G$
with label $y_0$ 
(Please refer to Appendix~\ref{sec:background} for more background on GNNs for graph classification, due to space limitation), the attacker attempts to generate an \emph{untargeted adversarial graph} $G'$
by perturbing the adjacency matrix $A$ of $G$ to be $A'$, such that the predicted label of $G'$ will be different from $y_0$. 
Let the \emph{adversarial perturbation} be a binary matrix $\Theta \in {\{0,1\}}^{N \times N}$. 
For ease of description, we fix the entries in the lower triangular part of $\Theta$ to be 0,  
i.e., $\Theta_{ij}=0 \ \forall j\leq i$, and each entry in the upper triangular part indicates whether the corresponding edge is perturbed or not.
Specifically, $\Theta_{ij}=1, j>i$ means 
the attacker  changes the edge status between nodes $i$ and $j$, i.e., adding the new edge $(i,j)$ if there is no edge between them in the original graph $G$ or deleting the existing edge $(i,j)$ from $G$. We keep the edge status between $i$ and $j$ unchanged if $\Theta_{ij}=0, j>i$. 
Then the perturbed graph $A'$ can be generated by a perturbation function $h$, i.e., $A' = h\left(A,\Theta\right)$, and $h$ is defined as follows:
\begin{equation}\label{eq:perturb}
    h\left(A,\Theta\right)_{ij}=h\left(A,\Theta\right)_{ji}=\begin{cases}A_{ij}&\Theta_{ij}=0, j>i, \\
    \neg A_{ij} &\Theta_{ij}=1, j>i.\end{cases}
\end{equation}
Moreover, the attacker ensures that the perturbation rate 
$r$ 
will not exceed a given budget $b$. 
Formally, we formulate 
generating adversarial structural perturbations to a target graph (or called \emph{adversarial graphs}) as  the following optimization problem:
\begin{equation}\label{eq:obj1}
\begin{split}
    \Theta^{*} =\ \mathop{\arg\min}_{\Theta}\  &  ||A'-A\|_0, \quad \\
    \mathrm{subject\ to} \quad  & A' = h\left(A,\Theta\right), \\
    & f\left(A'\right)\ne y_0, \\ 
    & r \leq b,
\end{split}
\end{equation}
where 
$r$ is defined as $r = \|A'-A\|_0 / N(N-1)$ and $\| M \|_0$ is the $L_0$ norm of $M$, which counts the number of nonzero entries in $M$. 

\begin{figure*}[t]
\centering
\includegraphics[width=7in]{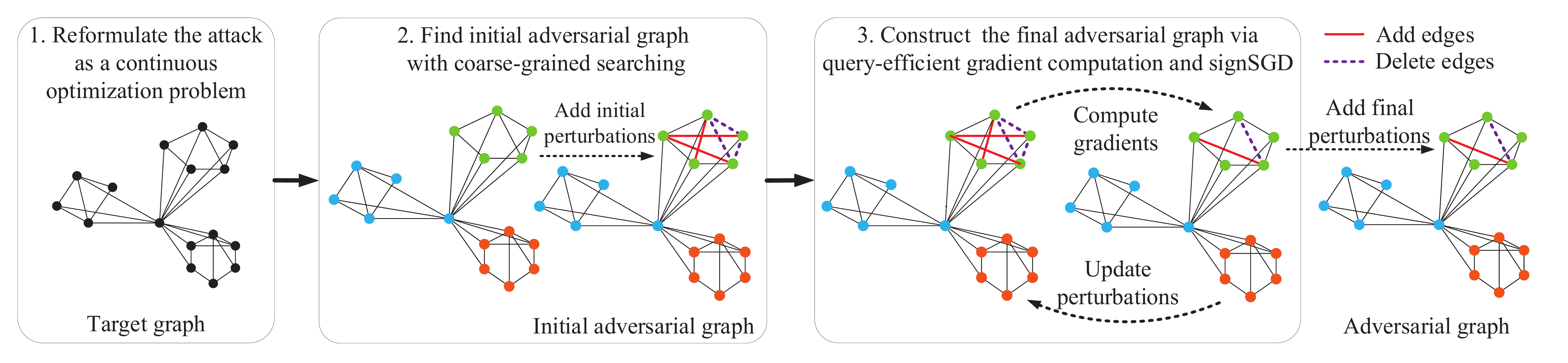}
\vspace{-7mm}
\caption{Overview of our hard label black-box attack: (1) We reformulate our 
attack as an continuous optimization problem that aims at minimizing 
perturbations on the target graph; (2) We design a coarse-grained searching algorithm to identify initial perturbations for efficient gradient descent computation; (3) We develop a query-efficient gradient computation algorithm, that only needs one query each time to compute the sign of gradients in signSGD. 
Finally, we obtain the adversarial graph via adding the final perturbations on the target graph.}\hfil
\vspace{-6mm}
\label{fig:flowchart}
\end{figure*}

\section{Constructing Adversarial Graphs}
\label{sec:algorithm}

In this section, we design our hard label black-box adversarial attack to construct adversarial graphs 
by solving the optimization problem in Eq. (\ref{eq:obj1}).

\subsection{Overview}
\label{sec:overviem}
Eq. (\ref{eq:obj1}) is an intractable optimization problem, and we cannot directly 
solve it. In order to address this issue, we convert the optimization problem into a tractable one and adopt a sign stochastic gradient descent (signSGD) algorithm to solve it with convergence guarantee. 
The signSGD algorithm computes gradients of graphs by iteratively querying the target GNN model. We also design two algorithms to 
reduce the number of queries: a coarse-grained searching algorithm by leveraging the graph structure and a query-efficient gradient computation algorithm that only requires one query in each time of computation. 
The overview of our attack framework is shown in Figure~\ref{fig:flowchart}. 
The attack consists of three phases.  
First, we relax the intractable optimization problem to a new tractable one 
(Section \ref{sec:re-formulate}).  
{Second, we develop a coarse-grained searching algorithm to 
identify a better initial adversarial perturbation/graph 
(Section \ref{sec:initial search}). }
Third, we propose a query-efficient gradient computation algorithm to deal with hard labels and construct the final adversarial graphs via  signSGD 
(Section \ref{sec:solve problem}).
The whole procedure of generating an adversarial graph for a given target graph is summarized in Algorithm \ref{alg:attack}.

\begin{algorithm}[t]
    \caption{Generating an adversarial graph for a target graph with a hard label black-box access}
    \label{alg:attack}
    \begin{algorithmic}[1]
        \Require
    A trained target GNN model $f$, a target graph $A$, perturbation budget $b$
        \Ensure
    Adversarial graph $A'$
    
    \State Search initial vector $\Theta_0$ via coarse-grained searching;
    \For{$t=1,2,\ldots,T$}
    \State Randomly sample $u_1, \ldots,u_Q$ from a Gaussian distribution;
    \State Compute $g(\Theta_t)$, $g(\Theta_t+\mu u_q)$ for  $q=1,\dots,Q$ via binary search;
    \State Compute $p(\Theta_t)$, $p(\Theta_t+\mu u_q)$  for  $q=1,\dots,Q$ using Eq. (\ref{eq:p_1});
    \State Estimate the gradient $\triangledown p\left(\Theta_t\right)$ using Eqs. (\ref{eq:sign}) and  (\ref{eq:gradient}); 
    \State Update $\Theta_{t+1}\leftarrow \Theta_t - \eta_t\triangledown p\left(\Theta_t\right)$;
    \EndFor
    \State Compute $A'=h(A,\Theta_T)$, $r = \|A'-A\|_0 / N(N-1)$;
    \If{$r\le b$} \textbf{return} $A'$ \# succeed
    \Else{} \textbf{return} $A$ \# failed
    \EndIf
    \end{algorithmic}
\end{algorithm}

\subsection{Reformulating the Optimization Problem}
\label{sec:re-formulate}
The optimization problem defined in 
Eq. (\ref{eq:obj1}) is intractable to solve. 
This is because the objective function involves $L_0$ norm related to the variables of adversarial perturbation  $\Theta$, which 
is naturally NP-hard. 
 To cope with this issue, 
 we use the following steps to reformulate the original optimization problem.

\noindent \textbf{Relaxing $\Theta$ to be continuous variables.}
We relax the binary entries $\{0,1 \}$ in $\Theta$ to be continuous variables ranging from $0$ to $1$, i.e., $\Theta_{ij} \in \left[0,1\right] \forall j>i$, such that we can approximate the gradients of the objective function. 
Each relaxed entry can be treated as the probability that  
the corresponding edge between two nodes is changed. Specifically, we perturb the edge status between node $i$ and node $j$ if $\Theta_{ij} \ge 0.5$; otherwise not. 
Thus, the perturbation function $h$ defined in Eq. (\ref{eq:perturb}) can be reformulated as follows:
\begin{equation}\label{eq:perturb new}
    h\left(A,\Theta\right)_{ij}=h\left(A,\Theta\right)_{ji}=\begin{cases}A_{ij}&\Theta_{ij}<0.5, j>i, \\
    \neg A_{ij} &\Theta_{ij} \ge 0.5, j>i.\end{cases}
\end{equation}

\begin{figure}[t]
\centering
\includegraphics[width=2.2in]{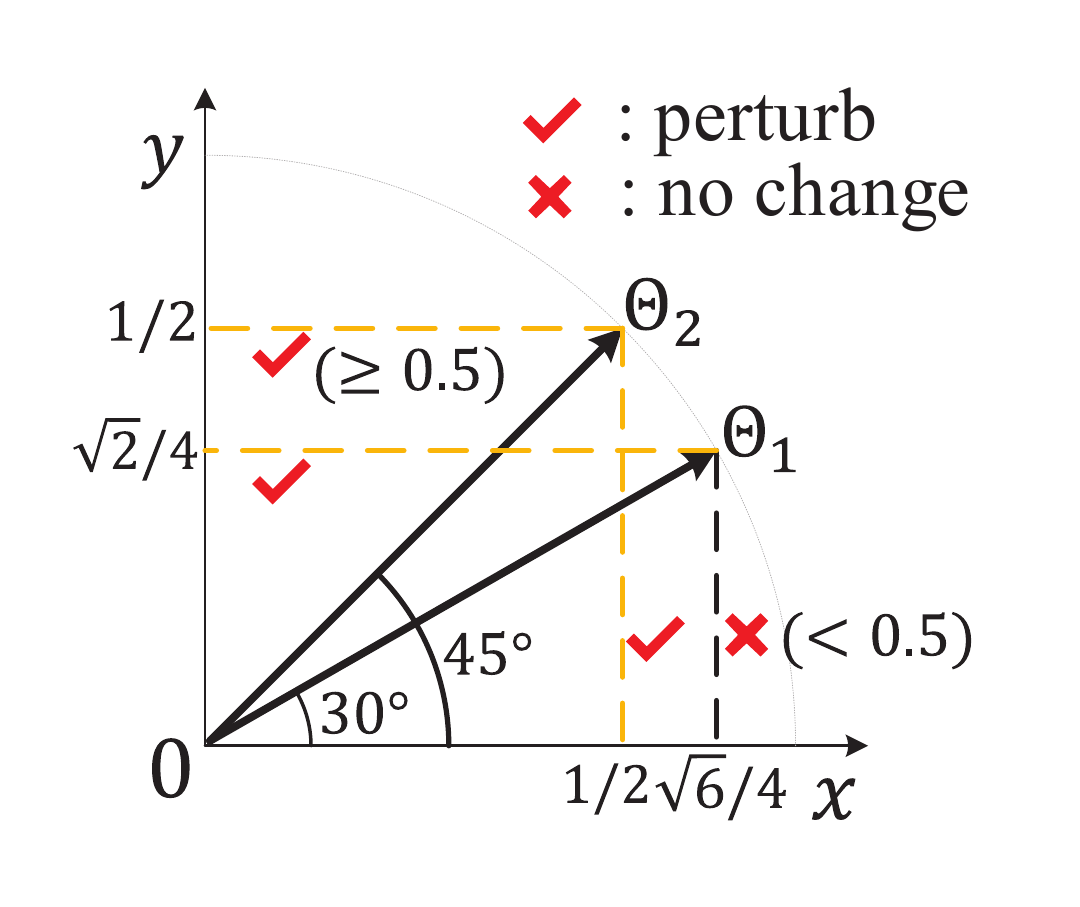}
\vspace{-5mm}
\caption{A toy sample: (i) Only one component of $\Theta_1$ exceeds 0.5, which means we only need to perturb one edge in the graph in the direction of $\Theta_1$; (ii) Both components of $\Theta_2$ achieve 0.5 and thus we need to perturb both the two edges in the direction of $\Theta_2$.}
 \vspace{-2mm}
\label{fig:toy sample}
\end{figure}

\noindent \textbf{Defining a new objective function.}
Next, we define a new objective function that replaces the $L_0$ norm with the $L_1$ norm.  
Similar to existing adversarial attacks against image classifiers~\cite{cheng2018query}, 
we can define a distance function $g\left(\Theta\right)$ for GNN to measure the distance from the target graph 
to the classification boundary as follows:
\begin{equation}\label{eq:g}
    g\left(\Theta\right)=\mathop{\arg\min}_{\lambda>0} \ \left\{f\left(h\left(A, \lambda\Theta_{norm}\right)\right)\ \ne y_0\right\},
\end{equation}
where 
$\Theta_{norm}$ is the normalized perturbation vector of the  perturbation vector $\Theta$\footnote{Without loss of generality, we transform the triangle perturbation matrix $\Theta$ to the corresponding vector form and use  perturbation vector and perturbation matrix interchangeably without otherwise mentioned. } 
that satisfies $\|\Theta_{norm}\|_2=1$. 
$g\left(\Theta\right)$ measures the distance from the original graph $A$ to the classification boundary, i.e., the minimal distance $\lambda$ if we start at $A$ and move to another class in the direction of $\Theta$ such that the predicted label of the perturbed graph $A'=h\left(A,\lambda\Theta_{norm}\right)$ changes. 
We also denote $\widehat{g}\left(\Theta \right)$ as a distance vector which starts from $A$ and ends at classification boundary at the direction of $\Theta$ with a length of $g\left(\Theta\right)$, i.e., $\widehat{g}\left(\Theta \right)=g\left(\Theta \right)\Theta_{norm}$. 

{A straightforward way of computing optimal $\Theta^{*}$ is to minimize $g(\Theta)$ because smaller $g(\Theta)$ may lead to less elements in $\Theta$ that exceed $0.5$. Thus, we should change less edges in $A$ for constructing the adversarial graphs.   
However, it is not effective enough 
as it does not consider the impact of the search direction, i.e., $\Theta$, on the attack.}
Specifically, the metrics of our attack is the number of 
perturbed edges
(i.e., the number of entries of $\Theta$ that exceed $0.5$) instead of the $L_2$ norm distance (i.e., $g\left(\Theta\right)$). 
The perturbations 
with different $\Theta_1$ and $\Theta_2$ may be different even if they share the equal distance (i.e., $g\left(\Theta_1\right)=g\left(\Theta_2\right)$). 
We explain this via a toy sample 
shown in Figure~\ref{fig:toy sample}. We assume that two distance vectors $\widehat{g}\left(\Theta_1 \right)$ and $\widehat{g}\left(\Theta_2 \right)$ with the dimension of $2$ have the same length of $\sqrt{2}/2$ in the direction of $\Theta_1$ and $\Theta_2$,  respectively. The \M{lengths} of the two components of $\widehat{g}\left(\Theta_1 \right)$ along with $x$-axis and $y$-axis are $\sqrt{6}/4$ and $\sqrt{2}/4$, respectively.
Thus, we only need to perturb one edge along with $x$-axis as only 
$\sqrt{6}/4 \geq 0.5$. However, the lengths of both two components of $\widehat{g}\left(\Theta_2 \right)$ are both $0.5$,
which means the attacker should perturb both edges because they both achieve the threshold 0.5. 
Motivated by this toy example and by considering both $\Theta$ and $g(\Theta)$, we define the following new objective function:
\begin{equation}\label{eq:p_0}
    p\left(\Theta\right)= \|clip(\widehat{g}\left(\Theta\right)-0.5)\|_0,
\end{equation}
where 
$clip(x)$ is a clip function which clips $x$ into $\left[0,1\right]$. $p\left(\Theta\right)$ denotes the number of elements of $\widehat{g}\left(\Theta\right)$ that exceed $0.5$. Thus, it can measure the desired perturbations in the direction of $\Theta$. Here, in order to calculate the gradients, we also replace the $L_0$ norm  in Eq. (\ref{eq:p_0}) with the $L_1$ norm as follows:
\begin{equation}\label{eq:p_1}
    p\left(\Theta\right)= \|clip(\widehat{g}\left(\Theta\right)-0.5)\|_1.
\end{equation}

\noindent \textbf{Converting the optimization problem.}
According to the definition of $p\left(\Theta\right)$, the attacker can find the optimal vector $\Theta^{*}$ by minimizing $p\left(\Theta\right)$. Finally, we convert the original optimization problem in Eq. (\ref{eq:obj1}) into a new one as follows:
\begin{equation}\label{eq:obj2}
\begin{split}
    \Theta^{*} =\ \mathop{\arg\min}_{\Theta}\  p\left(\Theta\right), \quad \mathrm{subject\ to} \quad r\leq b.
\end{split}
\end{equation}
Note that, (i) this optimization problem is designed for  non-targeted attacks. However, it can also be extended to targeted attacks via changing the condition in Eq. (\ref{eq:g}) to $f\left(h\left(A, \lambda\Theta_{norm}\right)\right)\ = y_c$, where $y_c$ is the target label; \M{(ii) Eq. (\ref{eq:obj2}) approximates Eq. (\ref{eq:obj1}). Note that, we cannot guarantee that Eq. (\ref{eq:obj1}) and (\ref{eq:obj2}) have exactly the same optimal values. Nevertheless, our experimental results show that solving Eq. (\ref{eq:obj2}) can achieve promising attack performance. 
}

\subsection{Coarse-Grained Searching}
\label{sec:initial search}
In this section, we develop a coarse-grained searching algorithm to efficiently identify an initial perturbation vector $\Theta_0$ that makes the corresponding adversarial graph have a different predicted label from $y_0$ in the direction specified by $\Theta_0$. 
We note that, it is difficult to find the valid $\Theta_0$
because the searching space is extremely large when the number of nodes is large. 
Specifically, a graph with $N$ nodes has $S = N\left(N-1\right)/2$ candidate edges. Each edge can be existent or nonexistent so that the search space has a volume of $2^{S}$, which increases exponentially as $N$ increases. The query and computation overhead of \M{traversing} all candidate graphs in the searching space is extremely large. 
Our coarse-grained searching algorithm aims to leverage the graph structure property to reduce the searching space. 

\begin{figure}[!t]
\centering
\includegraphics[width=2.8in]{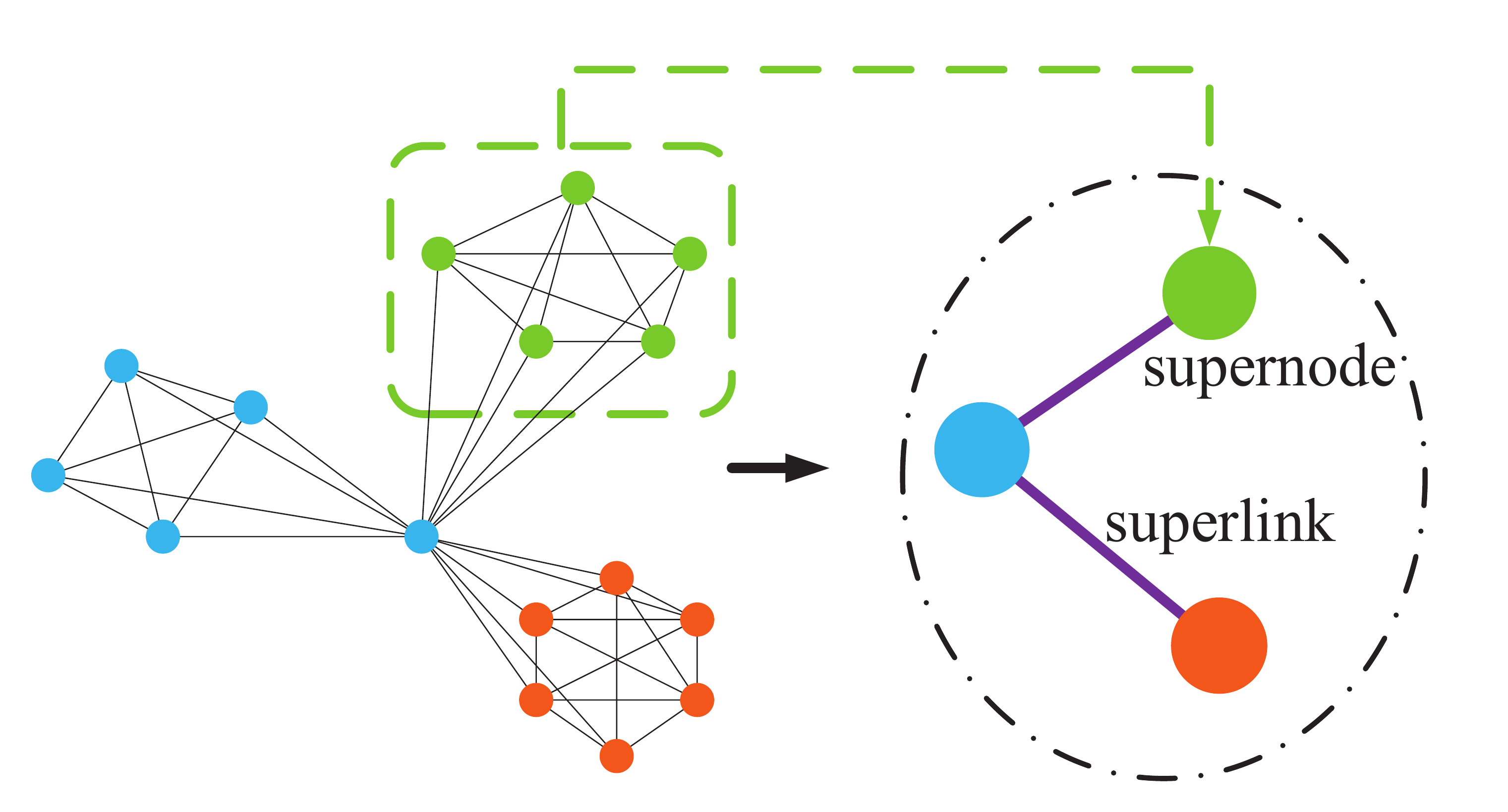}
\vspace{-4mm}
\caption{Coarse-grained searching. We partition the graph into several node clusters, denoted as \emph{supernodes}, and links between two supernodes are denoted as a \emph{superlink}.}
 \hfil
 \vspace{-6mm}
 \label{fig:CGS}
\end{figure}

We utilize the properties of a graph to reduce the number of the queries and to find a better initial $\Theta_0$ that incurs small perturbations.  Specifically, edges in a graph can reflect the similarity among nodes. For instance, there could be more number of edges within a set of nodes, but the number of edges between these nodes and other nodes is much smaller.  
This means this set of nodes are similar and we can group them into a node cluster. 
Inspired by graph partitioning~\cite{karypis1998fast}, 
we split the original graph into several node clusters,  where nodes within each cluster are more similar. We denote \textit{supernode} as one node cluster and \textit{superlink} as the set of links between nodes from two node clusters 
(See Figure \ref{fig:CGS}). Then, there are three components of a graph for us to search, i.e., (i) supernode; (ii) superlink; and (iii) the whole graph. 
We take turns to traverse these three types of searching spaces. 
{The reason why we target one specific type of components for  \M{perturbation} each time is that we can ensure the perturbations follow the same direction, and thus can 
more effectively generate an adversarial graph. 
Otherwise, perturbations added in different components will interfere with each other.} 
{Thus, randomly selecting $\Theta_0$ is not a good choice as it discards the structural information of the target graph. Our experimental results in Section \ref{sec:experiment results} also support our idea. }

Particularly, we first partition the graph into node clusters 
(or supernodes) using the popular and efficient Louvain algorithm~\cite{blondel2008fast}.
After that, we traverse each supernode. In each supernode $c$, we uniformly choose a fraction {$s \in [0,1]$} at random
to determine the number of perturbed edges $n$, i.e., $n = s \cdot N_c\left(N_c-1\right)/2$, where $N_c$ is the number of nodes in the supernode $c$. We then randomly select $n$ edges to be perturbed and query the target GNN model {to see whether the label of the target graph is changed}. We repeat the above process, e.g., $5 \cdot N_c$ times used in our experiments, and always keep the initial perturbation vector with minimal number of perturbed edges.
If we failed to find $\Theta_0$ that can change the target label after searching all supernodes, we then search the space within each superlink and finally the whole graph {if we still cannot find a successful $\Theta_0$}. 
During the searching process, we can thus maintain the perturbation vector with the smallest number of edges to be perturbed. 

Note that we can significantly reduce the searching overhead by searching supernodes, superlinks, and the whole graph in turn. First, we search supernodes before superlinks 
because the searching spaces defined by superlinks are larger than those of supernodes. It is not necessary to search within superlinks if we already find $\Theta_0$ within supernodes. 
Second, the size of the searching space defined by the whole graph is  $2^{S}$, which is query and time expensive. As we search it at last, we can find successful $\Theta_0$ in the former two phases (i.e., searching supernodes and superlinks) for most of the target graphs and only a few graphs need to search the whole space (see Section \ref{sec:experiment results}). 
Via performing coarse-grained searching, we can \emph{exponentially} reduce the time {and the number of queries} 
when the number of supernodes is far more smaller than the number of nodes. 
{The following theorem states the reduction in the time of space searching with our CGS algorithm:}

\begin{restatable}[]{thm}{reduction}
\label{theorem3}
Given a graph $G$ with $N$ nodes, 
the reduction, denoted as $\beta$, in the time of space searching with coarse-grained searching satisfies 
$\beta \approx O(2^{\kappa^4})$,
where $\kappa$ is the number of node clusters and we assume $\kappa \ll N$.
\end{restatable}
 
\begin{proof}
See Appendix~\ref{appendix:proof3}.
\end{proof}


\begin{figure}[t]
\vspace{-4mm}
\centering
\subfloat[Compute $p(\Theta)$ with binary search]{
\centering
\includegraphics[width=0.32\textwidth]{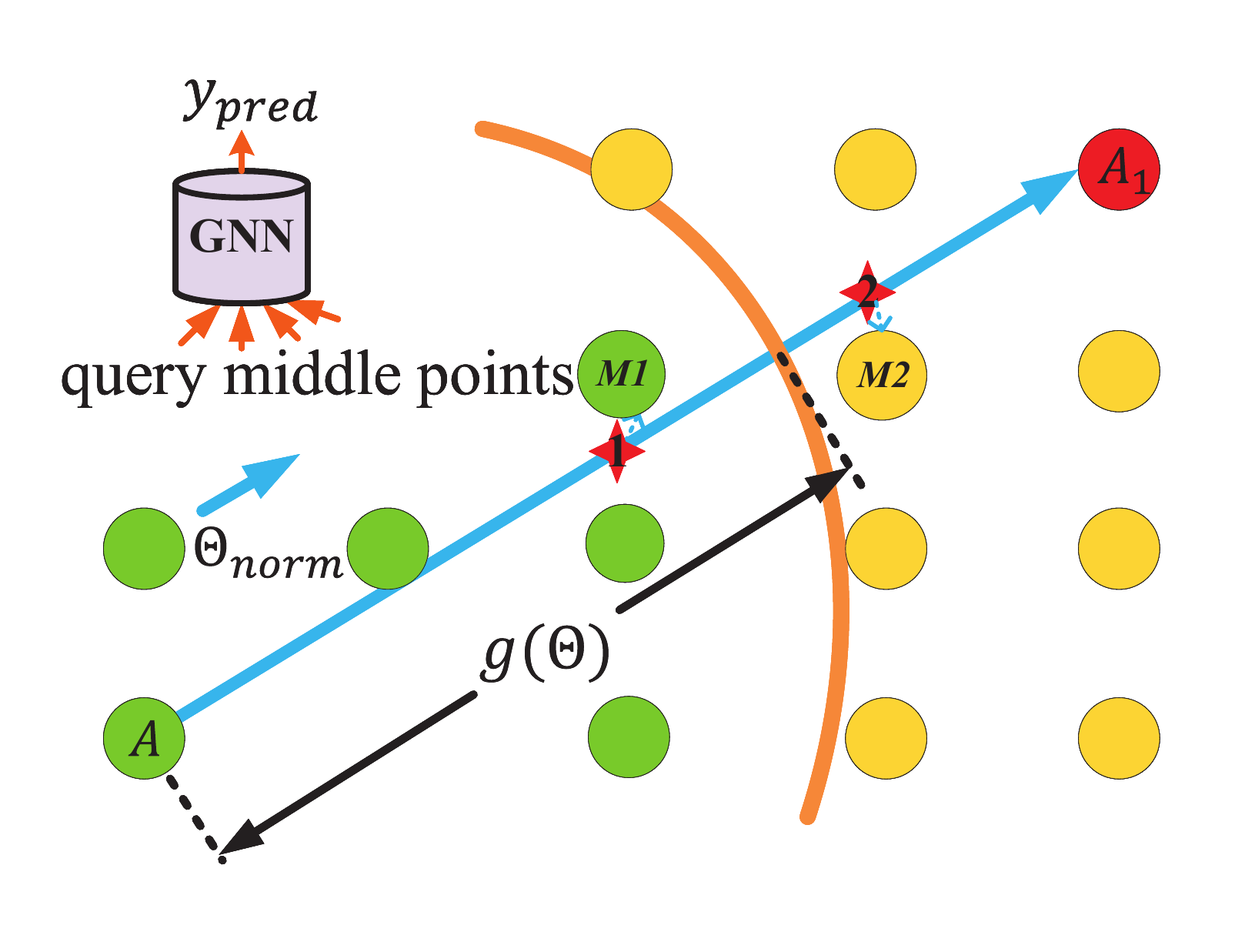}
\label{fig:3querymodel}}
\vspace{-2mm}
\\
\subfloat[Compute gradient of $p(\Theta)$ with QEGC]{
\centering
\includegraphics[width=0.32\textwidth]{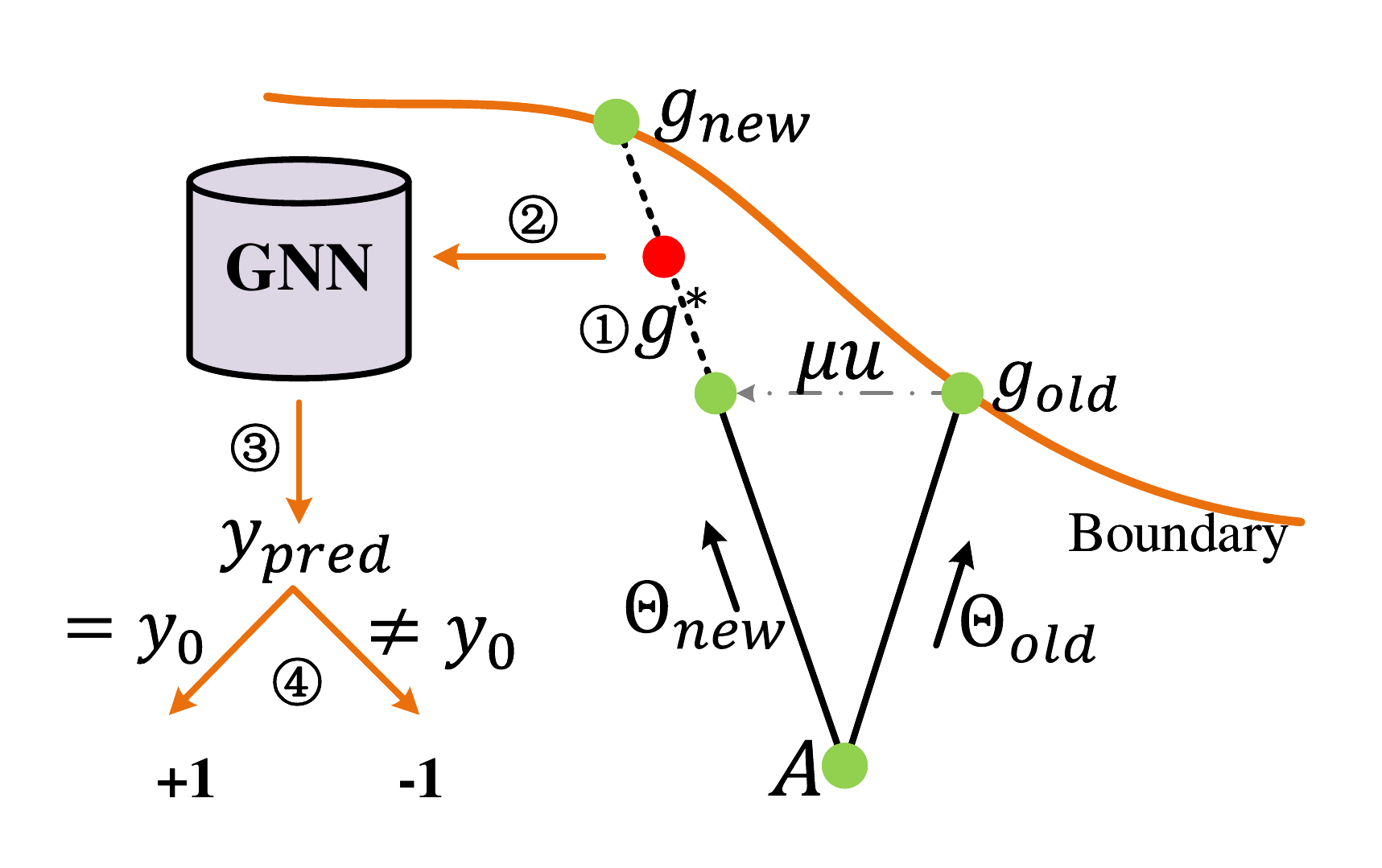}
\label{fig:3querymodel}}
\vspace{-2mm}
\caption{Constructing adversarial graphs. (a) Computing $g(\Theta)$ and $p(\Theta)$ by querying the target model  until we find the classification boundary, which will incur many queries when computing gradients of $p(\Theta)$ by using the zeroth order oracle; (b) Query-efficient gradient computation (QEGC). To compare $p_{new}$ and $p_{old}$ (i.e., compute the sign of gradient of $p(\Theta)$), we find $g^*$ in the direction of $\Theta_{new}$ such that $p^*=p_{old}$, and use the predicted label to judge if $p_{new}$ is larger than $p_{old}$ after querying $A^*=h(A,g^*\Theta_{new})$.}
\vspace{-6mm}
\label{fig:QEGC}
\end{figure}

\subsection{Generating Adversarial Graphs via \textit{SignSGD}}
\label{sec:solve problem}
Now we present our sign stochastic gradient descent (signSGD) algorithm to solve the attack optimization problem in Eq (\ref{eq:obj2}). 
Before presenting signSGD, we first describe the method to compute $p(\Theta)$, where only hard label is returned when querying the GNN model; and then introduce a query-efficient gradient computation  algorithm to compute the gradients of $p(\Theta)$. 

\noindent \textbf{Computing $p(\Theta)$ via binary search.}
We describe computing $p(\Theta)$ with only hard label black-box access to the target model. 

We first compute $g(\Theta)$ in Eq. (\ref{eq:g}) via repeatedly querying the target model and further obtain $p(\Theta)$ using Eq. (\ref{eq:p_1}). 
As shown in Figure \ref{fig:QEGC} (a), each edge in the edge space of $G$ can be either existent or nonexistent so that the searching space consists of lattice points \M{(i.e., a lattice point is a symmetrical binary matrix $M\in \{0, 1\}^{N\times N}$)} that have equal distance among each other. 
Suppose there is a classification boundary in the direction of $\Theta$. $g(\Theta)$ is the length of direction vector $\hat{g}(\Theta)$ that begins at the target graph $A$ and ends at the boundary.
We can first find a graph $A_1$ with a different label from $y_0$ using our CGS algorithm. Since there will be a classification boundary between $A$ and $A_1$, we can then 
conduct a binary search between 
{them,} i.e., we query the middle point of the range {[$A$, $A_1$] (e.g., $M_1$ in Figure \ref{fig:QEGC} (a))} and update the endpoints of the range based on the predicted label of the middle point in each iteration. The query process ends when the length of range decreases below a tolerance $\epsilon$. 
With such query process, we can obtain
$g(\Theta)$, and then we can compute $p(\Theta)$ easily.

\noindent \textbf{Computing the gradient of $p(\Theta)$ via query-efficient gradient computation (QEGC).} \label{sec:QEGC} 
We now propose a query efficient algorithm to compute the sign of 
gradient of $p(\Theta)$, that aims at saving queries used in signSGD in the next part.


With zeroth order oracle, we can estimate the sign of gradient of $p(\Theta)$ via computing $sign\left((p\left(\Theta+\mu u\right)-p\left(\Theta\right))/\mu u\right)$, where $u$ is a  normalized i.i.d direction vector sampled randomly from a Gaussian distribution, and $\mu$ is a step constant. The sign can be acquired by computing $p\left(\Theta+\mu u\right)$ and $p\left(\Theta\right)$ separately. However, we need multiple queries to obtain the value of $p(\Theta)$. As we need to update $\Theta$ with many iterations, it is query expensive to compute all $p\left(\Theta_t+\mu u\right)$ and $p\left(\Theta_t\right)$ at each iteration during the signSGD. Fortunately, we only need to know which $p$ is larger instead of the exact values of them. Thus, we propose a \emph{query-efficient gradient computation (QEGC)} algorithm to compute the sign of gradient with only one query a time as shown in Figure \ref{fig:QEGC} (b).

Suppose the current direction is $\Theta_{old}$ with $g\left(\Theta_{old}\right)=g_{old}$ and $p\left(\Theta_{old}\right)=p_{old}$. Now the direction steps forward with an increment of $\mu u$, i.e., $\Theta_{new} = \Theta_{old} +\mu u $. For simplicity of description, we assume $\Theta_{old}$ and $\Theta_{new}$ are both normalized vectors. We want to judge if $p_{new}$ is larger than $p_{old}$ or not.
The idea is that we transfer $p_{new}$ and $p_{old}$ to $g_{new}$ and $g_{old}$ respectively and compare their values. 
Specifically, for $p_{old}$, we find $g^{*}$ such that $p^{*} = \|clip(g^{*}\Theta_{new}-0.5)\|_1=p_{old}$. For $p_{new}$, the corresponding $g_{new}$ is the distance from $A$ to the classification boundary at the direction $\Theta_{new}$. Then we query the target model $f$ with graph $A^{*} = h\left(A, g^{*}\Theta_{new}\right)$ to figure out whether $g^{*}$ exceeds the boundary or not. We say that the classification boundary in the direction of $\Theta_{new}$ is closer than that of $\Theta_{old}$ if $f\left(A^{*}\right) \ne y_0$ because we cross the boundary with the same $p_{old}$ at the direction of $\Theta_{new}$, while we can only achieve the boundary (but not cross) at the direction of $\Theta_{old}$. Thus, $p_{new}$ is smaller than $p_{old}$ and the sign of gradient is $-1$. Similarly, $sign = +1$ if $f\left(A^{*}\right) = y_0$. 

In summary, we compute the sign of a gradient as follows:
\begin{equation}\label{eq:sign}
    sign\left(p\left(\Theta+\mu u\right)-p\left(\Theta\right)\right)=\begin{cases}+1&f\left(A^{*}\right) = y_0,\\
    -1 &f\left(A^{*}\right) \ne y_0,\end{cases} 
\end{equation}
where $A^{*}$ is the graph whose value of $p$ equals to $p\left(\Theta\right)$ in the direction of $\Theta + \mu u$. We can use Eq. (\ref{eq:sign}) to save the queries 
due to 
the following theorem. 


\begin{restatable}[]{thm}{monotonous}
\label{theorem:2}
Given a normalized direction $\Theta_{old}$ with $g_{old}$ and $p_{old}$, there is one and only one $g^{*}$ at the direction of $\Theta_{new}$ that satisfies $p^{*} = \|clip(g^{*}\Theta_{new}-0.5)\|_1=p_{old}$.
 \end{restatable}
 \begin{proof}
See Appendix \ref{appendix:proof2}.
\end{proof}

\noindent \textbf{Solving the converted attack problem via sign Stochastic Gradient Descent (signSGD).} 
We utilize the sign stochastic gradient descent (signSGD) algorithm~\cite{bernstein2018signsgd} to solve the converted optimization shown in Eq. (\ref{eq:obj2}). The reasons are twofold: 
(i) the sign operation that compresses the gradient into a binary value is suitable to the hard label scenario;
(ii) the sign of the gradient can approximate the exact gradient, which can significantly reduce the query overhead. 

Specifically, during the signSGD process, we use Eq. (\ref{eq:sign}) to compute the sign of gradient of $p(\Theta)$ in the direction of $u$. To ease the noise of gradients, we average the signs of $Q$ gradients in different directions to estimate the derivative of the vector $p(\Theta)$ as follows:
\begin{equation}\label{eq:gradient}
   \triangledown p\left(\Theta\right)=\frac{1}{Q}\sum_{q=1}^Q sign\left(\frac{p\left(\Theta+\mu u_q\right)-p\left(\Theta\right)}{\mu}u_q\right), 
\end{equation}
where $ \triangledown p\left(\Theta\right)$ is the estimated gradients of $p(\Theta)$, $u_q, q\in {1,2,\ldots,Q}$ are normalized i.i.d direction vectors sampled randomly from a Gaussian distribution, and $Q$ is the number of vectors.
\M{Recently, Maho et al.~\cite{maho2021surfree} proposed a black-box SurFree attack that also involves sampling the direction vector $u$ from a Gaussian distribution. However, the purpose of using $u$ is different from our method.  
Specifically, $u$ in the SurFree attack is used to compute the distance from the original sample to the boundary, while $u$ in our attack is used to approximate the gradients.}

\M{The sign calculated by Eq. (\ref{eq:sign}) depends on a single direction vector $u$. In contrast, Eq. (\ref{eq:gradient}) computes the sign of the average of multiple directions, and can better approximate the real sign of gradient of $p(\Theta)$.} Then, we use this gradient estimation to update the search vector $\Theta$ by computing $\Theta_{t+1}\leftarrow\Theta_t-\eta_t \triangledown p\left(\Theta_t\right)$, where $\eta_t$ is the learning rate in the $t$-th iteration. After $T$ iterations, we can construct an adversarial graph 
$A'=h(A, \Theta_T)$ 
\footnote{\M{Our attack can be easily extended to attack directed graphs via only changing the adjacency matrix for directed graphs.}}.
The following theorem shows the convergence guarantees of our signSGD for generating adversarial graphs. 

\begin{assumption}\label{assump:2}
At any time $t$, the gradient of the function $p(\Theta)$ is upper bounded by $\|\triangledown p\left(\Theta_t\right)\|_2 \le \sigma$, \M{where $\sigma$ is a non-negative constant.}
\end{assumption}

\begin{restatable}[]{thm}{convergence}
\label{theorem:1}

Suppose that $p\left(\Theta\right)$ has $L$-Lipschitz continuous gradients and 
Assumption \ref{assump:2} holds. If we randomly pick $\Theta_R$, \M{whose dimensionality is $d$, } from $\left\{\Theta_t\right\}_{t=0}^{T-1}$ with probability $P\left(R=t\right)=\frac{\eta_t}{\sum_{t=0}^{T-1} \eta_t}$, the convergence rate of our signSGD with $\eta_t=O\left(\frac{1}{\sqrt{dT}}\right)$ and $\mu=O\left(\frac{1}{\sqrt{dT}}\right)$ will give the following bound on $\mathbb{E}\left[\|\triangledown p\left(\Theta\right)\|_2\right]$
\begin{equation}
    \mathbb{E}\left[\|\triangledown p\left(\Theta\right)\|_2\right]=O\Big(\frac{\sqrt{d} L}{\sqrt{T}}+\frac{\sqrt{d}}{\sqrt{Q}}\sqrt{Q+d}\Big),
\end{equation}
 \end{restatable}
\begin{proof}
See Appendix \ref{appendix:proof1}.
\end{proof}


\section{Attack Results}
\label{sec:experiments}

In this section, 
we evaluate the effectiveness our hard label black-box attacks against GNNs for graph classification. 

\subsection{Experimental Setup}
\label{sec:experiment setup}
\noindent \textbf{Datasets.} We use three real-world graph datasets from three different fields to construct our adversarial attacks, i.e.,  COIL~\cite{riesen2008iam,coil} in the computer vision field, IMDB~\cite{yanardag2015deep} in the social networks field, and NCI1~\cite{wale2008comparison,shervashidze2011weisfeiler} in the small molecule field. 
Detailed statistics of these datasets are in Table \ref{tab:info of datasets}. 
By using 
datasets from different fields with different sizes, 
we can effectively evaluate the effectiveness of our attacks in different real-world scenarios. 
We randomly split each dataset into 10 equal parts, of which 9 parts are used to train the target GNN model and the other 1 part is used for testing. 

\noindent \textbf{Target GNN model.} We choose three representative GNN models, i.e., GIN~\cite{xu2018powerful}, SAG~\cite{lee2019self}, and GUNet~\cite{gao2019graph} as the target GNN model. We train these models based on the authors' public available source code. 
The clean training/testing accuracy (without attack) of the three GNN models on the three graph datasets are shown in Table~\ref{tab:info of GIN models}. \M{Note that these results are close to those reported in the original papers.} 
We can see that GIN achieves the best testing accuracy.  
Thus, we use GIN as the default target model in this paper, unless otherwise mentioned. 
We also observe that SAG and GUNet perform bad on COIL, and we thus do not conduct attacks on COIL for SAG and GUNet.

\begin{table}[t]\renewcommand{\arraystretch}{1.2}
\centering
    \caption{Dataset statistics.}
    \vspace{-3mm}  
    \label{tab:info of datasets}
      \begin{tabular}{c|c|c|c}
      \hline
        Dataset & IMDB & COIL & NCI1 \\
        \hline
        \hline
        Num. of Graphs  & 1000 & 3900 & 4110 \\
        Num. of Classes & 2 & 100 & 2 \\
        Avg. Num. of Nodes  & 19.77& 21.54 & 29.87 \\
        Avg. Num. of Edges& 96.53  & 54.24 & 32.30 \\
        \hline
        \hline
      \end{tabular}
\end{table}

\begin{table}[t]
\centering
    \caption{Clean accuracy of the three GNN models.}
    \vspace{-3mm}  
    \label{tab:info of GIN models}
      \begin{tabular}{c|c|cc}
      \hline
      \multirow{1}{*}{GNN model}
        & Dataset & Train acc & Test acc \cr
        \hline
        \hline
        \multirow{3}{*}{GIN}
        &COIL  & $82.17\%$ & $77.95\%$ \cr
        &IMDB & $69.44\%$ & $77.00\%$ \cr
        &NCI1 & $73.59\%$ & $77.37\%$ \cr
        \cline{1-4}
        \multirow{3}{*}{SAG}
        &COIL & $40.85\%$ & $42.56\%$\cr
        &IMDB & $64.78\%$ & $68.00\%$\cr
        &NCI1 & $73.18\%$ & $72.02\%$\cr
        \cline{1-4}
        \multirow{3}{*}{GUNet}
        &COIL & $31.25\%$ & $31.03\%$\cr
        &IMDB & $64.44\%$ & $70.00\%$\cr
        &NCI1 & $69.59\%$ & $76.16\%$\cr
        \cline{1-4}
        \hline
        \hline
      \end{tabular}
\vspace{-4mm}
\end{table}

\noindent \textbf{Target graphs.} We focus on generating untargeted adversarial graphs, i.e., an attacker tries to deceive the target GNN model to output each testing graph a wrong label different from its original label. 
In our experiments, we select all testing graphs that are correctly classified by the target GNN model as the target graph. For example, the number of target graphs for GIN are 304 on COIL, 77 on IMDB, and 318 on NCI1, respectively.  

\noindent \textbf{Metrics.} We use {four} metrics to evaluate the effectiveness of our attacks: (i) Success Rate (SR), 
i.e., the fraction of successful adversarial graphs over all the target graphs. 
(ii) Average Perturbation (AP), i.e., the average number of perturbed edges across the successful adversarial graphs.
(iii) Average Queries (AQ), i.e., the average number of queries used in the whole attack.
(iv) Average Time (AT), i.e., the average time used in the whole attack.
We count queries and time for all target graphs even if the attack fails. 
Note that an attack has better attack performance if it achieves a larger SR or/and a smaller AP, AQ and AT. 

\noindent \textbf{Baselines.} 
We compare our attack with state-of-the-art RL-S2V attack~\cite{dai2018adversarial}. We also choose random attack as a baseline. 

\begin{figure*}[!t]
\centering
\subfloat[NCI1:GIN]{
\includegraphics[width=2in]{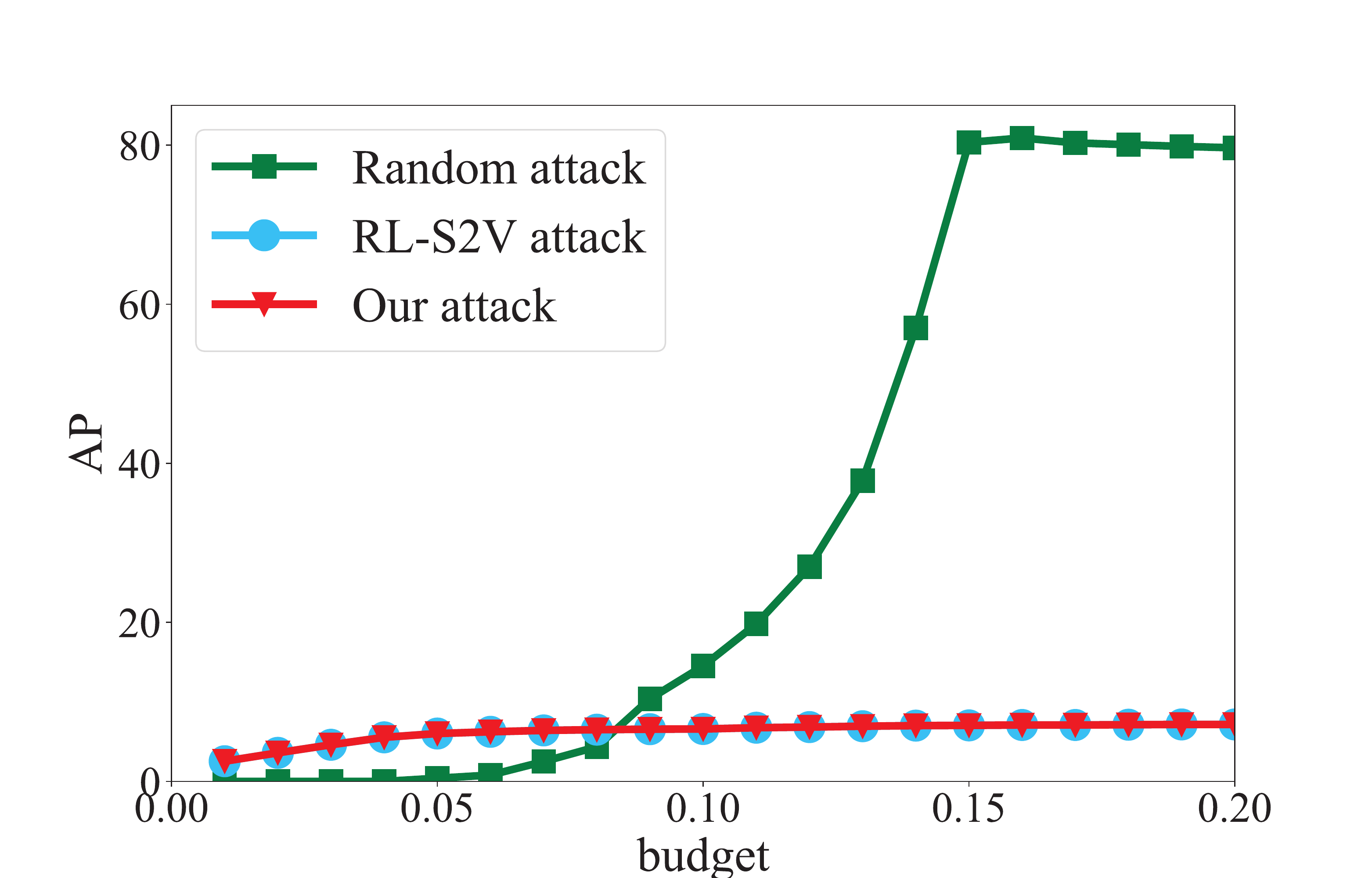}
\label{NCI1_AP}}
 \hfil
\subfloat[COIL:GIN]{\includegraphics[width=2in]{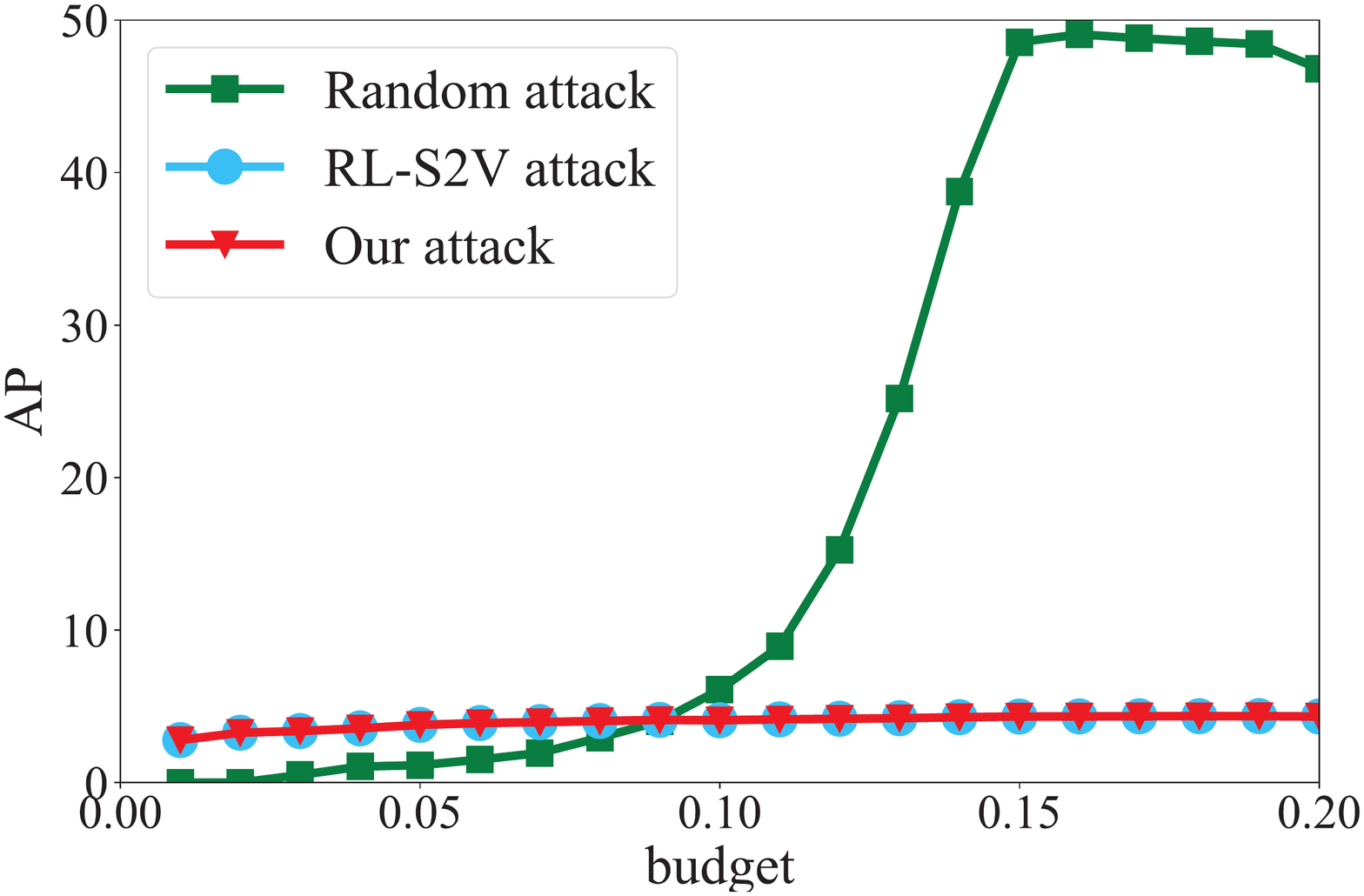}
\label{COIL_AP}}
 \hfil
\subfloat[IMDB:GIN]{\includegraphics[width=2in]{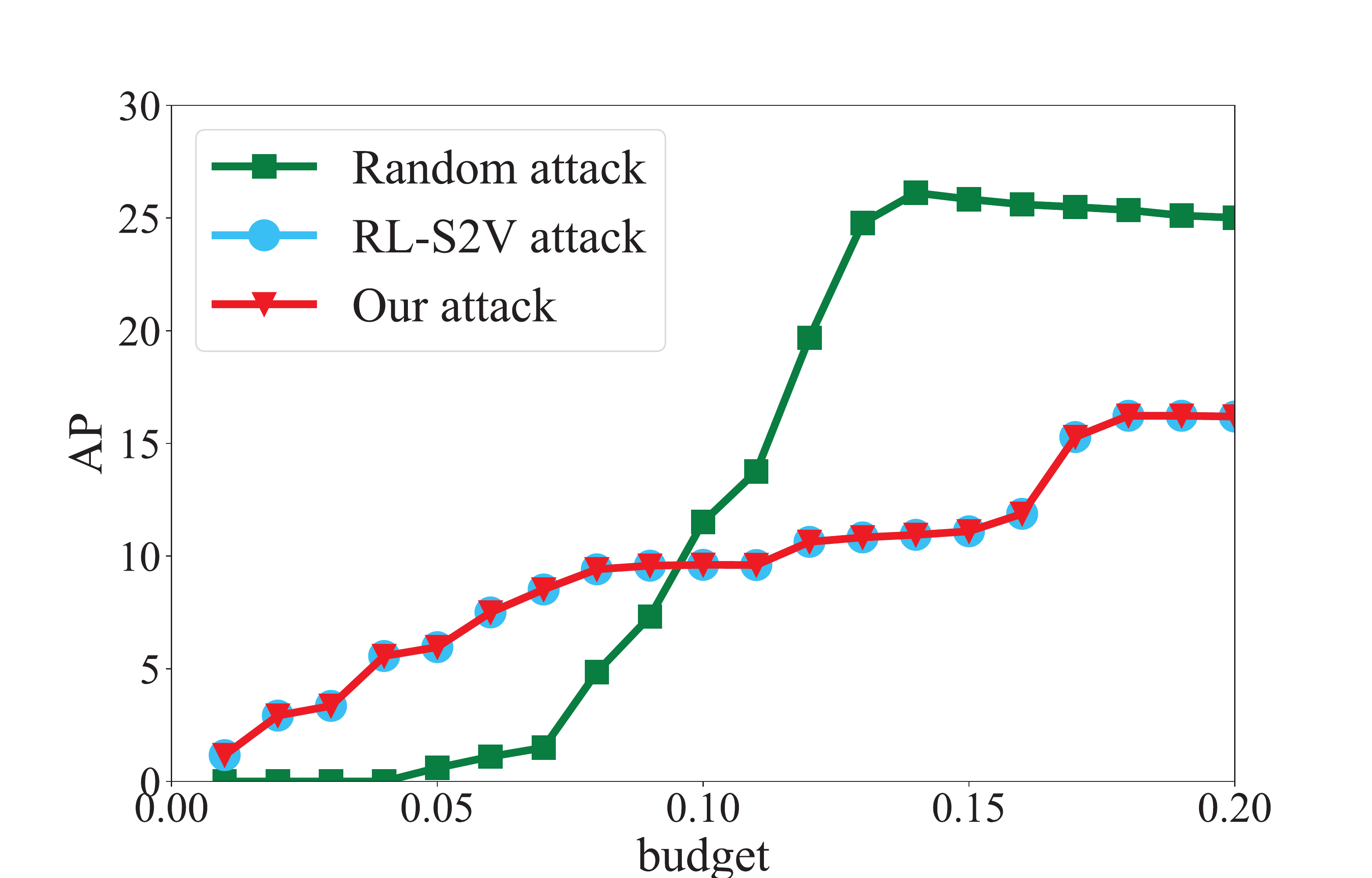}
\label{IMDB_AP}}
\vspace{-2mm}  
\caption{Average perturbation (AP) of our attack vs. budget $b$ on the three datasets against GIN.}
\label{fig:AP}
\vspace{-3mm}
\end{figure*}

\begin{figure*}[t]
\centering
\subfloat[NCI1:SAG]{\includegraphics[width=0.24\textwidth]{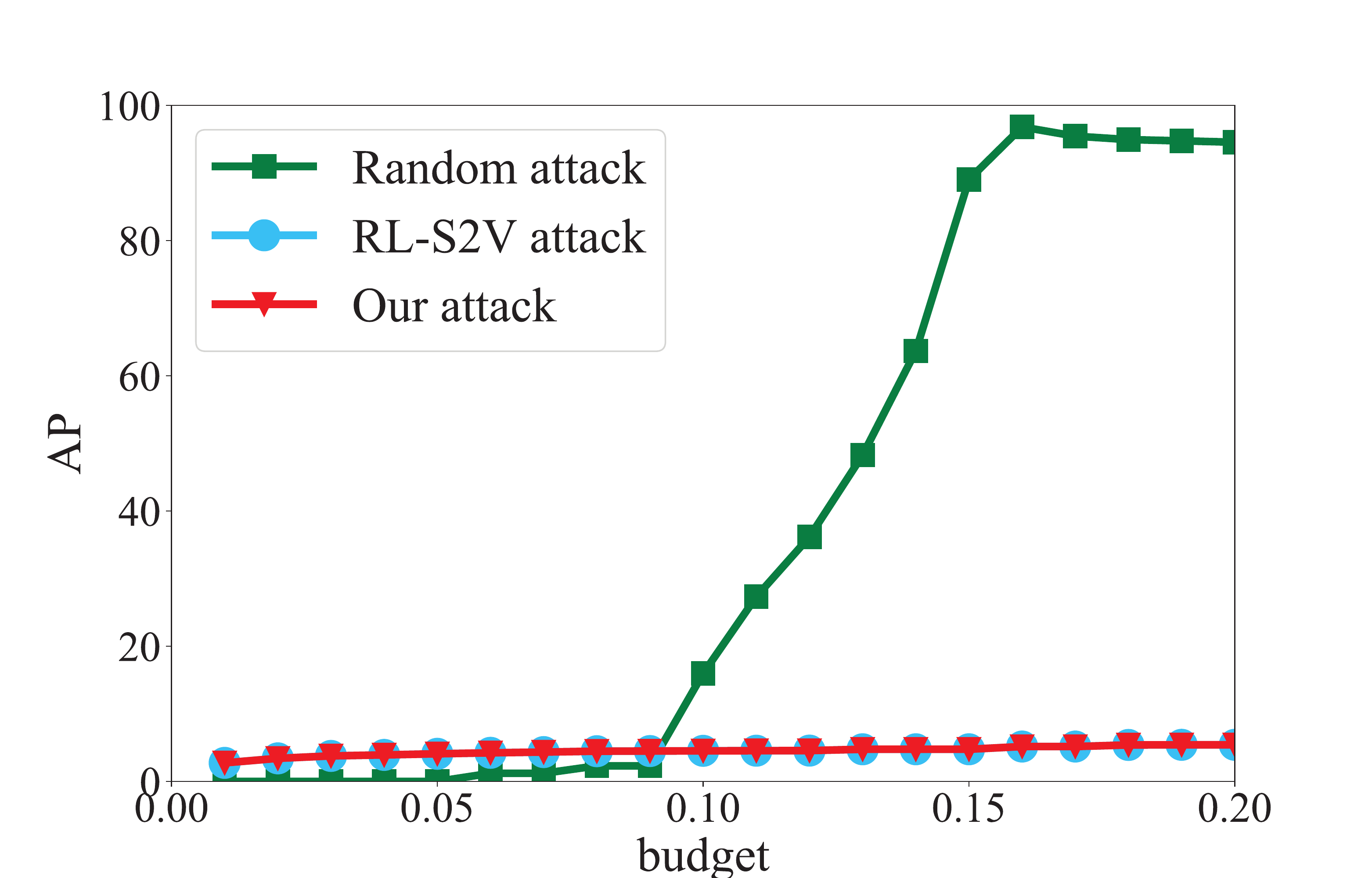}
\label{NCI1_AP_SAG}}
\subfloat[IMDB:SAG]{\includegraphics[width=0.24\textwidth]{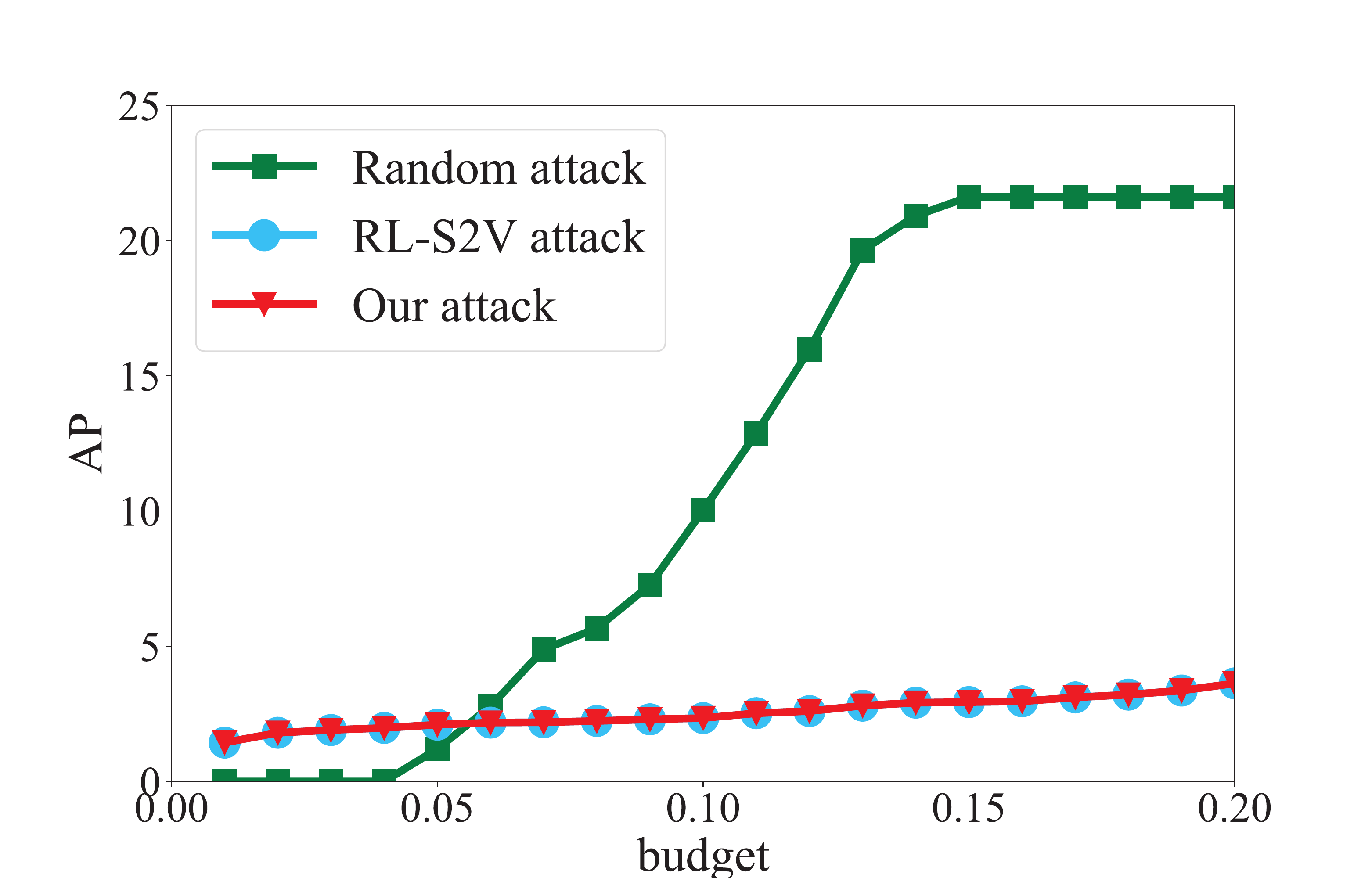}
\label{IMDB_AP_SAG}}
\vspace{-2mm}
\subfloat[NCI1:GUNet]{\includegraphics[width=0.24\textwidth]{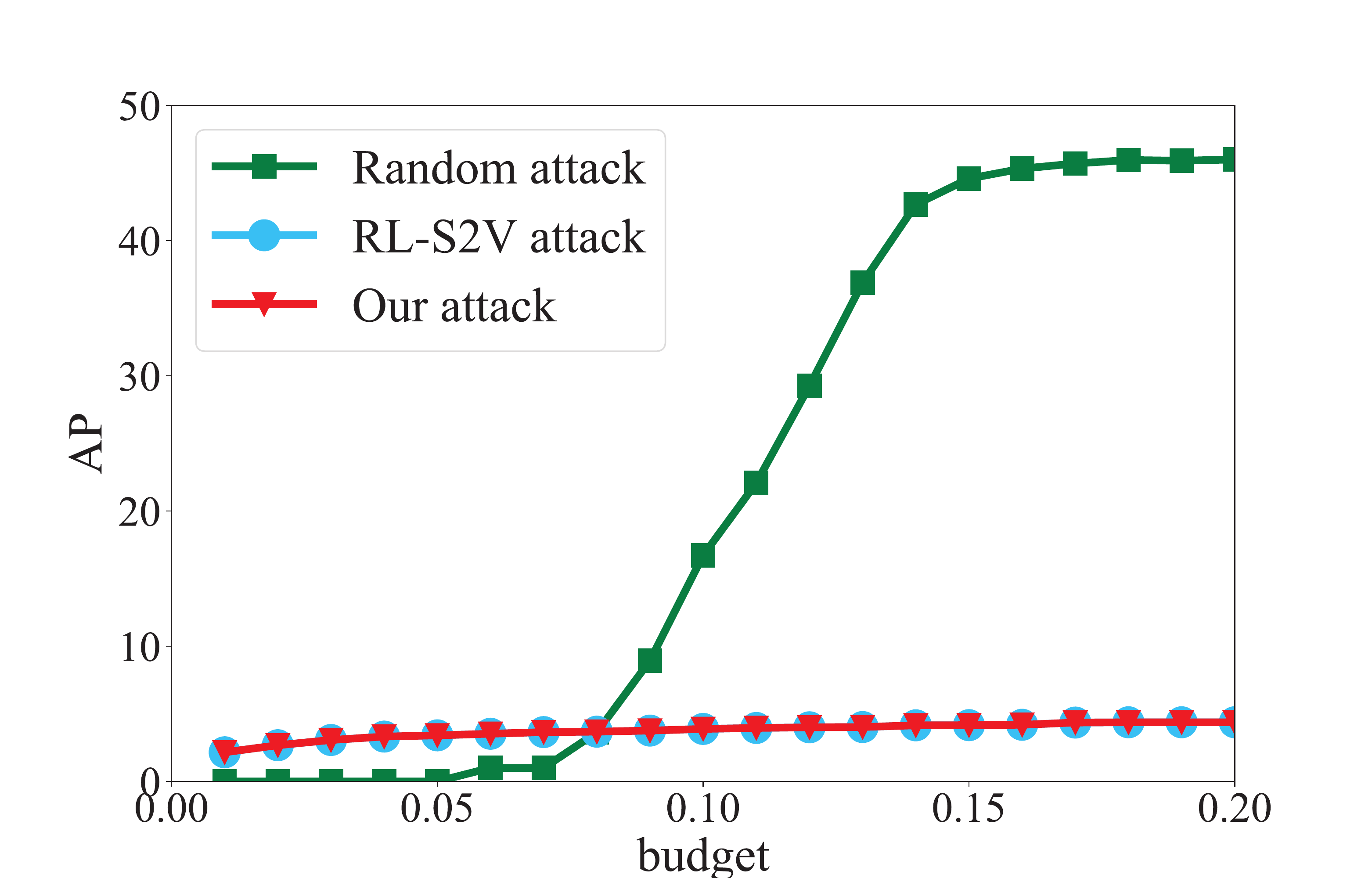}
\label{NCI1_AP_GUN}}
\subfloat[IMDB:GUNet]{\includegraphics[width=0.24\textwidth]{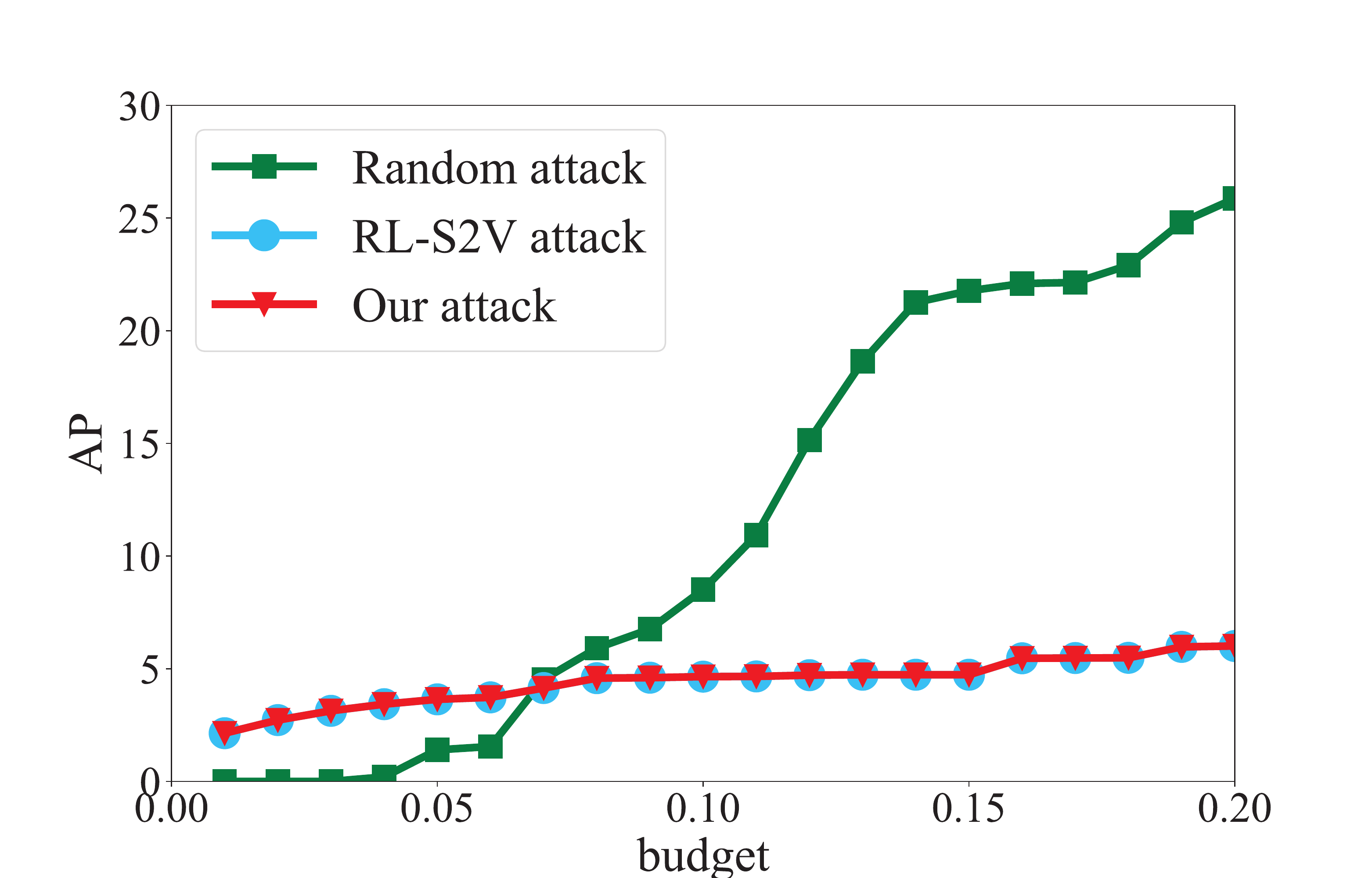}
\vspace{-2mm}   
\label{IMDB_AP_GUN}}
\caption{Average perturbation (AP) of our attack vs. budget $b$ on IMDB and NCI1 against SAG and GUNet.}
\label{fig:AP_SAG_GUNet}
\end{figure*}

\begin{itemize}[leftmargin=*]
\setlength{\itemsep}{0.1pt}
    \item {RL-S2V attack. RL-S2V is a reinforcement learning based adversarial attack that models the attack as a Finite Horizon Markov Decision Process. To attack each target graph, it first decomposes the action of choosing one perturbed edge in the target graph into two hierarchical actions of choosing two nodes separately. Then it uses Q-learning to learn the Markov decision process. 
    In the RL-S2V attack, the attacker needs to set a maximum number of perturbed edges before the attack. Thus, in our experiments,  we first conduct our attack to obtain the perturbation rate and then we set the perturbation rate of RL-S2V attack the same as ours. \M{Thus, the RL-S2V attack and our attack will have the same APs (see Figure \ref{fig:AP} and \ref{fig:AP_SAG_GUNet}).}
    For ease of comparison, we also tune RL-S2V to have a close number of queries as our attack. Then, we compare our attack with RL-S2V in terms of SR and AT. 
    }
    \item Random attack. 
    The attacker first  chooses a perturbation ratio uniformly at random. Then, given a target graph, the attacker randomly perturbs the corresponding number of edges in the target graph. 
For ease of comparison, the attacker will repeat this process and has the same number of queries as our attack, and choose the successful adversarial graph with a \emph{minimal} perturbation as the final adversarial graph. 
Note that, the random attack we consider is the strongest, as the attacker always chooses the successful adversarial graph with a minimal perturbation. 
\end{itemize}

\noindent \textbf{Parameter setting.} All the four metrics are impacted by the pre-set budget $b$. Unless otherwise mentioned, we set a default $b=0.2$. Note that we also study the impact of $b$ in our experiments. For other parameters such as $Q$ and $\mu$ in signSGD, we set $Q=100$ and $\mu=0.1$ by default. 
In each experiment, we repeat the trail 10 times and use the average results of these trails as the final results to ease the influence of randomness.



\begin{figure*}[!t]

\centering
\subfloat[NCI1:GIN]{\includegraphics[width=2in]{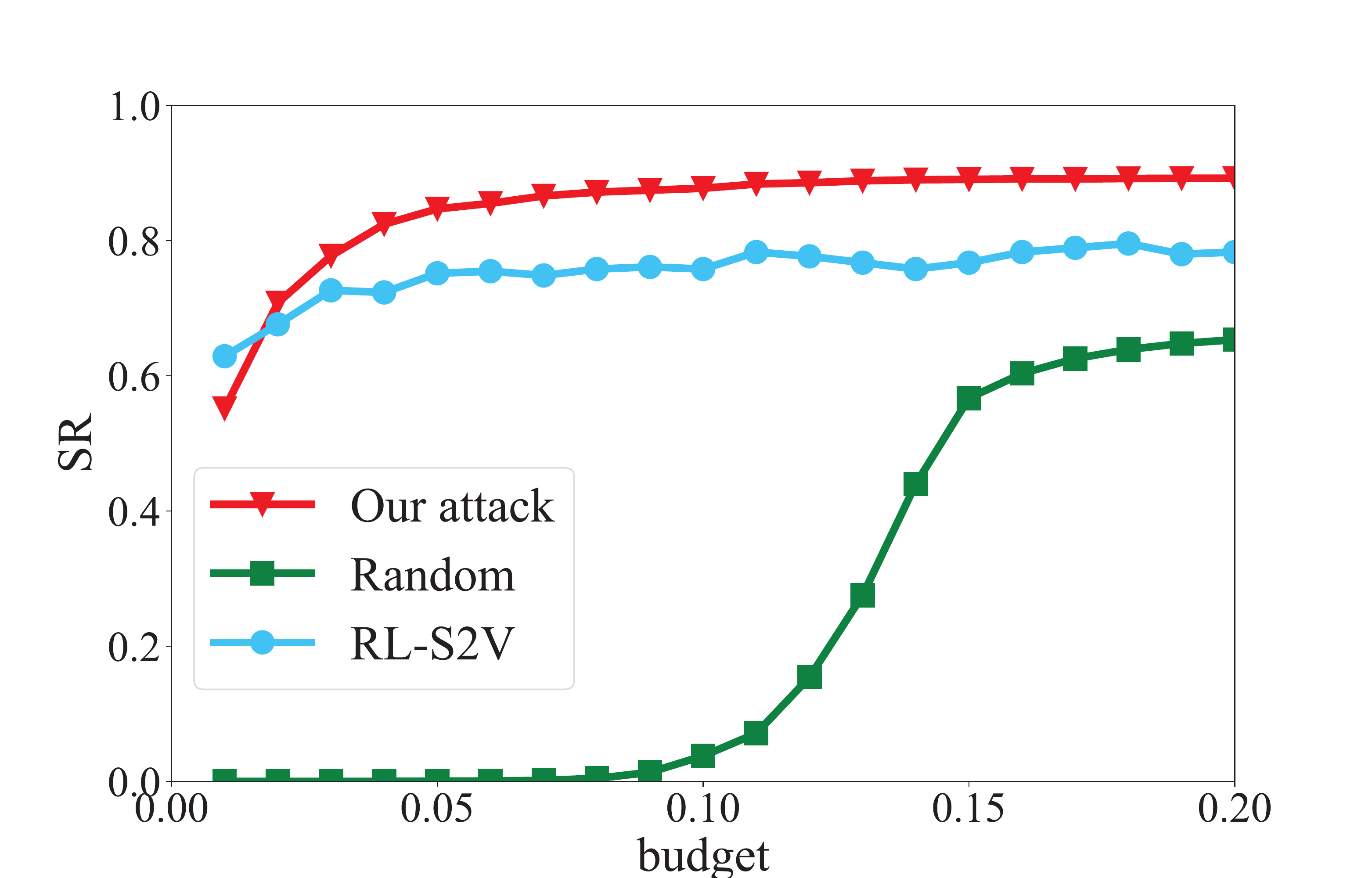}
\label{NCI1_SR}}
 \hfil
\subfloat[COIL:GIN]{\includegraphics[width=2in]{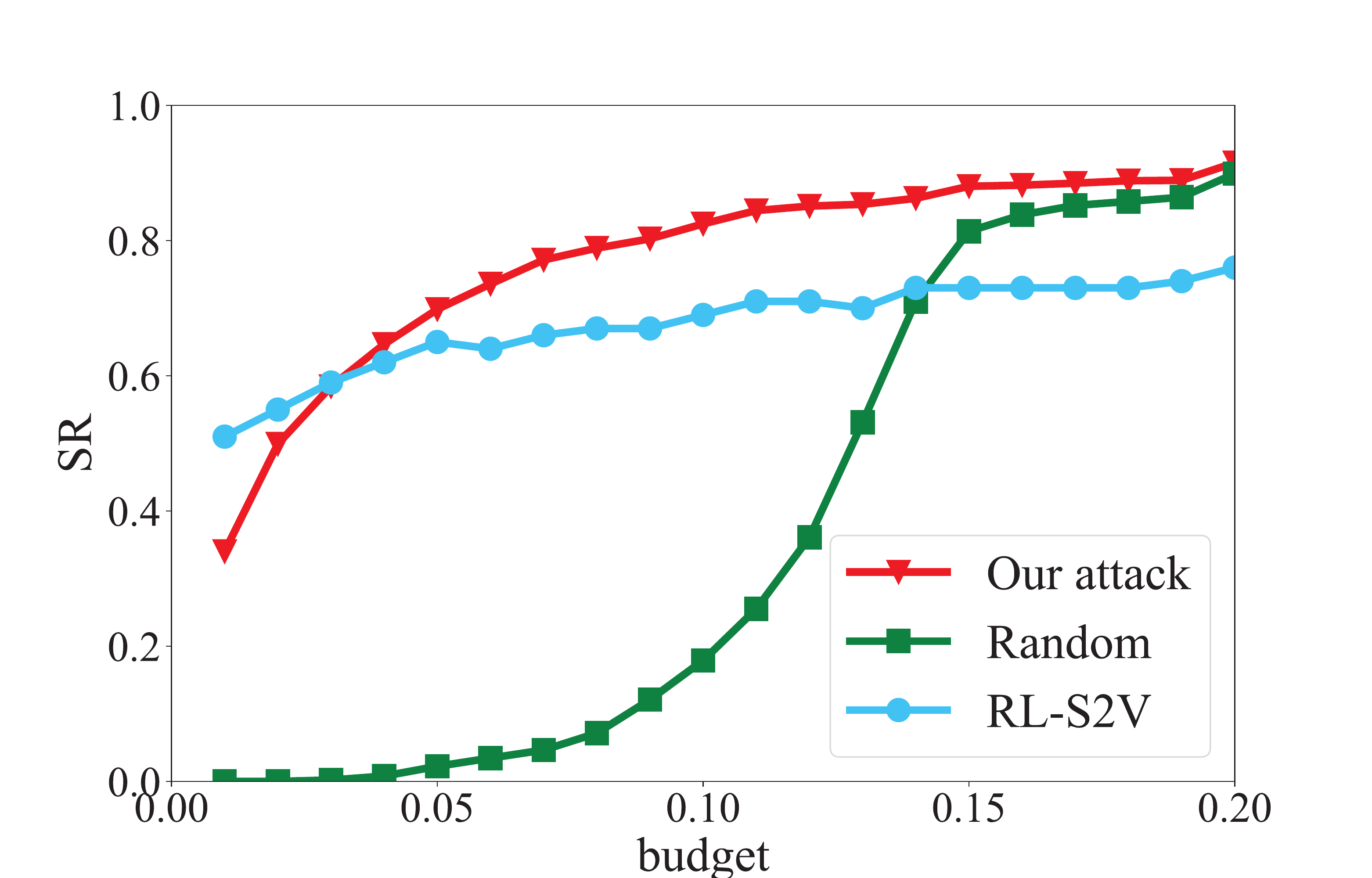}
\label{COIL_SR}}
 \hfil
\subfloat[IMDB:GIN]{\includegraphics[width=2in]{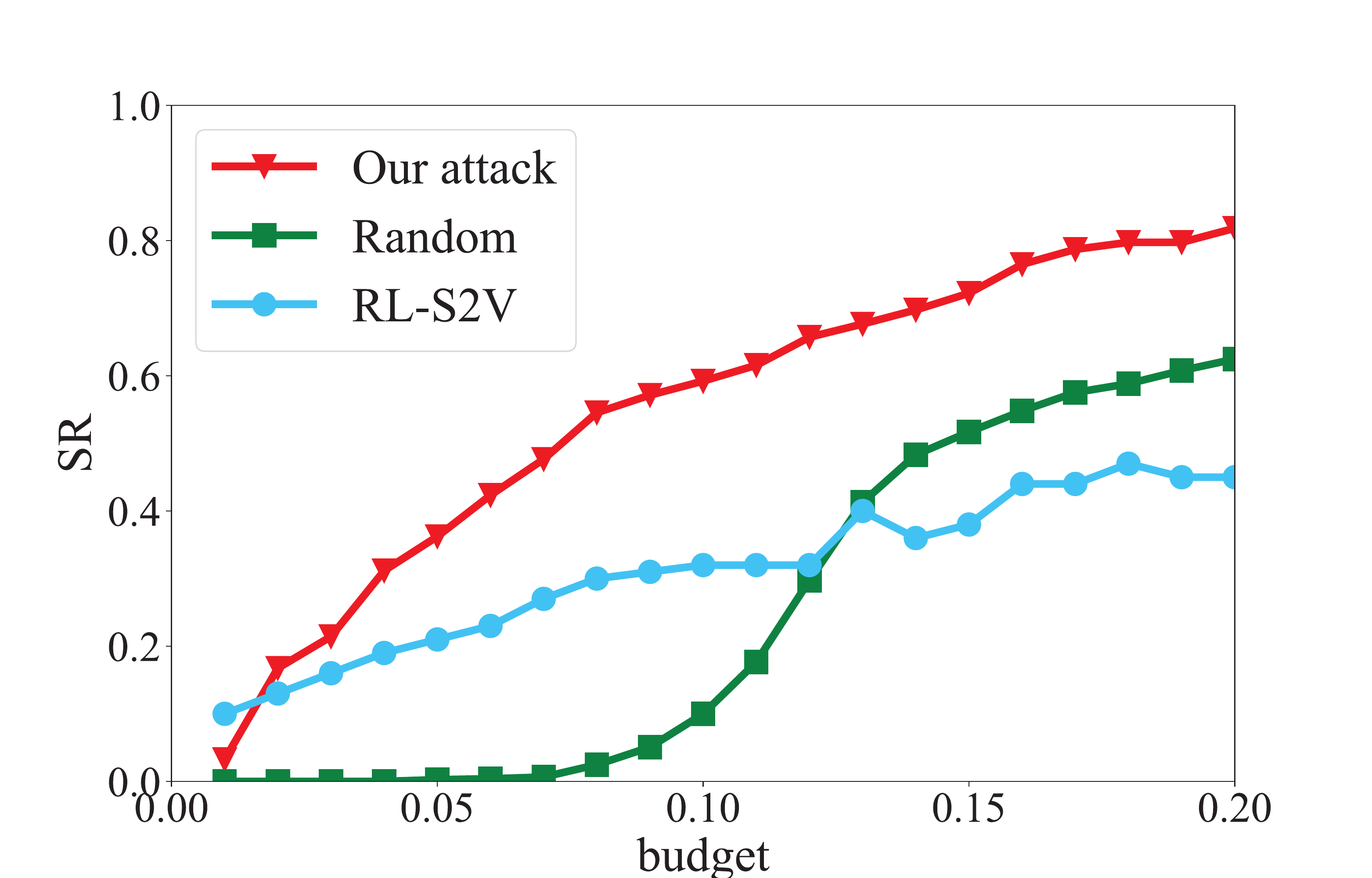}
\label{IMDB_SR}}
\vspace{-2mm}  
\caption{Successful rate (SR) of our attack vs. budget $b$ on the three datasets against GIN.}
\vspace{-4mm}
\label{fig:SR}
\end{figure*}

\begin{figure*}[t]

\centering
\subfloat[NCI1:SAG]{\includegraphics[width=0.24\textwidth]{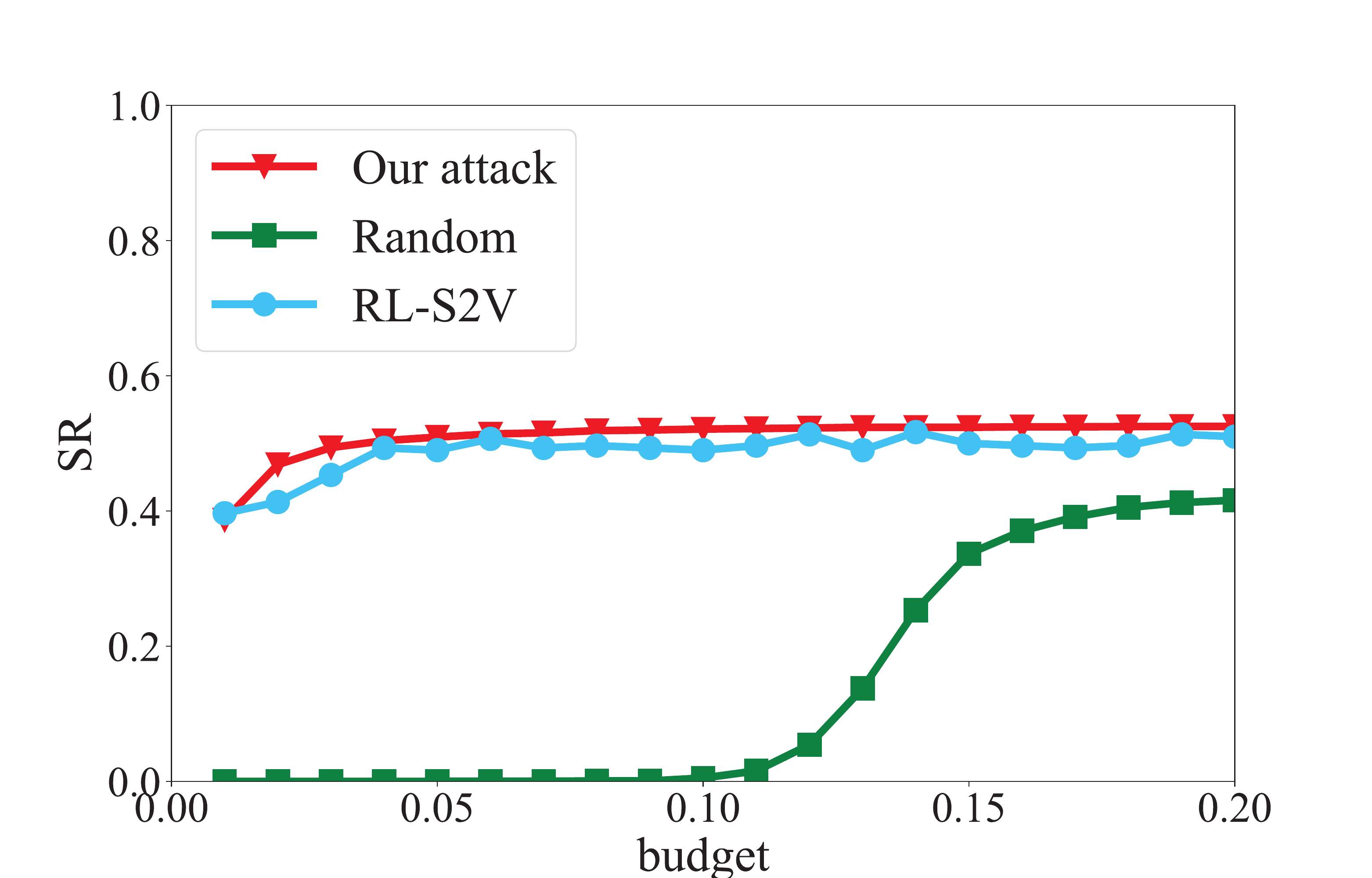}
\label{NCI1_SR_SAG}}
\subfloat[IMDB:SAG]{\includegraphics[width=0.24\textwidth]{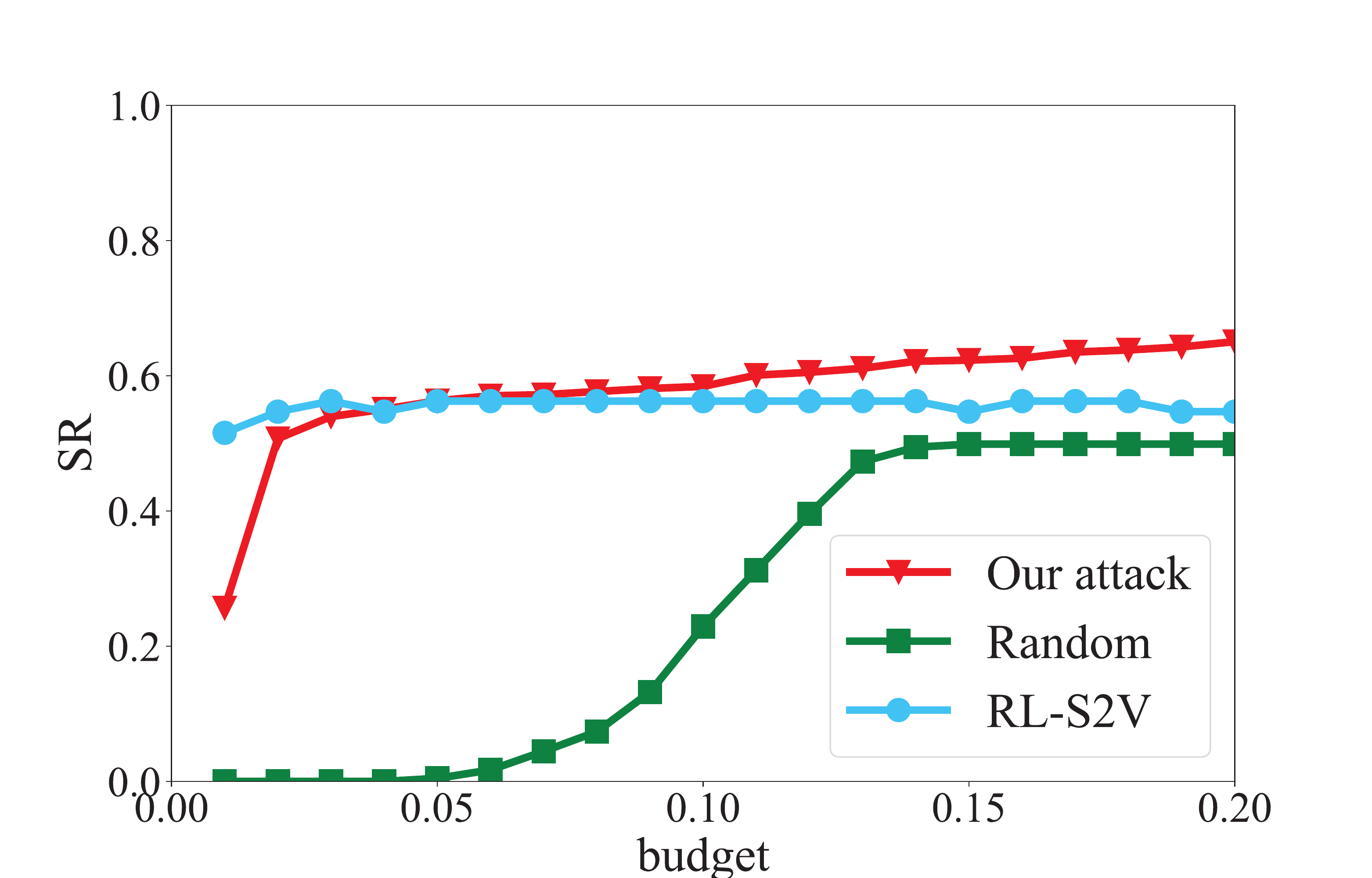}
\label{IMDB_SR_SAG}}
\subfloat[NCI1:GUNet]{\includegraphics[width=0.24\textwidth]{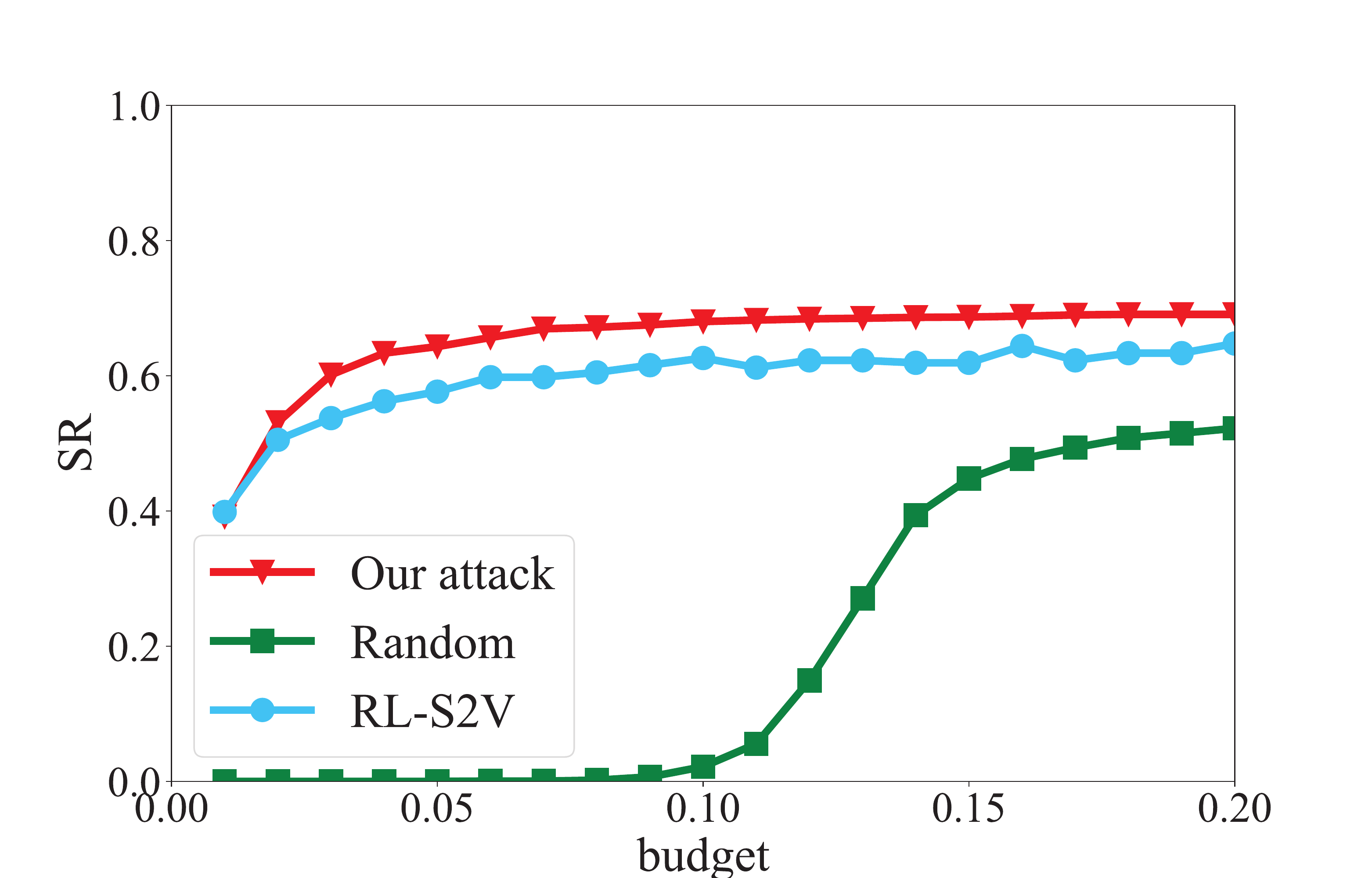}
\label{NCI1_SR_GUN}}
\subfloat[IMDB:GUNet]{\includegraphics[width=0.24\textwidth]{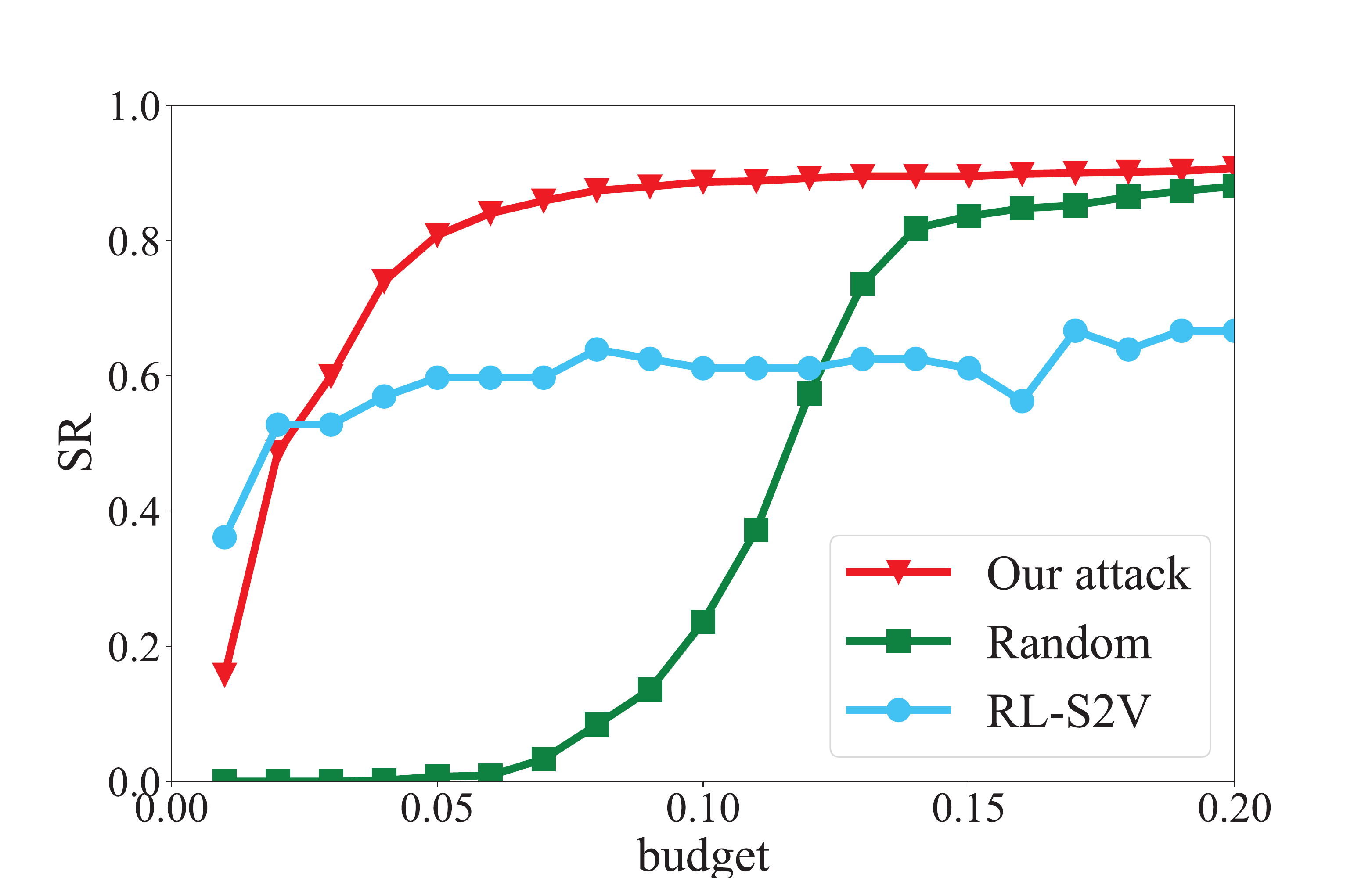}
\label{IMDB_SR_GUN}}
\vspace{-2mm}  
\caption{Successful rate (SR) of our attack vs. budget $b$ on  IMDB and NCI1 against SAG and GUNet.}
\vspace{-8mm}
\label{fig:SR_SAG_GUNet}
\end{figure*}



 \begin{table}[t]
\centering
  \fontsize{8}{11}\selectfont
  \caption{AQ and AT on three datasets.} 
  \vspace{-3mm}
    \begin{tabular}{c|c|ccc}
    \hline
    \multirow{1}{*}{Dataset}&
    Metric&Our&RL-S2V&Random \\
    \hline
    \hline
    \multirow{2}{*}{COIL}&AQ&1621&1728&1621\cr
    &AT (s)&121&245&97\cr
 
    \cline{1-5}
    \multirow{2}{*}{IMDB}&
    AQ&1800&1740&1800\cr
    &AT (s)&109&7291&88\cr
   
    \cline{1-5}
    \multirow{2}{*}{NCI1}&
    AQ&1822&1809&1822\cr
    &AT (s)&163&3071&104\cr
  
    \cline{1-5}
    \hline
    \hline
    \end{tabular}
    \vspace{-4mm}
\label{tab:AQ and AT}
\end{table}

\subsection{Effectiveness of Our Attack}
\label{sec:experiment results}
We conduct experiments to {evaluate} our hard-label black-box attacks. Specifically, we study the impact of the attack budget, the impact of our coarse-grained searching algorithm, and the impact of query-efficient gradient computation. 


\subsubsection{Impact of the budget on the attack}
Figure \ref{fig:SR} and ~\ref{fig:SR_SAG_GUNet} show the SR of the compared attacks with different budgets on the three datasets and three GNN models. We sample 20 different 
budgets ranging from 0.01 to 0.20 with a step of 0.01. We can observe that: (i) Our attack outperforms the baseline attacks significantly in {most} cases. For instance, with a budget $b$ less than 0.05, random attack fails to work on the three datasets, while our attack achieves a SR at least 40\%; {With a budget $b=0.15$, our attack against GIN achieves a SR of $72\%$ on IMDB, while the SR of RL-S2V is less than $40\%$.}
The results show that our proposed optimization-based attack is far more advantageous than the {baseline methods}. 
(ii) All methods have a higher SR with a larger budget. 
This is because a larger budget allows an attacker to perturb more edges in a graph. 

We further calculate AP of successful adversarial graphs with different budgets $b$, and show the results in Figure \ref{fig:AP} and  \ref{fig:AP_SAG_GUNet}. 
Note that, due to algorithmic issue, RL-S2V is set to have the same AP as our attack. 
We have several observations. (i) The AP of our attack is smaller for achieving a higher SR, which shows that our attack outperforms random attack significantly, even when the considered random attack is the strongest. For example, on the COIL dataset, our attack can achieve a SR of $91.52\%$ when $b=0.20$ and the corresponding AP of adversarial graphs is $4.33$. 
Under the same setting, random attack has a SR of only $9.25\%$. 
(ii) AP increases with budget $b$. It is obvious and reasonable because a larger budget means that the perturbed graphs with large perturbations have larger probabilities to generate successful adversarial graphs. 
(iii) The APs of our attack on three datasets are different. 
The reason is that these datasets have different average degree.
Specifically, IMDB is the most dense graph while NCI1 is the least dense. 
This result demonstrates that it takes more effort 
to change the state of supernodes or superlinks of graphs in the dense graph, and thus we need to perturb more edges. 

\M{To evaluate the types of perturbations, we record the number of added edges and removed edges for each dataset in our attack. With the target GNN model as GIN and $b=0.20$, the averaged (added edges, removed edges) on IMDB, COIL, and NCI1 are (8.46, 12.03), (2.51, 1.80), and (12.84, 1.12), respectively. Thus, we can see that we should remove more edges for denser datasets (e.g., IMDB) and add more edges for sparser datasets (e.g., COIL and NCI1). }

We also record AQ and AT of the three attack methods on the three datasets, as shown in Table \ref{tab:AQ and AT}. Recall that the three methods are set to have very close number of queries. We observe that RL-S2V has far more AT than our attack and random attack. This is because the searching space of RL-S2V is exponential to the number of nodes of the target graph. Random attack has the smallest AT, as it does not need to compute gradients. Our attack has similar AT as random attack, although it needs to compute gradients. 


\newcommand{\tabincell}[2]{\begin{tabular}{@{}#1@{}}#2\end{tabular}}
\begin{table}[!t]
\centering
  \fontsize{8}{11}\selectfont
  \caption{Coarse-grained searching with different strategies.}
  \vspace{-3mm}  
  \label{tab:coarsed level searching}
    \begin{tabular}{c|c|cccc}
    \hline
    \multirow{1}{*}{Dataset}&
    Strategy &SR&AP&AQ&AT (s) \\
    \hline
    \hline
    \multirow{3}{*}{COIL}&I&{\bf 0.89}&{\bf 8.88}&	{\bf 175}&{\bf 3.07}\cr
    &II&0.86&9.15&337&7.60\cr
    &III&0.84&14.46&339&29.29\cr
    \cline{1-6}
    \multirow{3}{*}{IMDB}&I&{\bf 0.79}&{17.27}&	293&{\bf6.46}\cr
    &II&{\bf 0.79}	&{\bf 17.22}&{\bf279}&6.66\cr
    &III&0.57&17.62&	308&	18.80\cr
     \cline{1-6}
    \multirow{3}{*}{NCI1}&I&0.88&{\bf12.57}&{\bf437}&{\bf7.55}\cr
    &II&{\bf0.89}&13.42&725&	12.55\cr
    &III&0.59&43.09&463&49.87\cr
     \cline{1-6}
    \hline
    \hline 
    \end{tabular}
\end{table}

\subsubsection{Impact of coarse-grained searching (CGS) on the attack}
In this experiment, we evaluate the impact of different strategies of CGS on the effectiveness of the attack. 
{Specifically, we will validate the importance of initial search in our entire attack. 
We use three methods to search the initial perturbation vector $\Theta_0$: 
(i) Strategy-I (i.e., our strategy):} supernode + superlink + whole graph, which means we search the space in the order of 
supernodes, superlinks and the whole graph (see Section~\ref{sec:initial search}); 
(ii) Strategy-II: superlink + supernode + whole graph; 
and (iii) Strategy-III: whole graph, which means we do not use CGS and search the whole space defined by the target graph directly. 
{Note that, this strategy also means that we start our signSGD based on a randomly chosen $\Theta_0$.}

Table \ref{tab:coarsed level searching} shows the attack results with different strategies against GIN. 
We have the following observations. 
(i) The SRs of strategy-I/-II are very close and both are much higher than that of strategy-III. For example, the SRs of strategy-I and -II  are 0.88 and 0.89 on NCI1, while that of strategy-III is 0.59.
(ii) The APs of strategy-I/-II are much less than \M{that} of strategy-III. For instance, AP of strategy-I on the NCI1 dataset is only 12.57, while that of strategy-III is 43.09, about 3.43 times more than the former. This result validates that CGS can find better initial vectors with less perturbations. 
(iii) Strategy-I has the least searching time and the least number of queries among the three strategies. For instance, it only requires 3.07 seconds to find $\Theta_0$ for target graphs  on  COIL, while Strategy-II requires 2x time. 
\M{(iv) The benefit of our CGS algorithm (e.g.,  Strategy III has a 1.94x AQ and 9.54x AT of our Strategy I on COIL) does not reach the theoretical level as stated in Theorem \ref{theorem3} (i.e., O($2^{\kappa^4}$)). The reason is that from a practical perspective, we assume that the attacker only has maximum number of queries as $5N$, which is exponentially much less than $2^S$. If we traverse the entire graph space with strategy-III as stated in Theorem \ref{theorem3}, AQ and AT will be enlarged to $\frac{2^S}{5N}$ 
times, which also explains the gap between strategy-I/II and strategy-III.}
In summary, 
our proposed CGS algorithm can effectively find initial perturbation vectors with higher success rates, less perturbations, less queries, and shorter time.

\begin{figure}[t]
\centering
\includegraphics[width=2.8in]{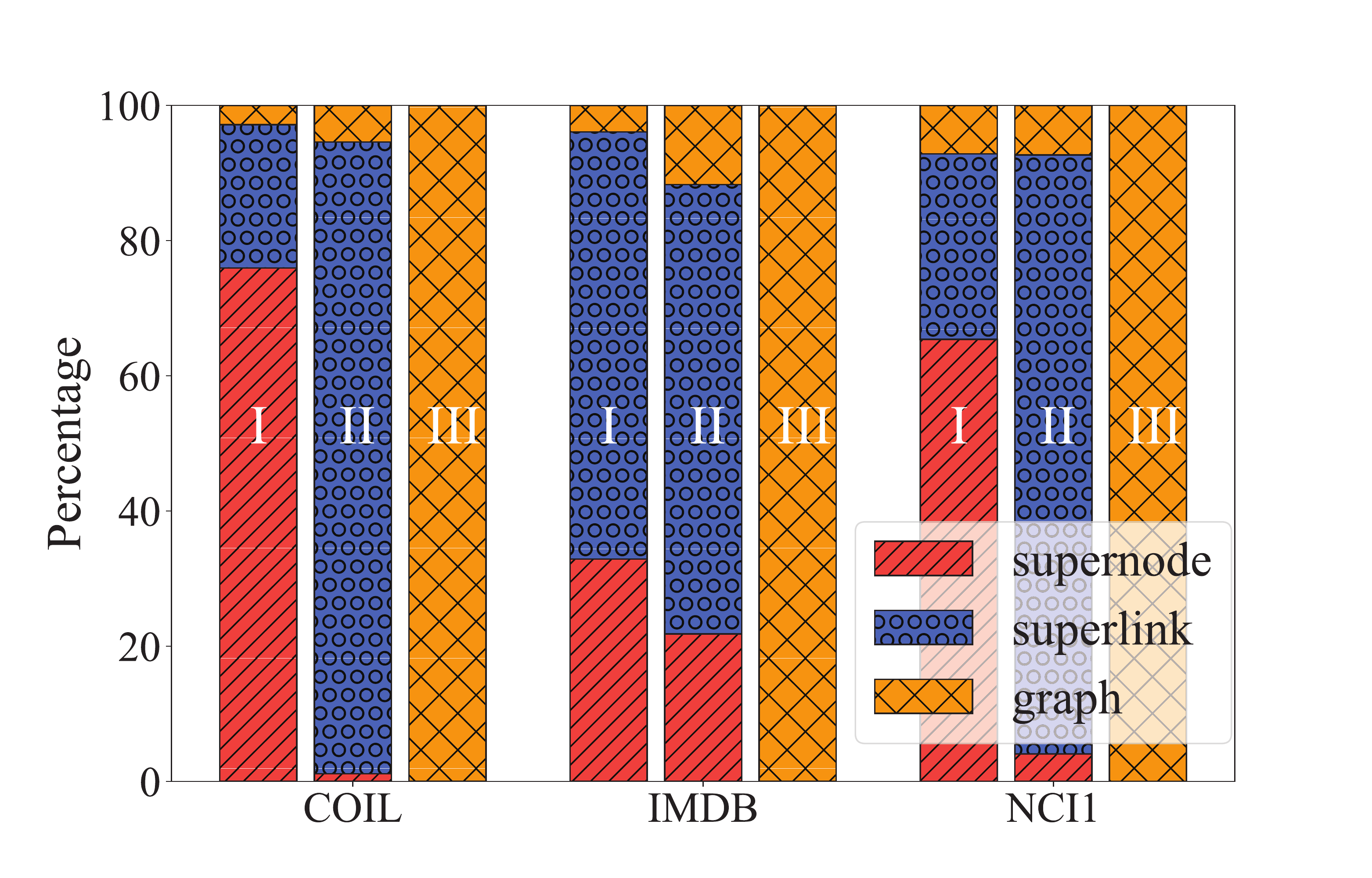}
\vspace{-4mm}
\caption{Percentage of the number of adversarial graphs with different initial perturbation vectors found in each component (i.e., supernode, superlink, and the whole graph) under three searching strategies.}
 \hfil
 \label{fig:percentage}
 \vspace{-4mm}
\end{figure}

We further analyze 
the percentages of adversarial graphs whose initial perturbation vectors $\Theta_0$ are found in each component (i.e., supernode, superlink, and graph).
Figure \ref{fig:percentage} shows the results. Each bar illustrates the percentages of adversarial graphs whose $\Theta_0$ is found by the three components. 
For instance, on COIL, 
we obtain $75.73\%$ adversarial graphs whose initial $\Theta_0$ 
is found in searching supernodes using strategy-I.
On COIL and NCI1, we find effective initial vectors $\Theta_0$ by using either strategy-I or strategy-II, i.e., either searching supernodes or superlinks first. 
Since the searching spaces of supernodes are often smaller than those of superlinks, strategy-I that searches within supernodes first is more suitable on these two datasets. Thus, strategy-I performs best among three strategies on these two datasets. However, on IMDB, we find $\Theta_0$ for most target graphs within the superlinks in both strategy-I/-II. Thus, strategy-II that searches within superlinks first is a better strategy for initial search on IMDB. From Table~\ref{tab:coarsed level searching}, we can also see that strategy-II performs slightly better than strategy-I in terms of AP and AQ. \M{Note that, the best searching strategies for different datasets are different. The possible reason is, due to the density of the datasets, i.e., strategy-II (i.e., searching superlinks first) is the best strategy for dense graphs (e.g., IMDB), while strategy-I (i.e, searching supernodes first) is for sparse graphs (e.g., NCI1), 
it is much harder to partition the graphs into supernodes in denser graphs than in sparser graphs.}

 \begin{table}[t]
\centering
  \fontsize{8}{11}\selectfont
  \caption{Impact of query-efficient  gradient  computation.} 
  \vspace{-3mm}  
  \label{tab:query efficient}
    \begin{tabular}{c|c|cccc}
    \hline
    \multirow{1}{*}{Dataset}
    &QEGC&SR&AP&AQ&AT (s) \cr
    \hline
    \hline
    \multirow{2}{*}{COIL}&Yes&{\bf0.92}&{\bf4.33}&{\bf1,622}&{\bf121.21}\cr
    &No&{\bf0.92}&4.44&9,859&808.76\cr
 
    \cline{1-6}
    \multirow{2}{*}{IMDB}&
    Yes&{\bf0.82}&16.19&{\bf1,800}&{\bf109.91}\cr
    &No&{\bf0.82}&{\bf16.05}&12,943&787.67\cr
   
    \cline{1-6}
    \multirow{2}{*}{NCI1}&
    Yes&{\bf0.89}&{\bf7.16}&{\bf1,822}&{\bf163.83}\cr
    &No&{\bf0.89}&7.60&10,305&1071.17\cr
    \cline{1-6}
    \hline
    \hline
    \end{tabular}
    \vspace{-4mm}
\end{table}

\begin{figure}[t]
\centering
\includegraphics[width=2.3in]{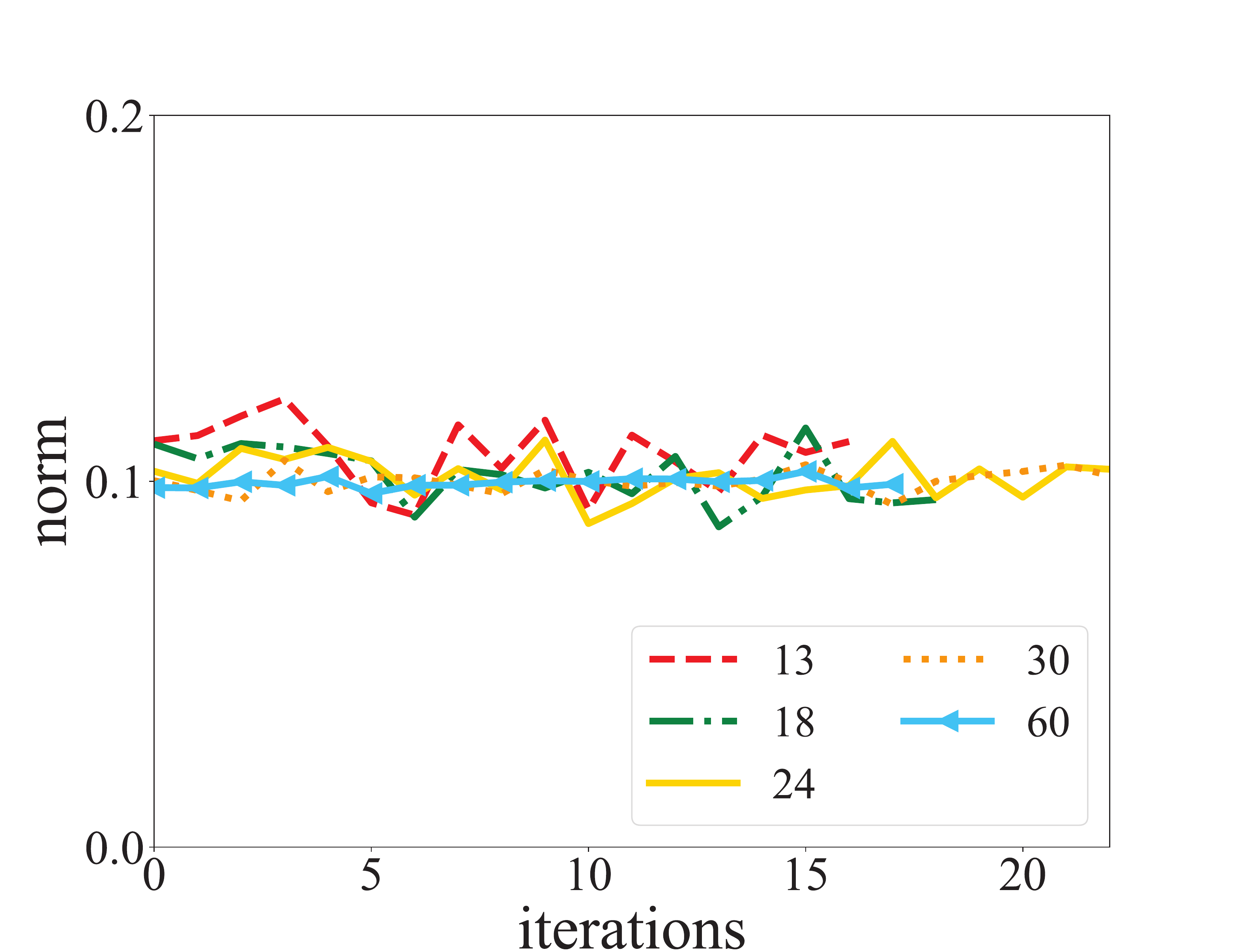}
\vspace{-2mm}
\caption{Gradient norms on the IMDB-BINARY dataset.}
\vspace{-4mm}
\label{fig:norms}
\end{figure}

 
   
  

\subsubsection{Impact of query-efficient gradient computation (QEGC)}
We further conduct experiments to evaluate the impact of QEGC. 
Table \ref{tab:query efficient} shows the results. 
We can observe that, under our attack, the number of required queries varies significantly, with and without QEGC. For instance, on IMDB, $AQ=12,943$ when we do not apply QEGC, while the $AQ$ is reduced to 1,800 when using QEGC, which is only $13.91\%$ of the former. The SR and AP vary slightly with and without QEGC. For example, the APs with and without QEGC on COIL are $4.33$ and $4.44$, respectively, and the difference is only $0.11$. These results demonstrate that QEGC can significantly reduce the number of queries and thus the attack time in our attacks, while maintaining high success rate and incurring small perturbations.


 
 \subsubsection{Gradient norms in our attack.}
 {The convergence property of our optimization based hard label black-box attack is based on Assumption \ref{assump:2}, which requires that the norm of gradient of $p(\Theta)$ should be bounded. 
 Here, we conduct an experiment to verify whether this assumption is satisfied. 
 Specifically, 
 we randomly choose 5 target graphs from IMDB that are successfully attacked by \M{our} attack. The number of nodes of these graphs are 13, 18, 24, 30, and 60, respectively. From  Figure \ref{fig:norms}, 
 we can observe that gradients norms are relatively stable
and are around 0.1 in all cases. 
 Therefore, Assumption \ref{assump:2} is satisfied in our attacks. }



\begin{figure*}[t]
\centering
\subfloat[NCI1]{\includegraphics[width=2.1in]{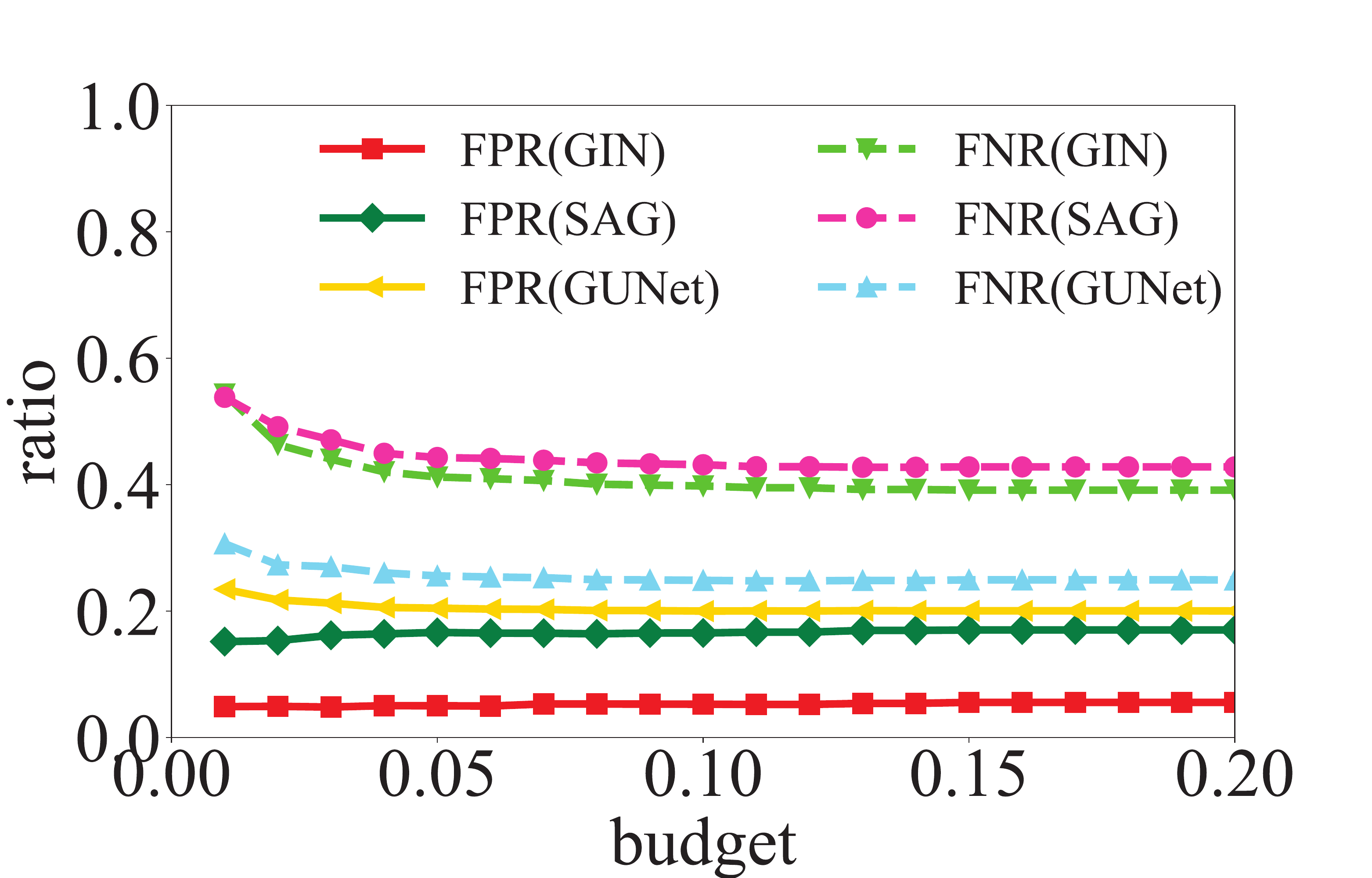}
\label{NCI1_our}}
 \hfil
\subfloat[COIL]{\includegraphics[width=2.1in]{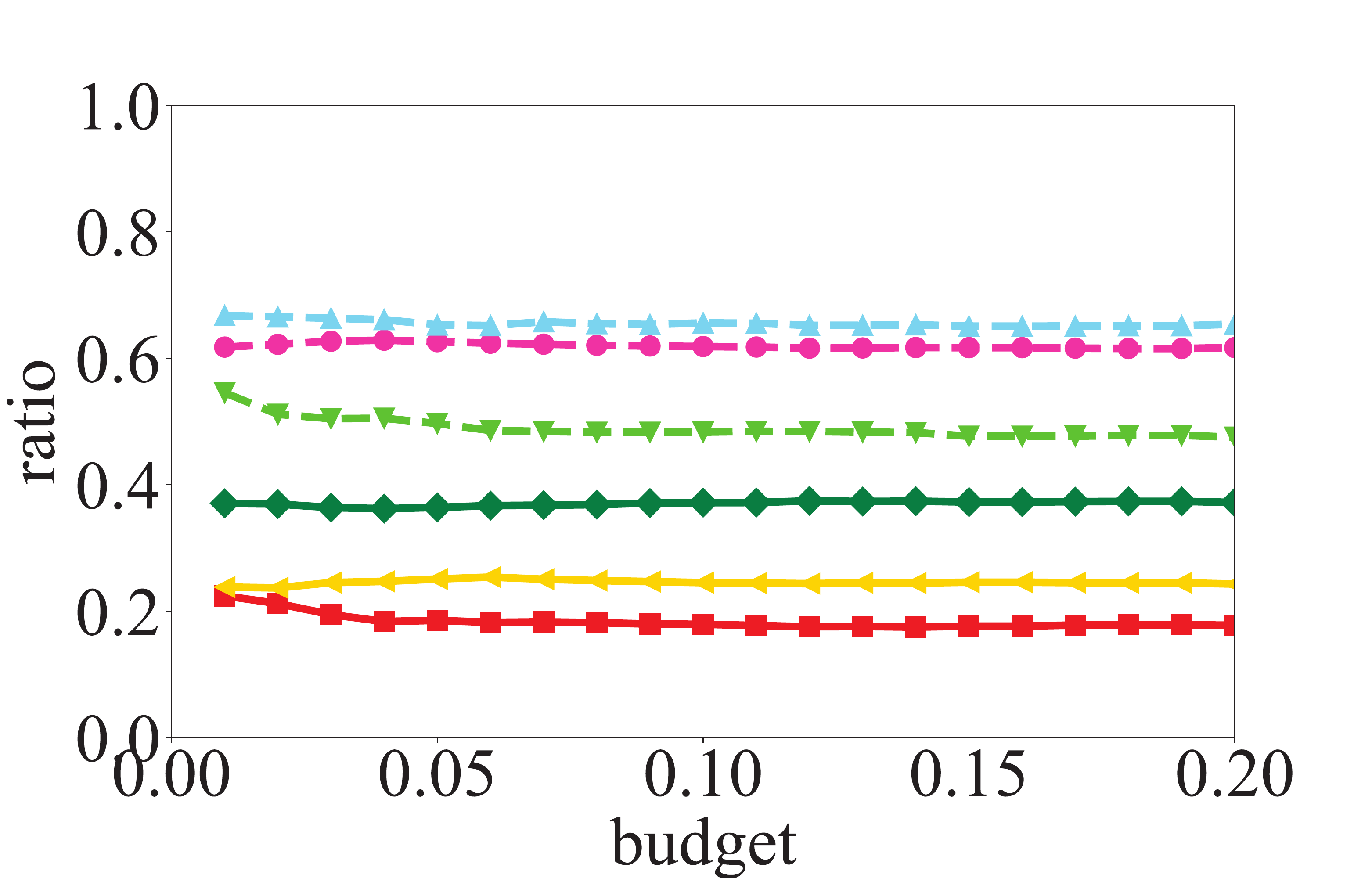}
\label{COIL_our}}
 \hfil
\subfloat[IMDB]{\includegraphics[width=2.1in]{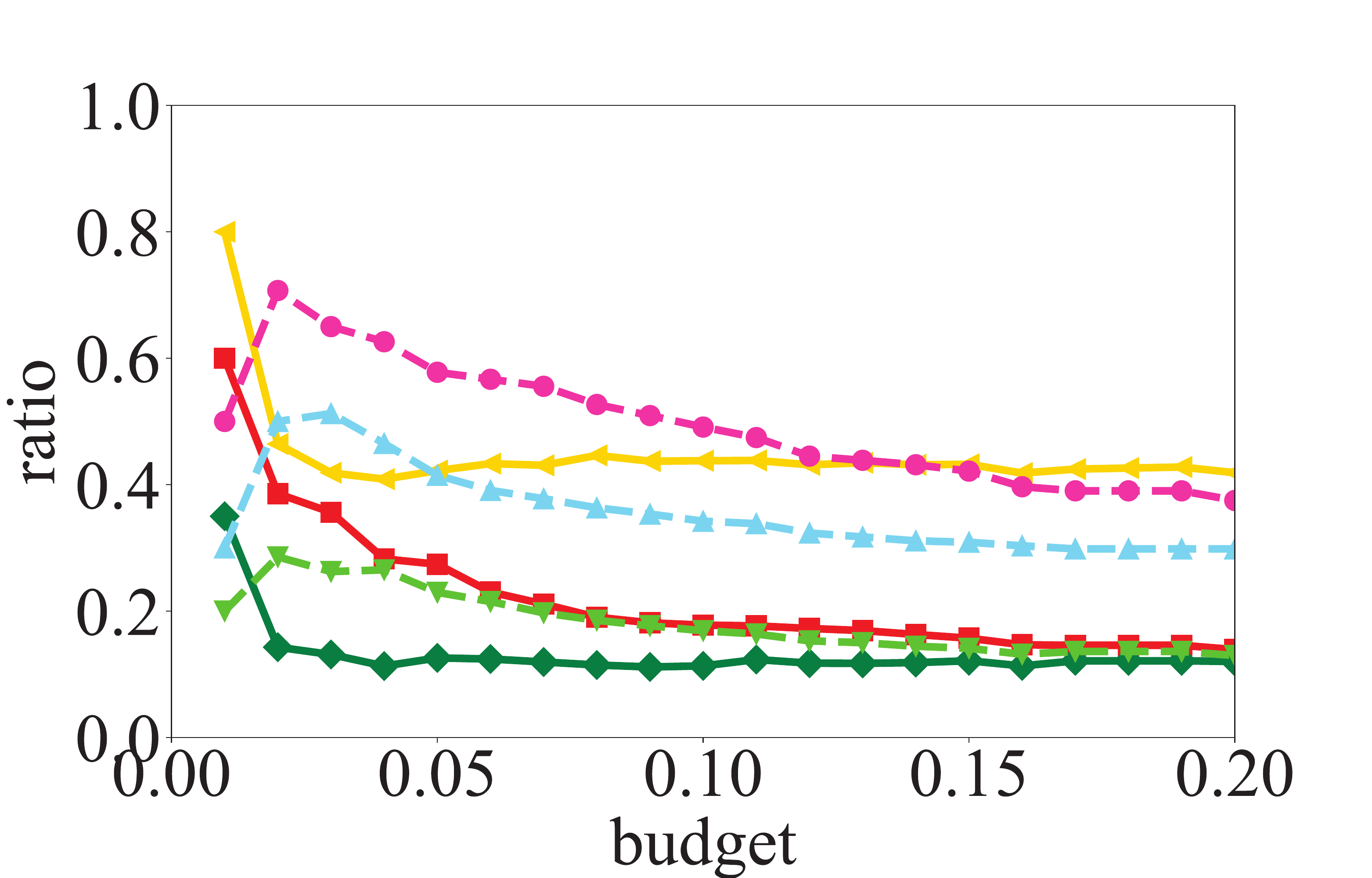}
\label{IMDB_our}}
\vspace{-4mm} 
\caption{Detection performance vs. budget $b$ on the testing dataset with the training dataset generated by our attack.}
\label{fig:detect our}
\vspace{-6mm}
\end{figure*}

\begin{figure*}[t]
\centering
\subfloat[NCI1]{\includegraphics[width=2.1in]{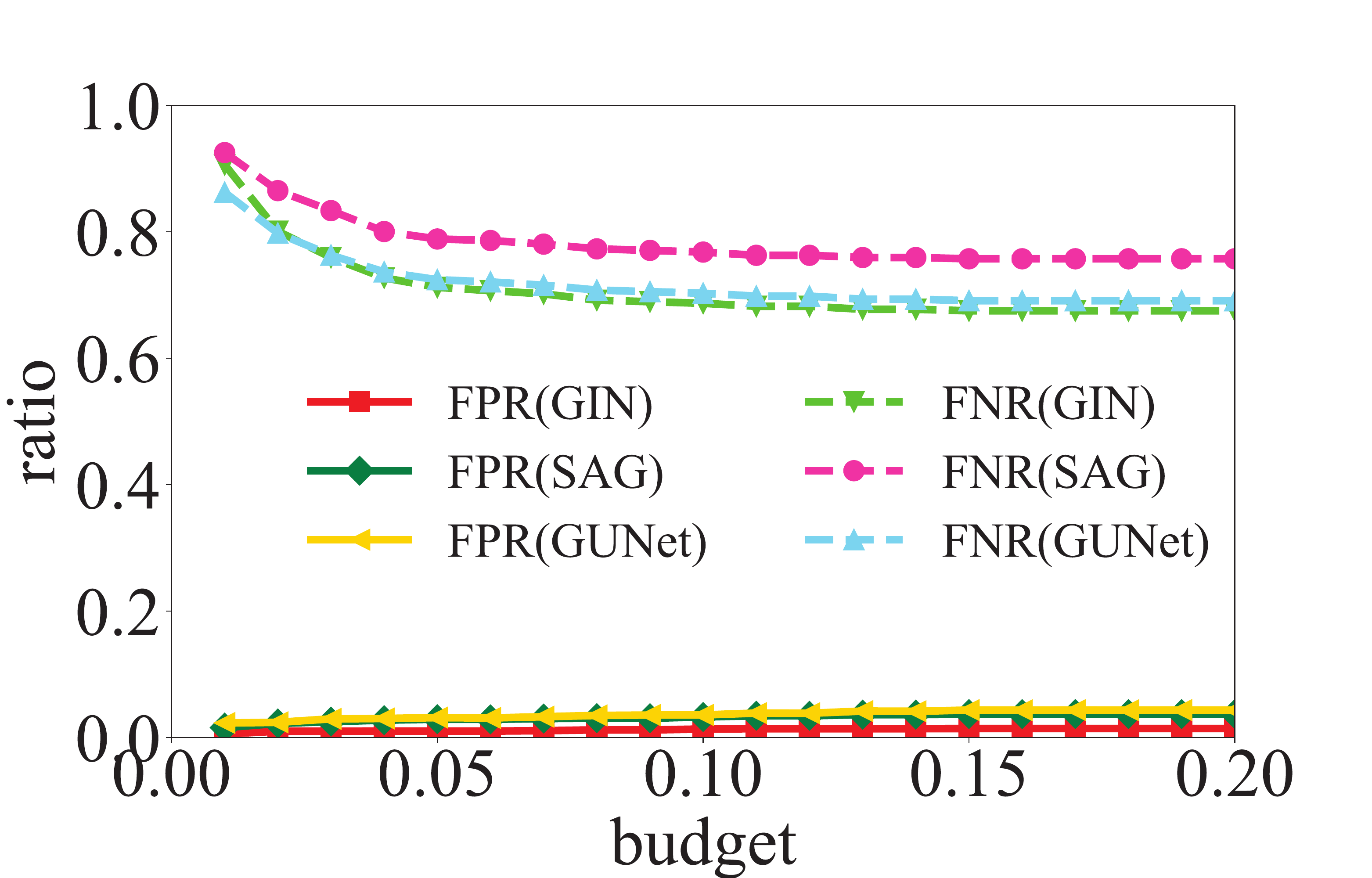}
\label{NCI1_typo}}
 \hfil
\subfloat[COIL]{\includegraphics[width=2.1in]{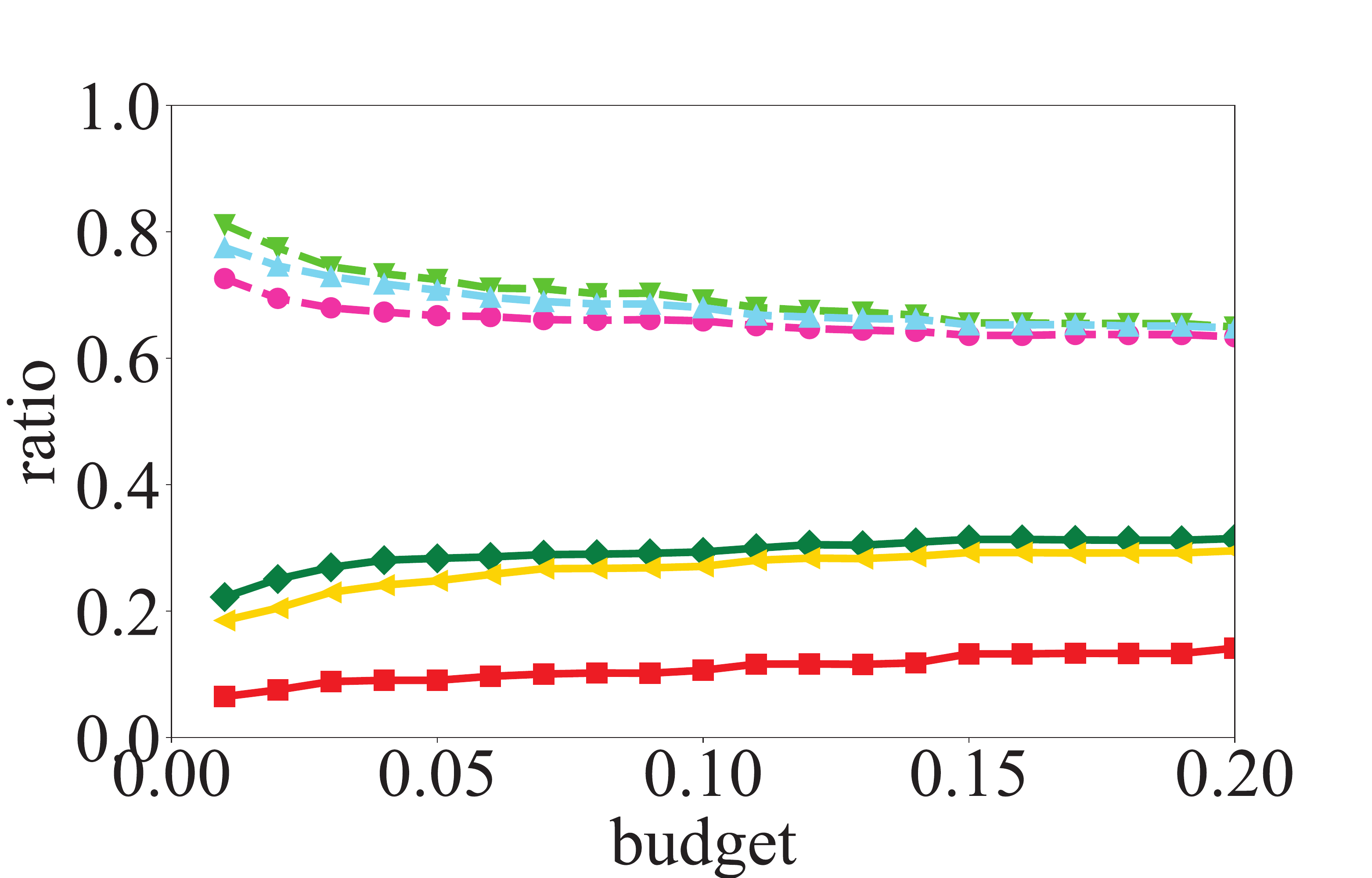}
\label{COIL_typo}}
 \hfil
\subfloat[IMDB]{\includegraphics[width=2.1in]{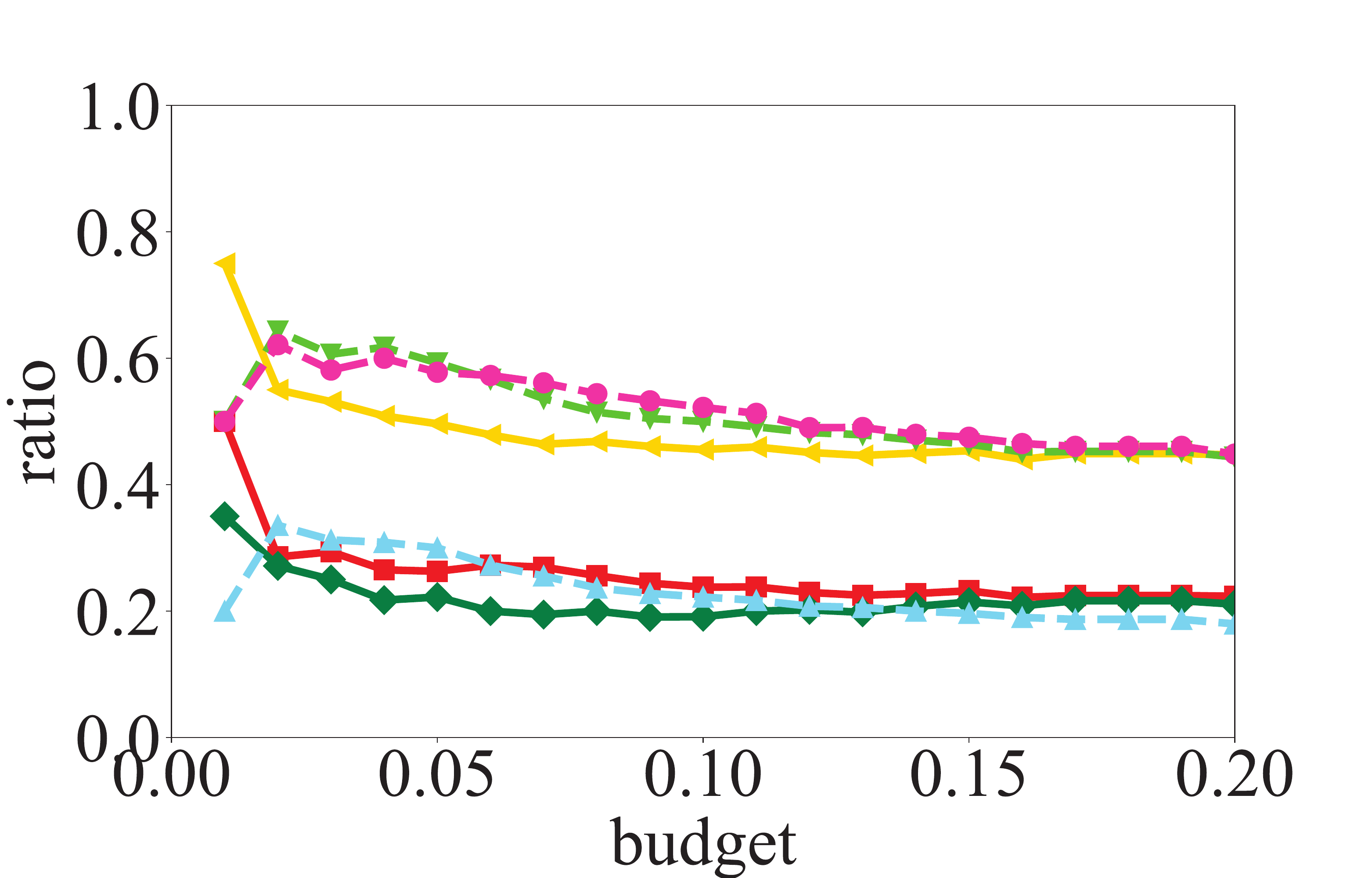}
\label{IMDB_typo}}
\vspace{-4mm}  
\caption{Detection performance vs. budget $b$ on the testing dataset with the training dataset generated by PGD attack.}
\label{fig:detect typo}
\end{figure*}

{\section{Defending against {Adversarial Graphs}}}
\label{sec:detection}
In this section, we propose two different defenses against adversarial graphs: one to detect adversarial graphs and the other \M{to} prevent adversarial graph generation. 

\vspace{+2mm}
\subsection{Adversarial Graph Detection}
{
We first train an adversarial graph detector and then use it to identify whether a testing graph is adversarially perturbed or not.}
We train GNN models as our detector.
Next, we present our methods of generating the training and testing graphs for building the detector,  
and utilize three different structures of GNN models to construct our detectors. 
Finally 
we evaluate our attack under these detectors. 

\subsubsection{Generating datasets for the detector}
\label{sec:detect dataset}
The detection process has two \M{phases}, i.e., training the detector and detecting testing (adversarial) graphs using the trained detector. 
Now, we describe how to generate the datasets for training and detection. 

\noindent \textbf{Testing dataset.} The testing dataset includes all adversarial graphs generated by our attack in Section \ref{sec:experiment results} and their corresponding normal (target) graphs.
We set labels of adversarial graphs and normal graphs to be $1$ and $0$ respectively. 

\noindent \textbf{Training dataset.} The training dataset 
contains normal graphs and adversarial graphs. 
Specifically, we first randomly select a number of normal graphs from the training dataset used to train the target GIN model (see Section \ref{sec:experiment setup}). 
For each sampled normal graph, the detector deploys an adversarial attack to generate the corresponding adversarial graph. 
We use all the sampled normal graphs and the corresponding adversarial graphs to form the training dataset. 
We consider that the detector uses two different attacks to generate the adversarial graphs: (i) the detector uses our attack, and (ii) the detector uses existing attacks. 
In our experiments, without loss of generality, we set the size of training dataset to be 3 times of the size of the testing dataset.

When using existing attacks,
the detector chooses the projected gradient descent (PGD) attack~\cite{xu2019topology}.  
PGD attack is a white-box adversarial attack against GCN model for node classification tasks. It first defines a perturbation budget as the maximum number of edges that the attacker can modify. Then it conducts projected gradient descent to minimize the attacker's objective function. Specifically, in each iteration, the attacker computes the gradients of the objective function w.r.t the edge perturbation matrix. Then it updates the edge perturbation matrix in the opposite direction of the gradient and further projects the edge perturbation matrix into the constraint set such that the number of perturbations is within the pre-set budget. We extend PGD attack 
for graph classification. 
Finally, we choose the three aforementioned GNNs (i.e., GIN~\cite{xu2018powerful}, SAG~\cite{lee2019self} and GUNet~\cite{gao2019graph}) to train the binary detectors on the constructed training dataset. 
\M{Note that, we do not use the aforementioned RL-S2V attack or the Random attack to generate training dataset because RL-S2V attack needs large AT and the random attack needs large AQ to produce a reasonable number of adversarial graphs (see Table \ref{tab:AQ and AT}). In contrast, the PGD attack is more efficient. }

\begin{figure*}[t]
\centering
\subfloat[NCI1]{\includegraphics[width=2in]{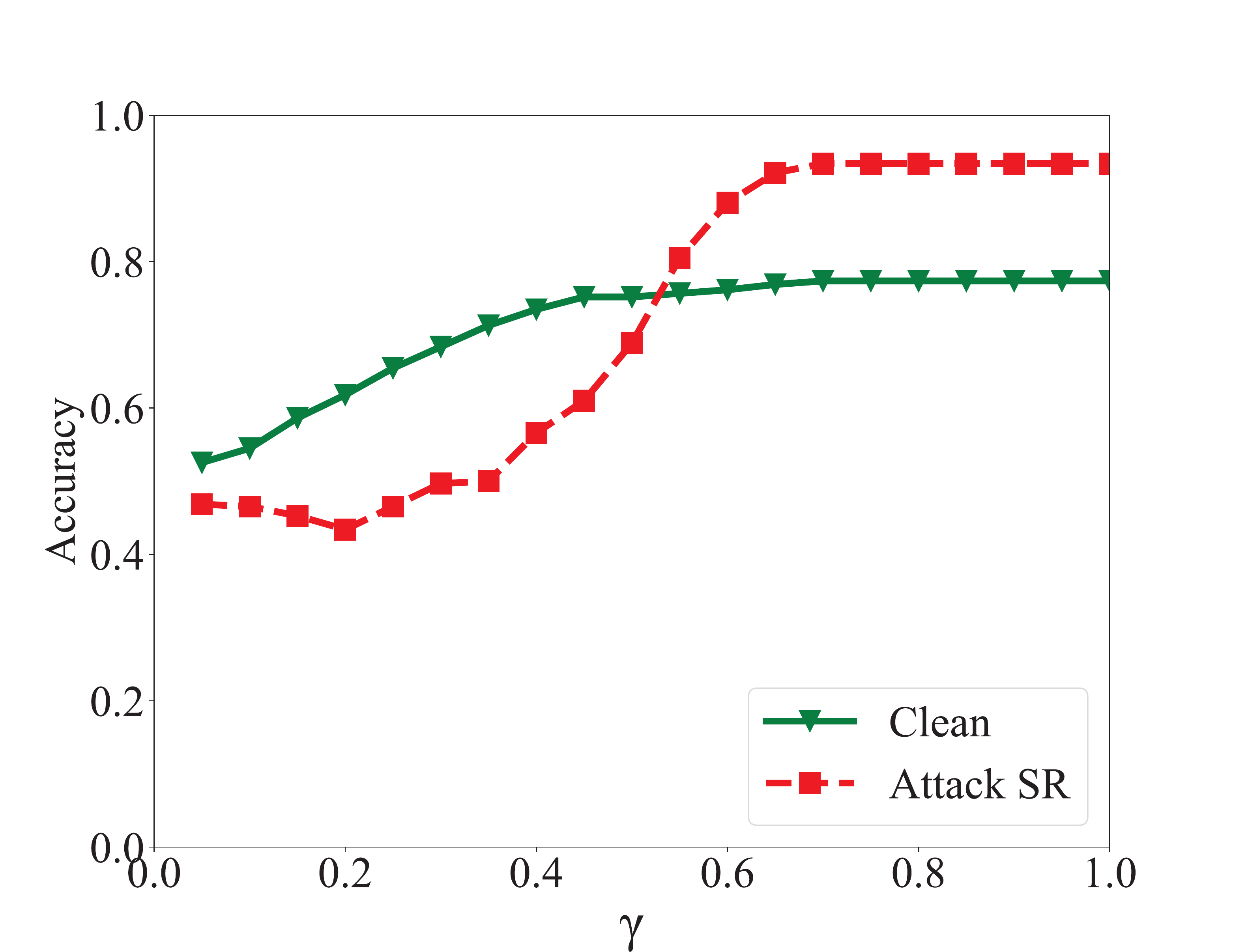}
\label{NCI1_defense}}
 \hfil
\subfloat[COIL]{\includegraphics[width=2in]{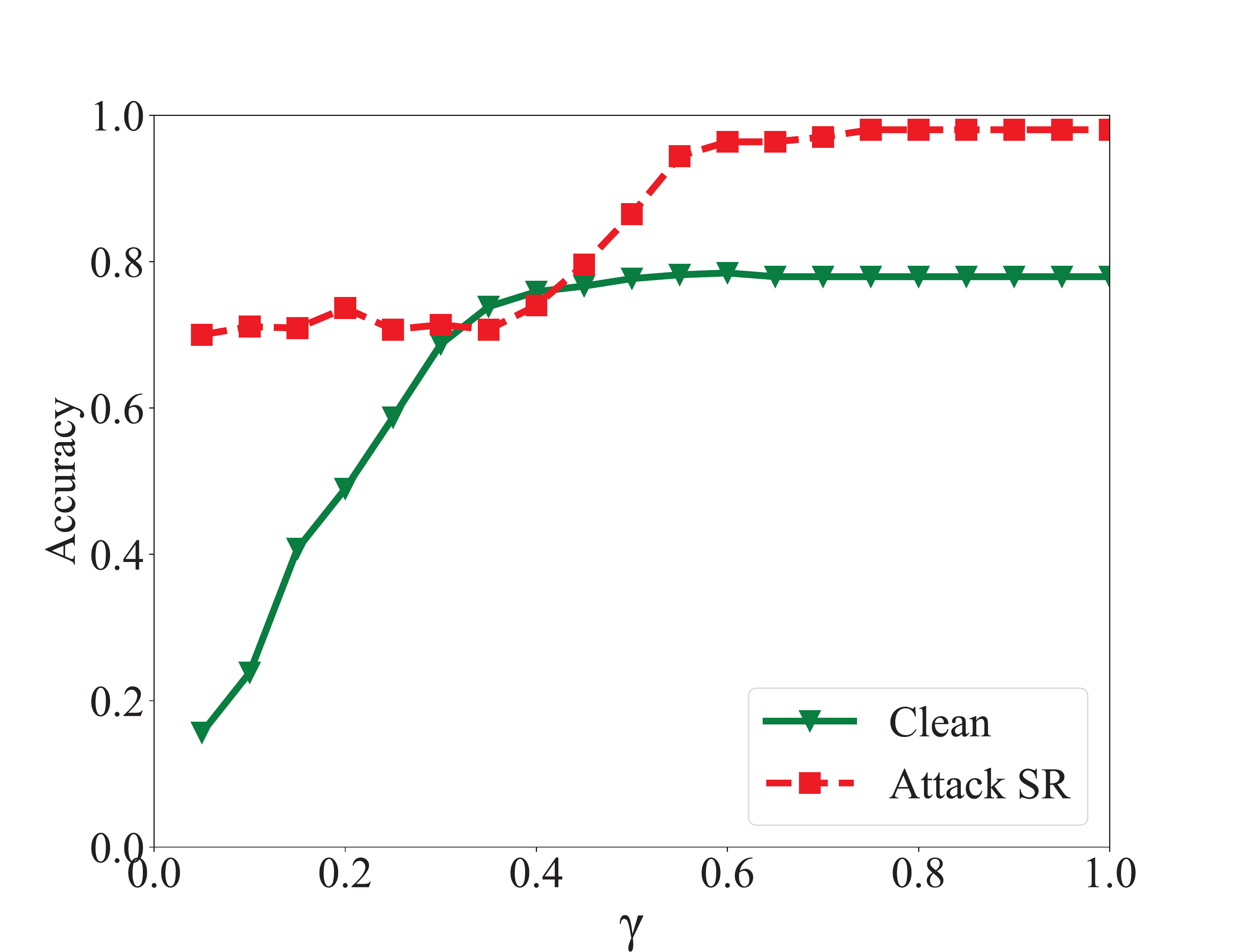}
\label{COIL_defense}}
 \hfil
\subfloat[IMDB]{\includegraphics[width=2in]{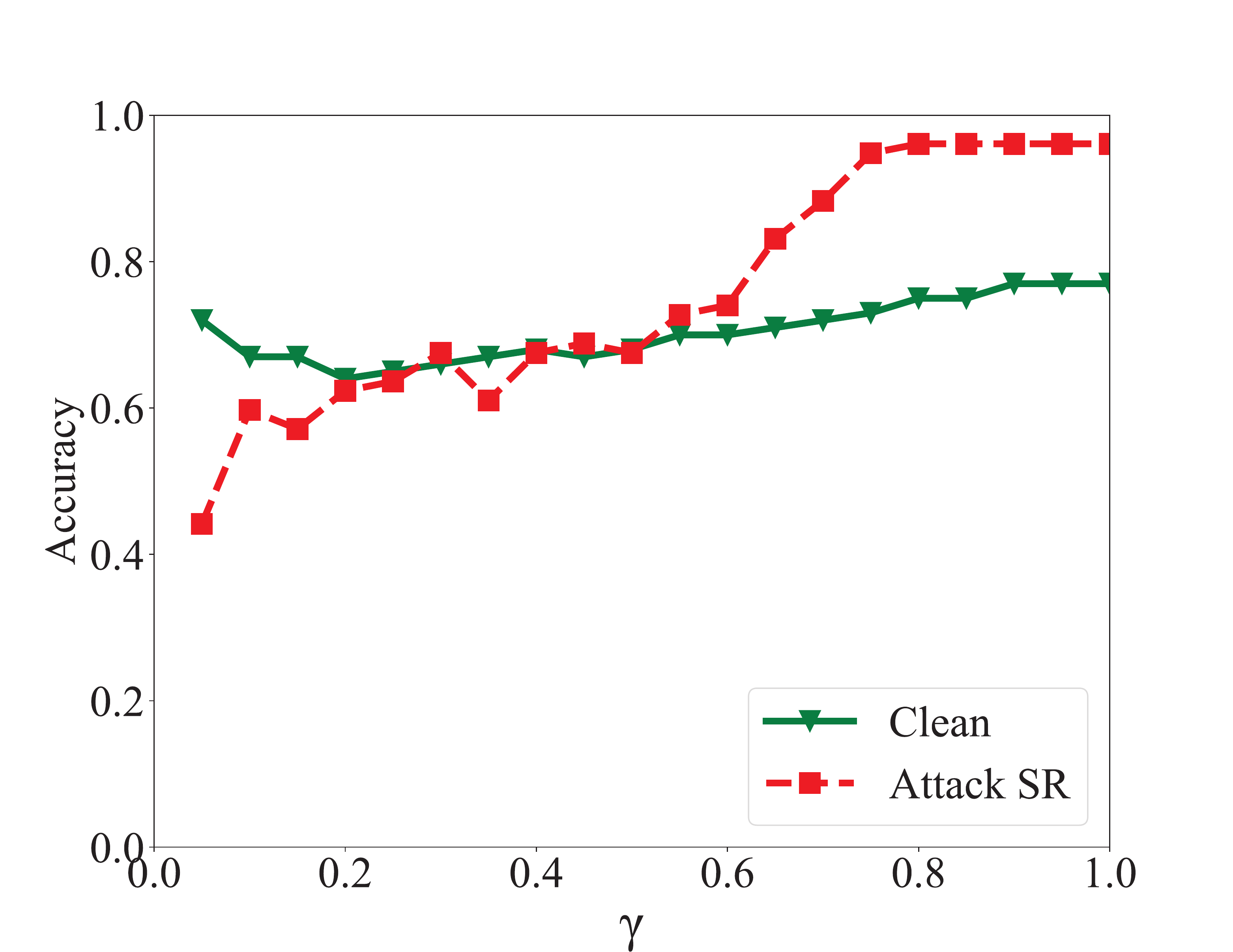}
\label{IMDB_defense}}
\vspace{-2mm}  
\caption{Defense performance against our attack for GIN vs. fraction $\gamma$ of kept top largest singular values.}
\label{fig:defense_GIN}
\vspace{-8mm}
\end{figure*}

\begin{figure*}[t]
\centering
\subfloat[NCI1:SAG]{\includegraphics[width=0.24\textwidth]{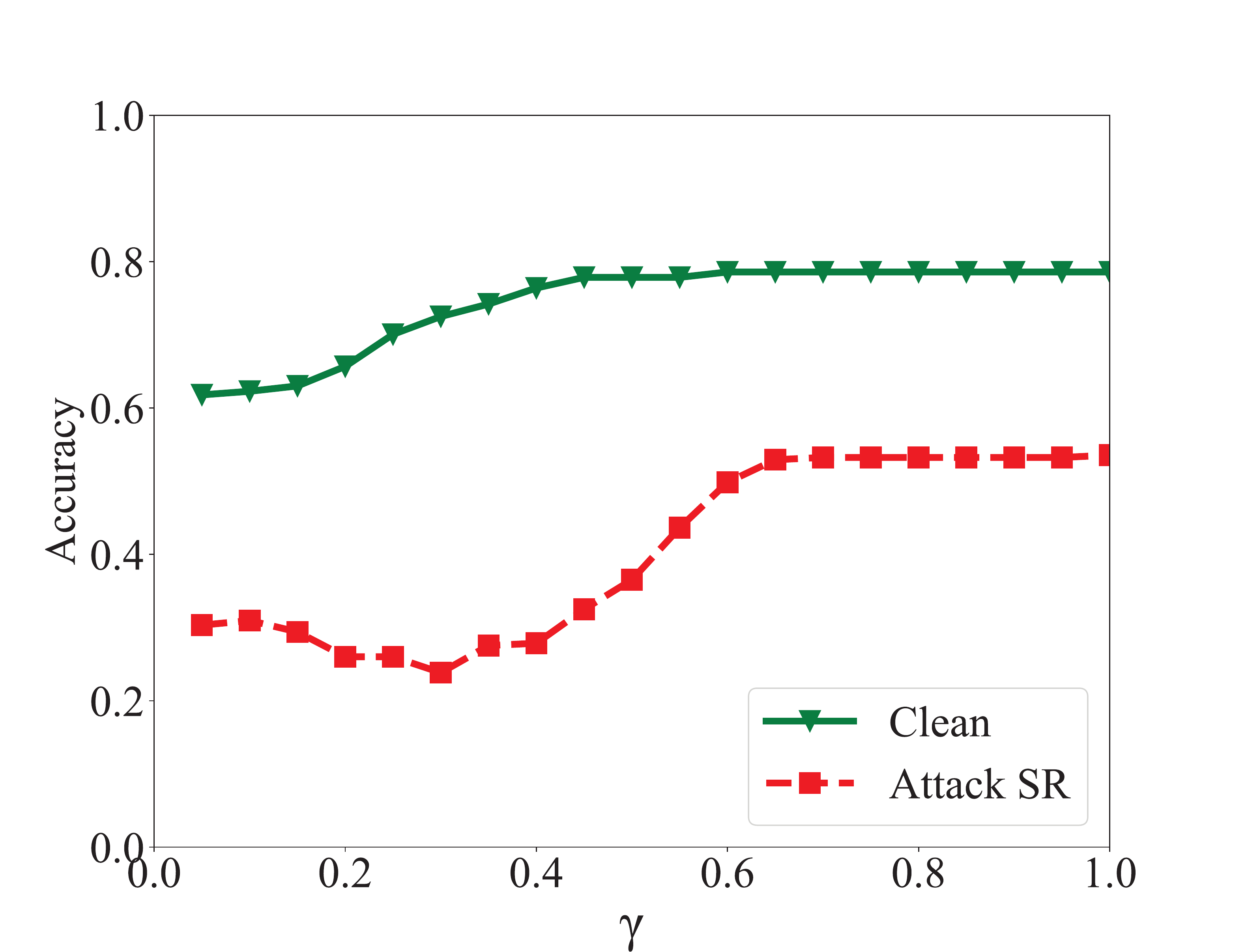}
\label{NCI1_defense_SAG}}
\subfloat[IMDB:SAG]{\includegraphics[width=0.24\textwidth]{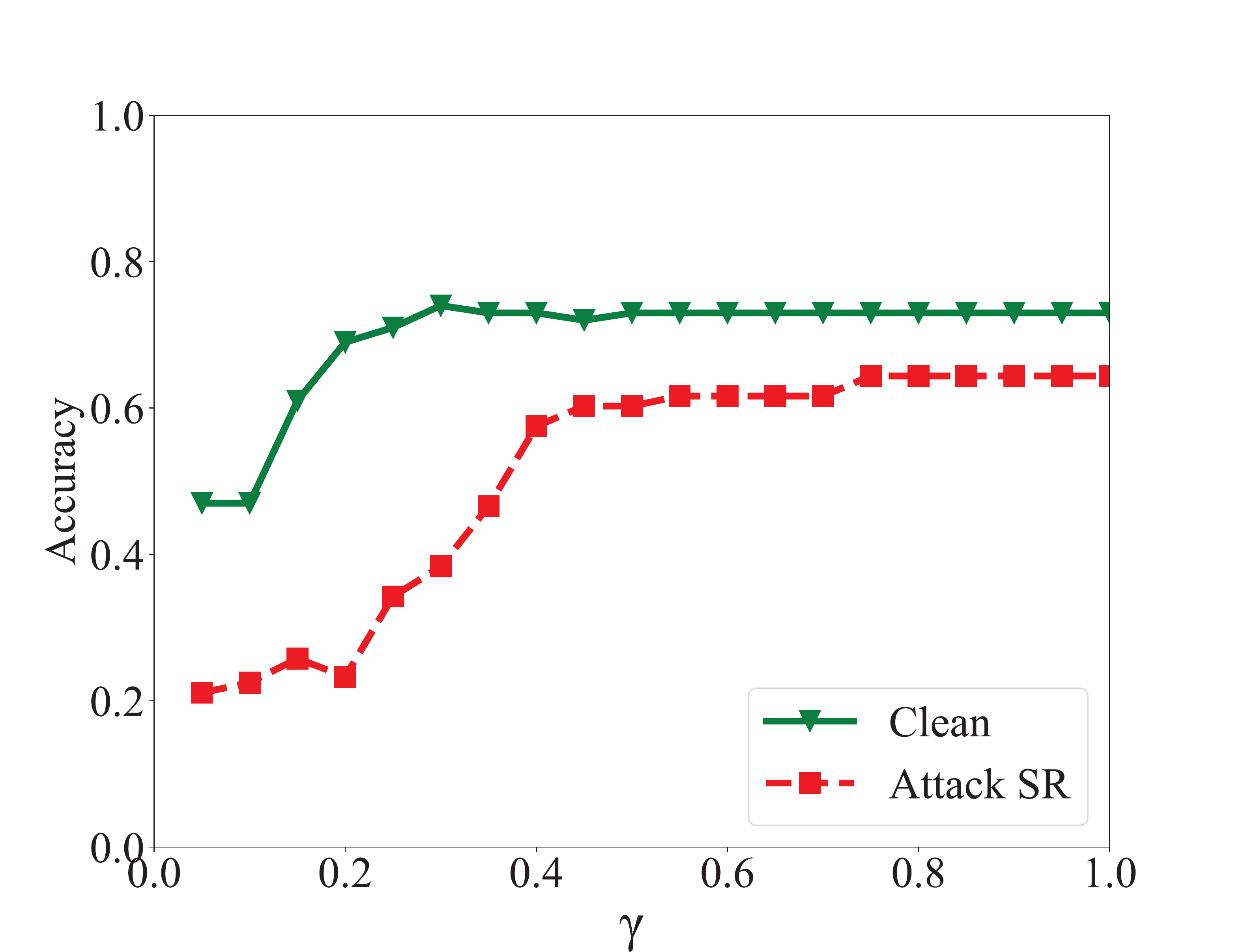}
\label{IMDB_defense_SAG}}
\subfloat[NCI1:GUNet]{\includegraphics[width=0.24\textwidth]{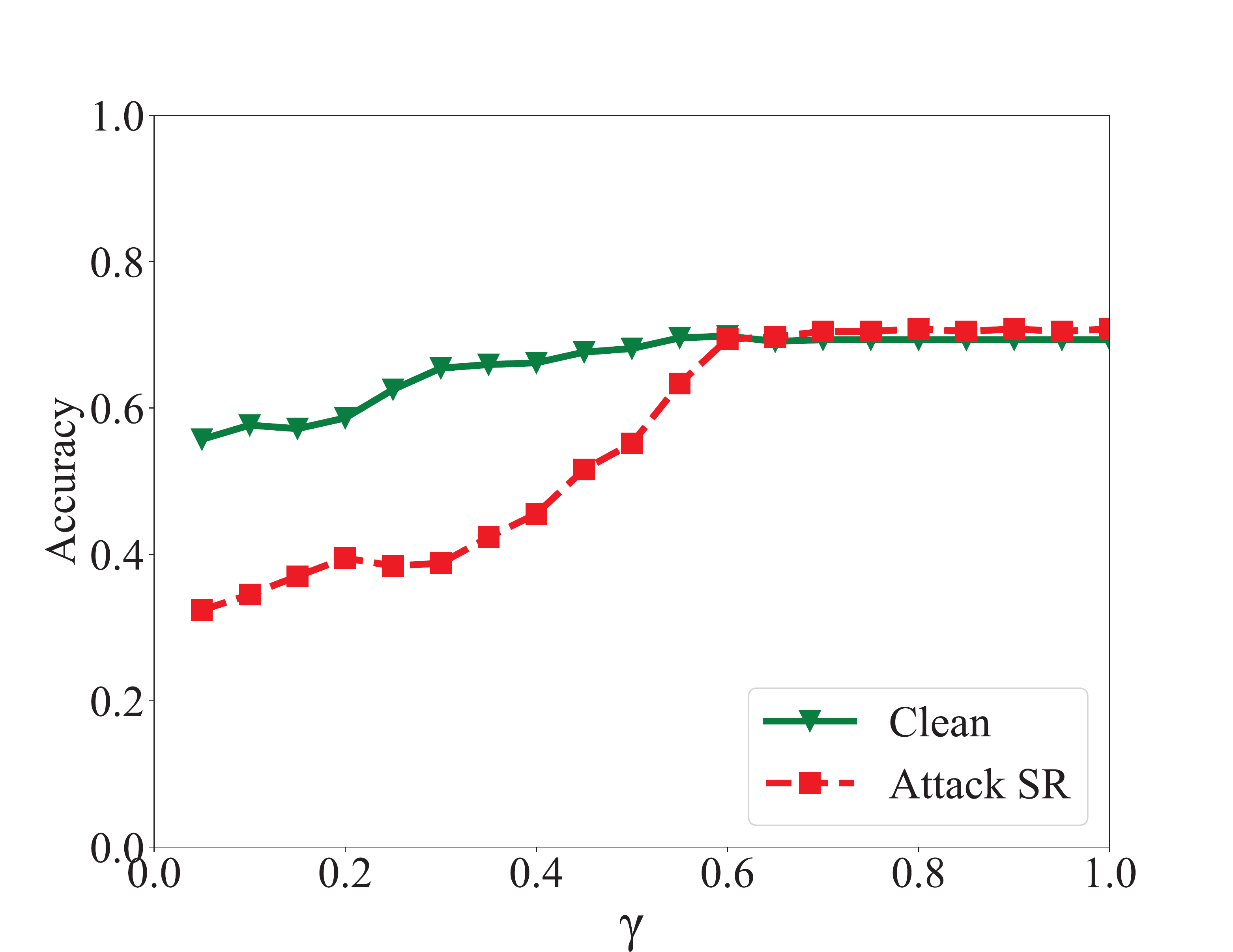}
\label{NCI1_defense_GUN}}
\subfloat[IMDB:GUNet]{\includegraphics[width=0.24\textwidth]{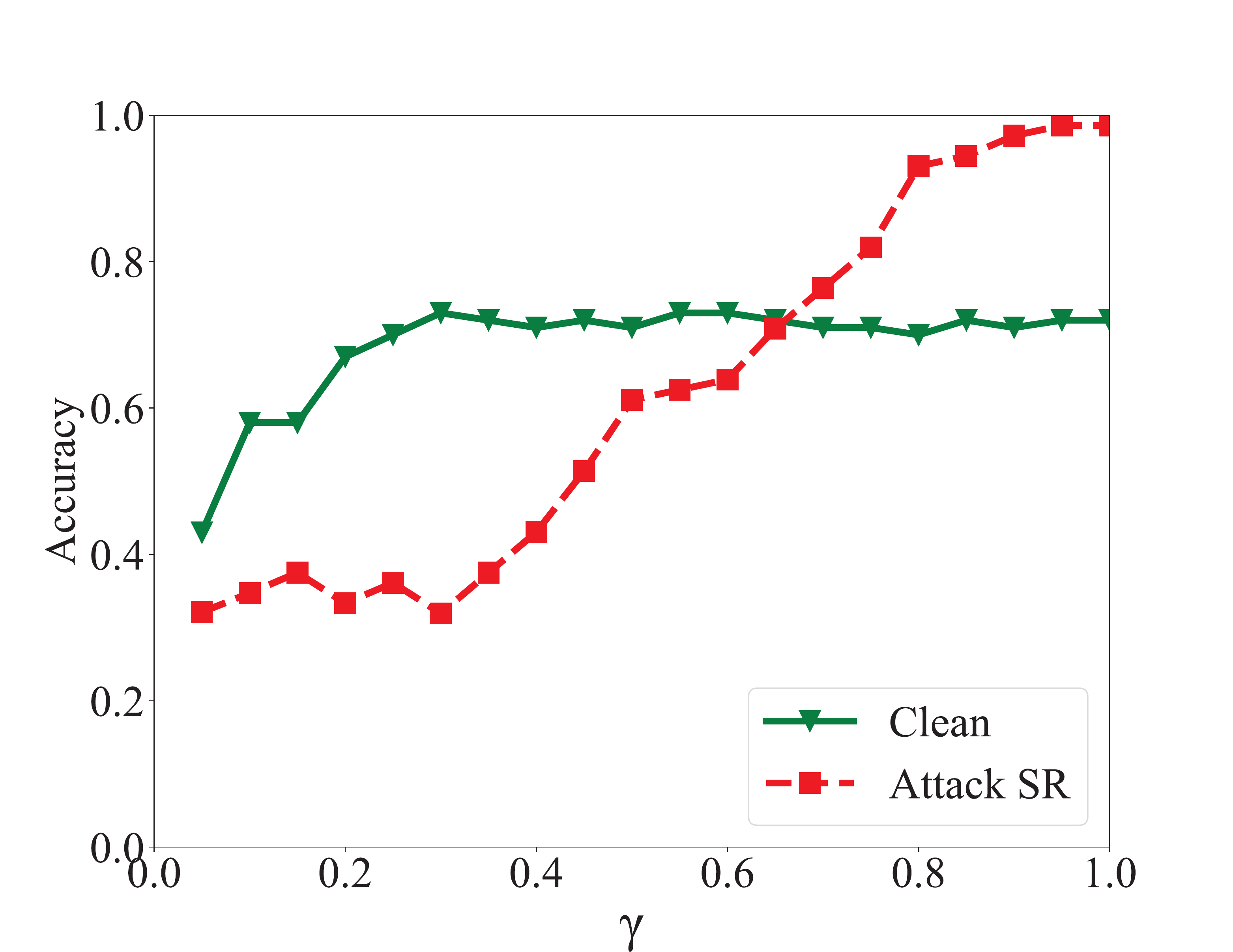}
\label{IMDB_defense_GUN}}
\vspace{-2mm}  
\caption{Defense performance against our attack for SAG and GUNets vs fraction $\gamma$ of kept top largest singular values. }
\vspace{-2mm}
\label{fig:defense_SAG_GUNeet}
\end{figure*}

\vspace{-3mm}
\subsubsection{Detection results}
\label{sec:detect result}
In the detection process, we use \textit{False Positive Rate (FPR)} and \textit{False Negative Rate (FNR)} to evaluate the effectiveness of the detector, where FPR indicates the fraction of normal graphs which are falsely predicted as adversarial graphs, while FNR stands for the fraction of adversarial graphs that are falsely predicted as normal graphs. Again, we repeat each experiment 10 trails and use the average results of them as the final results to ease the influence of randomness.

\noindent \textbf{Detector results under our attack.}
The detection performance vs. budget $b$ on the testing dataset with the \M{training} dataset generated by our attack is shown in Figure \ref{fig:detect our}. Solid lines indicate FPRs and dashed lines indicate FNRs. We have several observations. (i) The detection performance increases as the budget is getting larger, which means that the detector can distinguish more adversarial graphs if the average perturbations of adversarial graphs are larger. For example, the FPR of GUNet detector on IMDB dataset decreases from 0.80 to 0.42 when the budget increases. This is because when  more perturbations \M{are} added to the normal graphs, the structure difference between the corresponding adversarial graphs and the normal graphs is larger. Thus, it is easier to distinguish between them. 
(ii) The detection performances for a specific detector are different on the three datasets. For instance, when using GIN and the budget is $0.20$, the FPRs on the three datasets are similar while the FNRs are \M{quite} different, i.e., 0.48 (COIL), 0.39 (NCI1) and 0.13 (IMDB), respectively. 
This is because the average perturbations of adversarial graphs for COIL \M{are} the smallest (i.e., 4.33), then for NCI1 (i.e., 7.16) and for IMDB \M{are} the largest (i.e., 16.19). 
(iii) The detector is not effective enough. For example, the smallest FNR on the COIL dataset is 0.48, which means that at least 48$\%$ of adversarial graphs cannot be identified by the detector. 
We have similar observation on the NCI1 dataset. 
We guess the reason is that the difference between adversarial graphs and normal graphs is too small on these two datasets, and the detector can hardly distinguish between them. 

\noindent \textbf{Detection results under the PGD attack.}
The detection performance vs. budget $b$ on the testing dataset with the training dataset generated by PGD attack is shown in Figure \ref{fig:detect typo}. Similarly, we can observe that the detection performance is better when the budget is larger. 
However, the detection performance with the PGD attack is worse than that with our attack. For instance, on the NCI1 dataset, FNRs are around 0.70 using the three detectors with the PGD attack, while the detector with our attack achieves as low as 0.25. 
One possible reason is that, when the adversarial graphs in the training set are generated by our attack, the detector trained on these graphs can be relatively easier to generalize to the adversarial graphs in the testing dataset that are also generated by our attack.
On the other hand, when the adversarial graphs in the training set are generated by the PGD attack, it may be more difficult to generalize to the adversarial graphs in the testing dataset. 

\subsection{Preventing Adversarial Graph Generation}
We propose to equip the GNN model with a defense strategy to prevent adversarial graph generation. Here, we generalize the state-of-the-art low-rank based defense~\cite{entezari2020all} against GNN models for node classification to graph classification. 


\vspace{-2mm}
\subsubsection{Low-rank based defense}
{
The main idea is that only high-rank or low-valued singular components of the adjacency matrix of a graph are affected by the adversarial attacks. As these low-valued singular components contain little information of the graph structure, they can be discarded to reduce the effects caused by adversarial attacks, as well as maintaining testing performance of the GNN models. 
In the context of graph classification, 
we first conduct a singular value decomposition (SVD) to the adjacency matrix of each testing graph. Then, we keep the top largest singular values and discard the remaining ones. Based on the top largest singular values, we can obtain a new adjacency matrix, and the corresponding graph whose \M{perturbations} are removed. 

}

\subsubsection{Defense results}
\label{sec:defense results}
We calculate the SR of our attack and the clean testing accuracy after adopting the low-rank based defense. 
We use the target graphs described in Section \ref{sec:experiment setup} to calculate the SR. 
We use the original testing dataset without attack to compute the testing accuracy. 
For each target graph, we first generate a low-rank approximation of its adjacency matrix by removing small singular values and then feed the new graph into the target GNN model to see if the predicted label is correct or wrong. 
Figure~\ref{fig:defense_GIN} and~\ref{fig:defense_SAG_GUNeet}
 show the SR and clean testing accuracy with the low-rank based defense vs. fraction $\gamma$ of kept top largest singular values for the three GNN models, respectively. $\gamma$ ranges from 0.05 to 1.0 with a step of 0.05.
 We have several observations. 
  (i) When $\gamma$ is relatively small (e.g., $\le 0.35$), i.e., a small fraction of top singular values are kept, the clean accuracy decreases and even dramatically on NCI1 and COIL. One possible reason is that the testing graph structure is damaged. On the other hand, the SR also decreases, meaning adversarial perturbations in certain adversarial graphs are removed.  
 (ii) When $\gamma$ is relatively large (e.g., $\ge 0.35$), i.e., a large fraction of top singular values are kept, the clean accuracy maintains and the SR is relatively high as well. This indicates that adversarial perturbations in a few graphs are removed. 
The above observations indicate that the fraction $\gamma$ in low-rank based defense achieves an accuracy-robustness tradeoff. 
In practice, we need to carefully select $\gamma$ in order to obtain high robustness against our attack, as well as promising clean testing performance. 
\M{(iii) When $\gamma$ is extremely small, e.g., $\gamma=0.05$, which means that $95\%$ singular values of a graph are removed, the clean accuracy does not decrease much (e.g., see Figure \ref{fig:defense_SAG_GUNeet} (a) and (c)) or even slightly increases (e.g., see Figure \ref{fig:defense_GIN} (c)). The results are similar to in~\cite{rong2020dropedge}. We note that the labels of graphs may be different from the true labels when 95\% of their smallest singular values are removed even they are not perturbed. 
In our experiments, for ease of analysis, we assume that  these graphs have true labels as our goal is to evaluate the impact of the low-rank based defense, i.e., measure if the attack SR significantly decreases with the maintained clean accuracy. 
}
\subsection{Discussion}
\label{sec:detect discussion}
The defense results in Section \ref{sec:detect result} and Section \ref{sec:defense results} indicate that our adversarial attack is still effective even under detection {or prevention}. 
For example, on the NCI1 dataset, the best detection performance is obtained when the budget is 0.20 with GUNet and our attack to train the detector. However, the FNR and FPR are still 0.25 and 0.20, even if the detector has a full knowledge of our attack. What's worse, it is even harder to detect adversarial graphs in real-world scenarios as the detector often does not know the true attack. 
{Low-rank based defense can prevent 
adversarial graphs to some extent, while needing to scarify the testing performance on clean graphs, e.g., as large as 40\% of adversarial graphs on NCI1 cannot be prevented even we remove 95\% of the smallest singular values. 
}

The proposed detector is a data-level defense strategy and it attempts to block the detected adversarial graphs before they query the target model. It has two key limitations: (i) it is heuristic and (ii) it needs substantial number of adversarial graphs to train the detector and the detection performance highly depends on the quality of the training dataset, i.e., the structure difference between adversarial graphs and normal graphs should be large. 
{The proposed low-rank based defense is a model-level defense strategy and it equips the target GNN model with the smallest singular value removal such that the GNN model can accurately predict testing graphs even they are adversarially perturbed.}
\M{There are several possible ways to empirically strengthen our defense: (i) Locating the vulnerable regions of graphs based on the feedback of our attacks; (ii) Designing attack-aware graph partitioning algorithm as the method used in our attack is generic and does not exploit the setting of adversarial attacks. } 
(iii) \textit{Adversarial training}~\cite{hu2021robust,chen2020smoothing, jin2021robust}. It aims at training a robust GNN model by introducing a white-box adversarial attack and playing a min-max game when training the model. We do not adopt this method because there \M{does} not exist white-box attacks against the considered GNN models. 

Another direction is to provide 
the \textit{certified robustness}~\cite{wang2020certified,jin2020certified,bojchevski2020efficient} of GNN models against adversarial structural perturbations. 
We will also leave those kind of defenses to defend our hard label black box adversarial attacks as the future work.
\section{Related Work}
\label{sec:ad attack}
Existing studies have shown that GNNs are vulnerable to 
adversarial attacks~\cite{chang2020restricted, wang2019attacking,chen2020mga,ma2019attacking,zhang2020backdoor,wang2020evasion}, which deceive a GNN 
to produce wrong labels for specific target graphs (in graph classification tasks) or target nodes (in node classification tasks). According to the stages when these attacks occur, they can be classified into training-time poisoning attacks~\cite{wang2019attacking,zugner2018adversarial, liu2019unified,zugner2019adversarial,xi2020graph} and testing time 
adversarial attacks 
~\cite{tang2020adversarial,chen2018fast,lin2020adversarial,chen2020link, ma2020black,wang2020evasion}. 
In this paper, we focus on testing time adversarial attacks against classification attacks. 

\noindent \textbf{Adversarial attacks against node classification.}  Existing adversarial attacks mainly attack
GNN models for node classification. 
For node classification, the 
attacks can be divided into two categories, i.e., optimization based ones~\cite{takahashi2019indirect,wu2019adversarial,tian2021towards, ma2020black} and heuristic based ones that leverage greedy algorithms~\cite{wang2018attack,chen2018fast} or reinforcement learning (RL)~\cite{dai2018adversarial, sun2019node}. In order to develop an optimization based method, the attacker formulates the attack as an optimization problem and solves it via typical techniques such as gradient descent. For example, Xu et al.~\cite{xu2019topology} developed a CW-type loss as the attacker's objective function and utilized projected gradient descent to minimize the loss. 
As for heuristic based methods, an attacker can utilize a greedy based method, i.e., defining an objective function and traversing all candidate components (e.g., an edge or a node) for adding perturbations. The attacker can select the one that maximizes the objective function to perturb. This process will be repeated multiple times until the attacker \M{finds} an adversarial graph or the perturbations exceed the pre-set budget. For instance, Chen et al.~\cite{chen2018fast} proposed an adversarial attack to GCN, which selects the edge of the maximal absolute link gradient and adds it to graph as the perturbation in each iteration.

\noindent \textbf{Adversarial attacks against graph classification.} Only a few attacks aim to interfere with graph classification tasks~\cite{ma2019attacking,tang2020adversarial,dai2018adversarial}. For instance, Ma et al.~\cite{ma2019attacking} proposed a RL based adversarial attack to GNN, which constructs the attack by perturbing the target graph via rewiring. Tang et al.~\cite{tang2020adversarial} performed the attack against  Hierarchical Graph Pooling (HGP) neural networks via a greedy based method. 
Different from these attacks that are either white-box or grey-box, we study the most challenging \emph{hard label} and \emph{black-box} attack against graph classification in this paper. 
Our attack is both time and query efficient and is also effective, i.e., high attack success rate with small perturbations.  
 
\section{Conclusion} 
We propose a black-box adversarial attack to fool graph neural networks for graph classification tasks in the hard label setting. We formulate the adversarial attack as an optimization problem, which is intractable to solve in its original form. We then relax our attack problem and design a sign stochastic gradient descent algorithm to solve it with convergence guarantee. 
We also propose two algorithms, i.e., coarse-grained searching and query-efficient gradient computation, to decrease the number of queries during the attack. 
We conduct our attack against three representative GNN models on real-world  datasets from different fields. 
The experimental results show that our attack \M{is} more effective and efficient, 
when compared with the state-of-the-art attacks. 
Furthermore, we propose two defense methods to defend against our attack: one to detect adversarial graphs and the other \M{to} prevent adversarial graph generation. 
The evaluation results show that our attack is still effective, which highlights  advanced defenses in future work. 

\begin{acks}
We would like to thank our shepherd Pin-Yu Chen and the anonymous reviewers for their comments. This work is supported in part by the National Key R\&D Program of China under Grant 2018YFB1800304, 
NSFC under Grant 62132011, 61625203 and 61832013,  U.S. ONR under Grant N00014-18-2893, and BNRist under Grant BNR2020 RC01013. 
Qi Li and Mingwei Xu are the corresponding authors of this paper. 
\end{acks}

\bibliographystyle{ACM-Reference-Format}
\bibliography{reference}

\appendix

\section{Background: Graph Neural Network for Graph Classification}
\label{sec:background}
Graph Nerual Networks (GNNs) has been proposed~\cite{lee2019self, gao2019graph, xu2018powerful, hamilton2017inductive, hu2019strategies} to efficiently  process graph data such as social networks, moleculars, financial networks, etc.~\cite{he2020stealing,hamilton2017representation}. GNN learn embedding vectors for each node in the graph, which will be further used in various tasks, e.g., node classification~\cite{kipf2016semi}, graph classification~\cite{xu2018powerful}, community detection~\cite{chen2017supervised} and link prediction~\cite{zhang2018link}. 
Specifically, in each hidden layer, the neural network iteratively computes an embedding vector for a node via aggregating the embedding vectors of the node's neighbors in the previous hidden layer~\cite{xu2018representation}, which is called \textit{message passing}~\cite{gilmer2017neural}. Normally, only the embedding vectors of the last hidden layer will be used for subsequent tasks. For example, in node classification, a logistic regression classifier can be used to classify the final embedding vectors to predict the labels of nodes~\cite{kipf2016semi}; In graph classifications,  information of the embedding vectors in all hidden layers is utilized to jointly determine the graph's label~\cite{ying2018hierarchical,zhang2018end}. According to the strategies of message passing, 
various GNN methods have 
been 
designed for handling specific tasks.  
For instance, Graph Convolutional Network (GCN)~\cite{kipf2016semi}, GraphSAGE~\cite{hamilton2017inductive}, and Simplified Graph Convolution (SGC)~\cite{wu2019simplifying} are mainly for node classification, while Graph Isomorphism Network (GIN)~\cite{xu2018powerful}, SAG~\cite{lee2019self}, 
and Graph U-Nets (GUNet)~\cite{gao2019graph} are for graph classification. 
In this paper, we choose GIN~\cite{xu2018powerful}, SAG~\cite{lee2019self},  
and GUNet as the target GNN models. 
Here, we briefly review GIN as it outperforms other GNN models for graph classification. 

\noindent \textbf {Graph Isomorphism Network (GIN).}
Suppose we are given a graph $G = \left(A, X \right)$
with label $y_0$, where 
$A \in \left\{0,1\right\}^{N\times N}$ is the symmetric adjacent matrix indicating the edge connections in $G$, i.e., $A_{ij} =1$ if there is an edge between node $i$ and node $j$ and $A_{ij}=0$ otherwise. $N$ is the total number of nodes in the graph. $X \in \mathbb{R}^{N \times l}$ is the feature matrix for all nodes, where each row $X_i$ denote the associated $l$-dimensional feature vector of node $i$.
The process of message passing of an $K$-layer GIN can be formulated as follows~\cite{xu2018powerful}:
\begin{equation} \label{eq:GIN}
    h_v^{{k}} = MLP^{(k)}( (1+\epsilon^{(k)})\cdot h_v^{(k-1)} + \sum_{u \in \mathcal{N}_{v}} h_u^{(k-1)}),
\end{equation}
where $ h_v^{{k}} \in \mathbb{R}^{l_k}$ is the embedding vector of node $v$ at the $k$-th layer and, for all nodes, $h_i^{(0)}=X_i$, $MLP$ is a multi-layer perceptron {whose parameters are trained together with the whole GIN model},   $\epsilon^{(k)}$ is a learnable parameter at the $k$-th layer, and $\mathcal{N}_{v}$ is the set of neighbor nodes of node $v$.

To fully utilize the structure information, GIN collects the information from all depth to predict the label of a graph in graph classification tasks. That is, the graph's embedding vector is obtained as follows:
\begin{equation}
    h_G^{(k)} = READOUT(\{h_v^{(k)} | v \in G\}),
\end{equation}
where $h_G^{(k)}$ is the embedding vector of the whole graph at the $k$-th layer and the $READOUT(\cdot)$ function aggregates node embedding vectors in this hidden layer. $READOUT(\cdot)$ can be a simple permutation invariant function (e.g., summation) or a more sophisticated graph pooling function. In this paper, we choose the graph add pooling function {(i.e., adds node features of all nodes in a batch of graphs)} as the $READOUT(\cdot)$ function. GIN finally enables a fully-connected layer to each $h_G^{(k)}$ and sum the results to predict the label of the graph, i.e.,
\begin{equation}
    y_{pred} = softmax(\sum_{k=0}^{K-1} Linear(h_G^{(k)})),
\end{equation}
where $Linear$ is a fully-connected layer and $softmax(\cdot)$ is a softmax layer that maps the logits of GIN to values in $[0,1]$.

 
   
  

\section{Proof of Theorem 4.1}
\label{appendix:proof3}
We restate \cref{theorem3}:
\reduction*
Suppose the graph $G$ is partitioned into $\kappa$ clusters and each cluster has $d_i, i=1,2,\dots,\kappa$ nodes. Note that $N=\sum_{i=1}^\kappa d_i$.

The searching space without coarse-grained searching (CGS) is:
\begin{equation}
    S_{graph} = 2^{\frac{N(N-1)}{2}}
\end{equation}
The total searching space of all supernodes is:
\begin{equation}
    S_{node} = \sum_{i=1}^\kappa 2^{\frac{d_i(d_i-1)}{2}} 
\end{equation}
we define a convex function $f(x)=2^{\frac{x(x-1)}{2}}$ and use Jensen's inequality: 
\begin{equation}
     S_{node} =  \sum_{i=1}^\kappa f(d_i) \ge  \kappa \cdot f(\frac{1}{\kappa}\cdot \sum_{i=1}^\kappa d_i) = \kappa \cdot f(\frac{N}{\kappa}) = \kappa\cdot2^{\frac{d(d-1)}{2}}
\end{equation}
The equal sign of the inequality holds when $d_1=d_2=\dots=d_\kappa = d = \frac{N}{\kappa}$, which means that $\kappa$ clusters contain equal number of nodes.
Similarly, the total searching space of superlinks is:
\begin{equation}
\label{eq:S link}
    S_{link} =\frac{1}{2} \sum_{i=1}^\kappa \sum_{j=1,j\ne i}^\kappa 2^{d_id_j} 
\end{equation}
We define a cluster of convex functions $f_i(x) = 2^{d_ix}, i=1,2,\dots,\kappa$ and again deploy Jensen's inequality to Eq.(\ref{eq:S link}):
\begin{equation}
\begin{split}
     S_{link} & =\frac{1}{2} \sum_{i=1}^\kappa \sum_{j=1,j\ne i}^\kappa f_i(d_j) \ge \frac{1}{2} \sum_{i=1}^\kappa (\kappa-1) \cdot f(\frac{1}{\kappa-1} \sum_{j=1,j\ne i}^\kappa d_j ) \\
     &=  \frac{\kappa-1}{2} \sum_{i=1}^\kappa f(\frac{N-d_i}{\kappa-1}) = \frac{\kappa-1}{2} \sum_{i=1}^\kappa 2^{\frac{Nd_i-d_i^2}{\kappa-1}},
\end{split}
\end{equation}
where the equal sign holds when $d_j=\frac{N-d_i}{\kappa-1}$,$j=1,2,\dots,\kappa, j\ne i$. We further define a convex function $f_l(x)=2^{\frac{Nx-x^2}{\kappa-1}}$ and use Jensen's inequality once again to the above equation, we have :
\begin{equation}
    \begin{split}
         S_{link} & \ge \frac{\kappa-1}{2} \sum_{i=1}^\kappa 2^{\frac{Nd_i-d_i^2}{\kappa-1}}= \frac{\kappa-1}{2} \sum_{i=1}^\kappa f_l(d_i) \\
         & \ge \frac{\kappa-1}{2} \cdot \kappa \cdot f_l(\frac{N}{\kappa}) =  \frac{\kappa(\kappa-1)}{2} 2^{\frac{N^2}{\kappa^2}} = \frac{\kappa(\kappa-1)}{2} 2^{d^2},
    \end{split}
\end{equation}
where the equal sign of the second inequality holds when $d_1=d_2=\dots =d_\kappa=d=\frac{N}{\kappa}$. In general situations, we can assume that this condition holds. Thus, if we fist search within $S_{node}$ and then search within $S_{link}$, $\beta$ can be approximated as follows:
\begin{equation}
\begin{split}
      \beta &= \frac{S_{graph}}{S_{node}+S_{link}} \\ & \approx 2^{\frac{N(N-1)}{2}}  \div [\kappa\cdot2^{\frac{d(d-1)}{2}} + \frac{\kappa(\kappa-1)}{2} \cdot 2^{d^2}]
\end{split}
\end{equation}

Now suppose $d=t\kappa$ (thus $N=\kappa d=t\kappa^2$), where $t>>1$ often in practice. Then, we have: 
\begin{equation}
    \begin{split}
        \beta & \approx 2^{\frac{N(N-1)}{2}}  \div [\kappa\cdot2^{\frac{d(d-1)}{2}} + \frac{\kappa(\kappa-1)}{2} \cdot 2^{d^2}] \\
        & = \frac{2^{\frac{t^2\kappa^4-t\kappa^2+t^2\kappa^2}{2}}}{\kappa\cdot2^{t^2\kappa^2-\frac{t\kappa}{2}}+\frac{\kappa^2-\kappa}{2}\cdot2^{\frac{3t^2\kappa^2}{2}}} \\
        & > \frac{2^{\frac{t^2\kappa^4-t\kappa^2+t^2\kappa^2}{2}}}{\kappa^2\cdot2^{\frac{3\kappa^2\kappa^2}{2}}+\kappa^2\cdot2^{\frac{3t^2\kappa^2}{2}}} \\
        &> \frac{2^{\frac{t^2\kappa^4}{2}}}{2\kappa^2\cdot2^{\frac{3t^2\kappa^2}{2}}} \\
        & = \frac{1}{2\kappa^2}\cdot2^{\frac{t^2(\kappa^4-3\kappa^2)}{2}} \\
    \end{split}
\end{equation}

Finally, $\beta$ in general situations satisfies:
\begin{equation}
     \beta \approx O(2^{\kappa^4})
\end{equation}
%

\section{Proof of Theorem 4.2}
\label{appendix:proof2}
We first restate \cref{theorem:2}:
\monotonous*

    We proof theorem \ref{theorem:2} by showing that $p(\Theta)$ is a monotone increasing function of $g(\Theta)$. Without lose of generality, we assume two constants with $0 < g_1 < g_2$. They represent two points at the same direction $\Theta$ which have distances of $g_1$ and $g_2$ respectively from the original graph $A$. Then we have 
    \begin{equation}
    \begin{split}
    p_1 = \|clip(g_1\Theta-0.5)\|_1  \\
    p_2 = \|clip(g_2\Theta-0.5)\|_1
    \end{split}
    \end{equation}
    For simplicity, we assume that $\Theta = \left\{\Theta_1,\dots,\Theta_d\right\}$ here is a normalized direction vector. We denote $I_+$ as the set of indexes where the corresponding components of $\Theta$ are positive, i.e., $I_+ = \left\{i_1, i_2,\dots, i_l)\right\}$ where $l=\left|I_+ \right|$ and $\Theta_i > 0\ for\ i \in I_+$. As the $clip(\cdot)$ function limits the inputs into $[0,1]$ which will set all negative values as $0$, we can rewrite $p_1$ and $p_2$ as follows
    \begin{equation}
    \begin{split}
    p_1 = \sum_{i \in I_+} (clip(g_1\Theta-0.5))_i \\
    p_2 = \sum_{i \in I_+} (clip(g_2\Theta-0.5))_i
    \end{split}
    \end{equation}  
    Furthermore, the components of $g_1\Theta-0.5$ and $g_2\Theta-0.5$ may also be negative because of the $-0.5$ term. We thus further denote $I_+^{(1)}$ where $g_1\Theta_i-0.5>0\ \forall\ i \in I_+^{(1)}$ and $I_+^{(2)}$ where $g_2\Theta_j-0.5>0\ \forall\ j \in I_+^{(2)}$. It is obvious that $I_+^{(1)} \subseteq I_+^{(2)}$ as $0<g_1 < g_2$ and $\Theta_k>0 \forall k\in I_+^{(1)} \cup I_+^{(2)}$. Then we have
    \begin{equation}
        \begin{split}
            p_2-p_1 &= \sum_{i \in I_+} (clip(g_2\Theta-0.5))_i-\sum_{i \in I_+} (clip(g_1\Theta-0.5))_i \\
            &=  \sum_{i \in I_+^{(2)}} (clip(g_2\Theta-0.5))_i-\sum_{i \in I_+^{(1)}} (clip(g_1\Theta-0.5))_i \\
            &= \sum_{i \in I_+^{(1)}} (clip(g_2\Theta-0.5)-clip(g_1\Theta-0.5))_i \\ & + \sum_{j \in I_+^{(2)} \setminus I_+^{(1)}} (clip(g_2\Theta-0.5))_j \\
            & \ge  \sum_{i \in I_+^{(1)}} (g_2-g_1)\Theta_i + \sum_{j \in I_+^{(2)} \setminus I_+^{(1)}} (clip(g_2\Theta-0.5))_j \\
            & \ge 0
        \end{split}
    \end{equation}
    The equal sign holds when one of the following two conditions satisfied:
    
    (i) $I_+^{(1)} = I_+^{(2)} = I_+$, which means that $g_1$ and $g_2$ are both large enough such that all positive components of $g_1\Theta-0.5$ and $g_2\Theta-0.5$ exceed 1.0. Under this condition, we will perturb all edges correspond to the positive components of $g_1\Theta-0.5$.
    
    (ii) $I_+^{(1)} = I_+^{(2)} = \varnothing$, which means that $g_1$ and $g_2$ are both small enough such that all positive components of $g_1\Theta-0.5$ and $g_2\Theta-0.5$ lower than 0. Under this condition, we do not perturb any edge.
    
    Note that, the two conditions above will never be satisfied during our signSGD because we always start our gradient descent at an initial direction $\Theta_0$ with a moderate $g$ value. At each time $t$, when we step to a new direction, i.e., $\Theta_{t+1}$, we may go into an extreme condition where $I_+^{(t+1)} = I_+$ or $I_+^{(t+1)} = \varnothing$. However, both conditions will be rejected as the former leads to large perturbations and the later add no perturbations thus will never change the label of target graph. Therefore, during our signSGD, we will always have
    \begin{equation}
        p_1<p_2, \quad \forall\  0<g_1<g_2
    \end{equation}
    Then $p$ is a monotonically increasing function of $g$ thus $p$ and $g$ can be mutually uniquely determined.

\section{Proof of Theorem 4.3}
\label{appendix:proof1}
We first restate  
\cref{theorem:1}:
\convergence* 
Recall that $\eta_t$ is the learning rate of signSGD in Algorithm \ref{alg:attack} at the $t$-th iteration, $\mu > 0$ is the smoothing parameter and $d$ is the dimension of $\Theta$.

We first define some notations as follows:
\begin{equation} \label{eq:appen1}
    \dot{\triangledown}p(\Theta_t;u_q) = \frac{p(\Theta_t+\mu u_q)-p(\Theta_t)}{\mu}u_q
\end{equation}

\begin{equation}\label{eq:appen2}
    \hat{\triangledown}p(\Theta_t;u_q) =  sign(\frac{p(\Theta_t+\mu{u}_q)-p({\Theta}_t)}{\mu}{u}_q)
\end{equation}

\begin{equation}\label{eq:appen3}
    p_{\mu}(\Theta) = \mathbb{E}_{u}[p(\Theta+\mu{u})]
\end{equation}

\begin{equation}\label{eq:appen4}
    \delta_l = \sqrt{\mathbb{E}[(\dot{\triangledown}p(\Theta_t;u_q)-\triangledown p_{\mu}(\Theta_t))_l^2]}
\end{equation}
where $ p_{\mu}(\Theta)$ is the randomized smoothing function of $p(\Theta)$. We can observe that $\hat{\triangledown}p(\Theta_t;u_q) = sign( \dot{\triangledown}p(\Theta_t;u_q) )$.
Moreover, 
the corresponding estimated gradients are defined as:
\begin{equation}
\small
\label{eq:appen5}
    \dot{p}_t \approx \frac{1}{Q}\sum_{q=1}^{Q}  \frac{p(\Theta_t+\mu u_q)-p(\Theta_t)}{\mu}{u}_q=  \frac{1}{Q}\sum_{q=1}^{Q} \dot{\triangledown}p(\Theta_t;u_q)
\end{equation}

\begin{equation}\label{eq:appen6}
\small
    \hat{p}_t \approx \frac{1}{Q}\sum_{q=1}^{Q}  sign(\frac{p(\Theta_t+\mu u_q)-p(\Theta_t)}{\mu}{u}_q) = \frac{1}{Q}\sum_{q=1}^{Q}  \hat{\triangledown}p(\Theta_t;u_q)
\end{equation}

Next, we introduce some lemmas. 
\begin{lemma} \label{lemma1}
    $\small \left|(\triangledown p_{\mu}(\Theta_t))_l\right| Pr[sign((\hat{p}_t)_l) \ne sign((\triangledown p_{\mu}(\Theta_t))_l)] \le \frac{\delta_l}{\sqrt{Q}}$.
\end{lemma}
\begin{proof}
The proof can be found in \cite{cheng2019sign} Lemma 2.
\end{proof}

\begin{lemma}\label{lemma2}
    $\mathbb{E}[\| \dot{\triangledown}p(\Theta_t;u_q)- \triangledown p_{\mu}(\Theta_t)\|_2^2] \le \frac{4(Q+1)}{Q}\sigma^2+\frac{2}{Q}C(d,\mu)$,
where $C(d,\mu)=2d\sigma^2+\frac{\mu^2 L^2 d^2}{2}$.
\end{lemma}
\begin{proof}
The proof can be found in \cite{liu2018signsgd} proposition 2 with $b=1$, $q=Q$, $\alpha_b=1$ and $\beta_b=0$. As the number of objective function is just one in our optimization problem, so we can choose $b=1$. Then $\alpha_b$ and $\beta_b$ can be further fixed.
\end{proof}

\begin{lemma}
    \label{lemma3}
    $p_{\mu}(\Theta_1) - p_{\mu}(\Theta_T) \le p_{\mu}(\Theta_1) - p^{*} + \mu^2 L$,
where $p^{*}$ is the minimal value of $p(\Theta)$.
\end{lemma}
\begin{proof}
The proof can be found in \cite{liu2018zeroth} Lemma C.
\end{proof}

\begin{lemma} \label{lemma4}
    $\mathbb{E}[\|\triangledown p(\Theta)\|_2] \le \sqrt{2}\mathbb{E}[\|\triangledown p_{\mu}(\Theta) \|_2]+\frac{\mu Ld}{\sqrt{2}}$
\end{lemma}
\begin{proof}
The proof can be found in \cite{liu2018signsgd}.
\end{proof}

Now we prove our Theorem \ref{theorem:1}. 
As $p(\Theta)$ has an $L$-Lipschitz continuous gradient, it is known from \cite{nesterov2017random} that $p_{\mu}(\Theta)$ also has $L$-Lipschitz continuous gradient. Based on the $L$-smoothness of $p_{\mu}(\Theta)$, we have
\begin{equation}\label{eq:appen7}
\small
\begin{split}
    & p_{\mu}(\Theta_{t+1}) \le p_{\mu}(\Theta_{t}) + \left \langle \triangledown p_{\mu} (\Theta_t), \Theta_{t+1}-\Theta_t \right \rangle + \frac{L}{2} \|\Theta_{t+1}-\Theta_t \|_2^2  \\
    & =  p_{\mu}(\Theta_{t})- \eta_t \left \langle \triangledown p_{\mu} (\Theta_t), \hat{p}_t \right\rangle + \frac{L}{2}\eta_t^2\| \hat{p}_t\|_2^2  \\
\end{split}
\end{equation}
Moreover,  we define $(S_t)_l=\frac{1}{Q}\left|\sum_{q=1}^Q \hat{\triangledown}p(\Theta_t; u_q)_l\right|$, and thus $\hat{p}_t = S_t \odot sign(\hat{p}_t )$ and $\|\hat{p}_t \|_2 = \|S_t\|_2$. We can also have
 \begin{equation}\label{eqn:30}
     \small
     \begin{split}
          & \left \langle \triangledown p_{\mu} (\Theta_t), \hat{p}_t \right\rangle  = \|\triangledown p_{\mu} (\Theta_t)\|_2 \|\hat{p}_t\|_2 cos(\alpha_{1t}) \\
          &= \|\triangledown p_{\mu} (\Theta_t)\|_2\|S_t\|_2cos(\alpha_{1t}) \frac{cos(\alpha_{2t})}{cos(\alpha_{2t})}\frac{\|sign(\hat{p}_t)\|_2}{\|sign(\hat{p}_t)\|_2} \\
          &= \|\triangledown p_{\mu} (\Theta_t)\|_2\|sign(\hat{p}_t)\|_2cos(\alpha_{2t}) \cdot \frac{cos(\alpha_{1t})}{cos(\alpha_{2t})} \frac{\|S_t\|_2}{\sqrt{d}} \\
          &=  \left \langle \triangledown p_{\mu} (\Theta_t), sign(\hat{p}_t) \right\rangle \cdot  \frac{cos(\alpha_{1t})}{cos(\alpha_{2t})} \frac{\|S_t\|_2}{\sqrt{d}}, 
     \end{split}
 \end{equation}
 where $\alpha_{1t}$ is the angle between $\triangledown p_{\mu} (\Theta_t)$ and $\hat{p}_t$ and $\alpha_{2t}$ is the angle between $\triangledown p_{\mu} (\Theta_t)$ and $sign(\hat{p}_t)$. 
 Substituting Eq. (\ref{eqn:30}) into Eq. (\ref{eq:appen7}), and defining  $\hat{\eta}_t = \eta_t\cdot\frac{cos(\alpha_{1t})}{cos(\alpha_{2t})} \frac{\|S_t\|_2}{\sqrt{d}}$,we have
 \begin{equation}
     \footnotesize
     \begin{split}
          p_{\mu}(\Theta_{t+1}) & \le p_{\mu}(\Theta_{t}) -\hat{\eta}_t \left \langle \triangledown p_{\mu} (\Theta_t), sign(\hat{p}_t) \right\rangle+ \frac{dL}{2}\hat{\eta}_t^2\frac{cos(\alpha_{2t})^2}{cos(\alpha_{1t})^2} \\
          &= p_{\mu}(\Theta_{t})-\hat{\eta}_t \|\triangledown p_{\mu} (\Theta_t) \|_1 + \frac{dL}{2}\hat{\eta}_t^2\frac{cos(\alpha_{2t})^2}{cos(\alpha_{1t})^2} \\
          &+  2 \hat{\eta}_t\sum_{l=1}^d \left|(\triangledown p_{\mu} (\Theta_t))_l\right|\mathcal{I}[sign((\hat{p}_t)_l) \ne sign((\triangledown p_{\mu}(\Theta_t))_l)]
     \end{split}
 \end{equation}
Let $c_t=\frac{cos(\alpha_{2t})}{cos(\alpha_{1t})}$ and take expectation on both sides, we have
\begin{equation} \label{eq:appen8}
    \footnotesize
    \begin{split}
        &\mathbb{E}[p_{\mu}(\Theta_{t+1})-p_{\mu}(\Theta_{t})] \le -\hat{\eta}_t \|\triangledown p_{\mu} (\Theta_t) \|_1 + \frac{dL}{2}\hat{\eta}_t^2 c_t^2\\
       & + 2 \hat{\eta}_t\sum_{l=1}^d \left|(\triangledown p_{\mu} (\Theta_t))_l \right| Prob[(\hat{p}_t)_l \ne sign((\triangledown p_{\mu}(\Theta_t))_l)]
    \end{split}
\end{equation}
Applying Lemma \ref{lemma1} into the inequality, we have
\begin{equation}\label{eq:appen9}
\footnotesize
\begin{split}
 &\mathbb{E}[p_{\mu}(\Theta_{t+1})-p_{\mu}(\Theta_{t})] \le - \hat{\eta}_t \|\triangledown p_{\mu} (\Theta_t) \|_1 +  \frac{dL}{2}\hat{\eta}_t^2 c_t^2+\frac{ 2 \hat{\eta}_t}{\sqrt{Q}} \sum_{l=1}^d \delta_l
\end{split}
\end{equation}

Note that
\begin{equation}\label{eq:appen10}
\footnotesize
\begin{split}
    & \sum_{l=1}^d \delta_l \le  \|\delta\|_1 \le \sqrt{d} \|\delta \|_2 \\
    & = \sqrt{d} \sqrt{\mathbb{E}[\| \dot{\triangledown}p(\Theta_t;u_q)- \triangledown p_{\mu}(\Theta_t)\|_2^2]} \\
    & \le \sqrt{\frac{d}{Q}}\sqrt{ 4(Q+1)\sigma^2+2C(d,\mu)},
\end{split}
\end{equation}
where we apply Lemma \ref{lemma2} in the last inequality in Equation (\ref{eq:appen10}). Substituting Equation (\ref{eq:appen10}) into Equation (\ref{eq:appen9}), we have
\begin{equation}\label{eq:appen11}
\small
    \begin{split}
    \hat{\eta}_t \|\triangledown p_{\mu} (\Theta_t) \|_1  &\le \mathbb{E}[p_{\mu}(\Theta_{t})-p_{\mu}(\Theta_{t+1})]  +  \frac{dL}{2}\hat{\eta}_t^2 c_t^2  \\ &+ \frac{2\sqrt{d}\hat{\eta}_t}{Q} \sqrt{4(Q+1)\sigma^2+2C(d,\mu)}
    \end{split}
\end{equation}

By summing all inequalities for all $t$s we obtain
\begin{equation}\label{eq:appen12}
\small
\begin{split}
    \sum_{t=1}^T \hat{\eta}_t \mathbb{E}[\|\triangledown p_{\mu}(\Theta_t)\|_1] \le \mathbb{E}[p_{\mu}(\Theta_1)-p_{\mu}(\Theta_T)] +  \frac{dL}{2}\sum_{t=1}^T\hat{\eta}_t^2 c_t^2 \\
    + \sum_{t=1}^T\frac{2\sqrt{d}\hat{\eta}_t}{Q} \sqrt{4(Q+1)\sigma^2+2C(d,\mu)}
\end{split}
\end{equation}
Further substituting Lemma \ref{lemma3} into  Inequality (\ref{eq:appen12}), we have
\begin{equation}\label{eq:appen13}
\small
\begin{split}
    \sum_{t=1}^T \hat{\eta}_t \mathbb{E}[\|\triangledown p_{\mu}(\Theta_t)\|_1] \le  p_{\mu}(\Theta_1) - p^{*} + \mu^2 L +  \frac{dL}{2}\sum_{t=1}^T\hat{\eta}_t^2 c_t^2\\
    + \sum_{t=1}^T\frac{2\sqrt{d}\hat{\eta}_t}{Q} \sqrt{4(Q+1)\sigma^2+2C(d,\mu)}
\end{split}
\end{equation}

Dividing $\sum_{t=1}^T \hat{\eta}_t$ on both sides and use the property that $\|\triangledown p_{\mu}(\Theta_t)\|_2 \le \|\triangledown p_{\mu}(\Theta_t)\|_1$, the inequality (\ref{eq:appen13}) can be changed into
\begin{equation}
\small
    \begin{split}
        \sum_{t=1}^T \frac{ \hat{\eta}_t}{\sum_{t=1}^T \hat{\eta}_t} \mathbb{E}[\|\triangledown p_{\mu}(\Theta_t)\|_2] \le \frac{ p_{\mu}(\Theta_1) - p^{*} + \mu^2 L }{\sum_{t=1}^T \hat{\eta}_t}\\
        +\frac{dL}{2} \frac{\sum_{t=1}^T\hat{\eta}_t^2c_t^2}{\sum_{t=1}^T \hat{\eta}_t} + \frac{2\sqrt{d}}{Q} \sqrt{4(Q+1)\sigma^2+2C(d,\mu)}
    \end{split}
\end{equation}

If we randomly pick $R$ from $\left\{1,\dots,T\right\}$ with probability $P\left(R=t\right)=\frac{\hat{\eta}_t}{\sum_{t=1}^{T} \hat{\eta}_t}$, we will have 
\begin{align}
\label{eq:appenxx}
\small
& \mathbb{E}[\|\triangledown p_{\mu}(\Theta_R)\|_2] = \mathbb{E}[\mathbb{E}_R [\|\triangledown p_{\mu}(\Theta_R)\|_2]]  \nonumber \\
& = \mathbb{E} [\sum_{t=1}^T P(R=t)\|\triangledown p_{\mu}(\Theta_t)\|_2]
\end{align}

Applying Lemma \ref{lemma4} into the Equation (\ref{eq:appenxx}), we have
\begin{equation} \label{eq:append14}
\small
    \begin{split}
         \mathbb{E}[\|\triangledown p(\Theta)\|_2] \le \frac{\sqrt{2}( p_{\mu}(\Theta_1) - p^{*} + \mu^2 L) }{\sum_{t=1}^T \hat{\eta}_t} 
         + \frac{dL}{\sqrt{2}} \frac{\sum_{t=1}^T\hat{\eta}_t^2c_t^2}{\sum_{t=1}^T \hat{\eta}_t} \\
         + \frac{\mu Ld}{\sqrt{2}} + \frac{2\sqrt{2d}}{Q} \sqrt{4(Q+1)\sigma^2+2C(d,\mu)}
    \end{split}
\end{equation}

By choosing $\mu = O(\frac{1}{\sqrt{dT}})$ and $\eta_t=\eta=O(\frac{1}{\sqrt{dT}})$, the convergence rate in (\ref{eq:append14}) simplifies to
\begin{equation}
\small
     \mathbb{E}[\|\triangledown p(\Theta)\|_2] \le O(\frac{\sqrt{d}L}{\sqrt{T}}+\frac{\sqrt{d}}{\sqrt{Q}}\sqrt{Q+d}).
\end{equation}

\end{document}